\documentclass{article}

\usepackage{silence}
\usepackage[table,dvipsnames]{xcolor}
\definecolor{mydarkblue}{rgb}{0,0.08,0.45}
\definecolor{note_fontcolor}{rgb}{0.800781, 0.800781, 0.800781}

    \PassOptionsToPackage{numbers, compress, sort}{natbib}

     \usepackage[final]{neuripsx}

\usepackage[utf8]{inputenc} %
\usepackage[T1]{fontenc}    %
\usepackage[hyphens]{url}            %
\usepackage[
    colorlinks,
    bookmarks=true,
    breaklinks=true,
    linkcolor=mydarkblue,
    citecolor=mydarkblue,
    filecolor=mydarkblue,
    urlcolor=mydarkblue,]{hyperref}       %
\usepackage{booktabs}       %
\usepackage{amsfonts}       %
\usepackage{nicefrac}       %
\usepackage{microtype}      %
\usepackage{multicol}
\usepackage{subcaption}
\usepackage{amssymb, amsmath, amsthm}
\usepackage{thmtools}
\usepackage{mathtools}
\usepackage{tcolorbox}

\usepackage{subfiles}

\usepackage{wrapfig}

\usepackage{algorithm}
\usepackage[noend]{algpseudocode}
\makeatletter
\renewcommand{\ALG@name}{\netsort{} Program}
\makeatother

\usepackage[inline]{enumitem}
\usepackage{breakurl}
\usepackage[capitalize,nameinlink]{cleveref}
\usepackage{mymacros}

\newtheorem{thm}{Theorem}[section]
\newtheorem{cor}[thm]{Corollary}
\newtheorem{fact}[thm]{Fact}

\newtheorem{assm}[thm]{Assumption}
\newtheorem{heur}[thm]{Heuristic}
\newtheorem{setup}[thm]{Setup}
\newtheorem{lemma}[thm]{Lemma}
\newtheorem{prop}[thm]{Proposition}
\theoremstyle{definition}
\newtheorem{defn}[thm]{Definition}
\theoremstyle{remark}
\newtheorem{remk}[thm]{Remark}
\newtheorem{exmp}[thm]{Example}

\newtheorem{cond}{Condition}

\makeatletter
\@ifpackageloaded{hyperref}%
  {\newcommand{\mylabel}[2]%
    {\protected@write\@auxout{}{\string\newlabel{#1}{{#2}{\thepage}%
      {\@currentlabelname}{\@currentHref}{}}}}}%
  {\newcommand{\mylabel}[2]%
    {\protected@write\@auxout{}{\string\newlabel{#1}{{#2}{\thepage}}}}}
\makeatother

\usepackage{bbm}
\crefname{thm}{\text{Theorem}}{\text{Theorems}}
\crefname{assm}{\text{Assumption}}{\text{Assumptions}}
\crefname{defn}{\text{Definition}}{\text{Definitions}}
\crefname{prop}{\text{Proposition}}{\text{Propositions}}
\crefname{cor}{\text{Corollary}}{\text{Corollaries}}
\crefname{lemma}{\text{Lemma}}{\text{Lemmas}}
\crefname{algorithm}{\text{Program}}{\text{Programs}}

\newcommand{\bigtheta}{\Theta}
\newcommand{\bigvtheta}{\Theta}

\newcommand{\trsp}{\top}
\newcommand{\distto}{\xrar{\mathrm{d}}}
\newcommand{\asto}{\xrar{\mathrm{a.s.}}}
\newcommand{\probto}{\xrar{\mathrm{p}}}
\newcommand{\disteq}{\overset{\mathrm{d}}{=}}

\newcommand{\cdc}{\mathfrak{c}}

\newcommand{\Ss}{\mathcal{S}}

\newcommand{\defeq}{\mathbin{\overset{\mathrm{def}}{=}}}

\newcommand{\Atype}{\mathsf{A}}
\newcommand{\Gtype}{\mathsf{G}}
\newcommand{\Htype}{\mathsf{H}}

\newcommand{\Jac}[2]{\f{\dd #1}{\dd #2}}

\renewcommand{\cite}{\citep}

\newcommand{\Vt}[1]{\mathrm{V}_{#1}}

\newcommand{\netsor}{{$\textsc{Netsor}$}}
\newcommand{\netsort}{{$\textsc{Netsor}\trsp$}}
\newcommand{\netsorplus}{{$\textsc{Netsor}^+$}}
\newcommand{\netsortplus}{{$\textsc{Netsor}\trsp^+$}}
\newcommand{\netsormin}{{$\textsc{Netsor}^-$}}

\newcommand{\Sigmain}{\Sigma^{\mathrm{in}}}
\newcommand{\muin}{\mu^{\mathrm{in}}}
\newcommand{\tSigma}{\Sigma}
\newcommand{\tmu}{\mu}

\newcommand{\loss}{\mathcal{L}}

\newtcolorbox[auto counter,crefname={Box}{Boxes}]{pabox}[2][]{%
title=Box~\thetcbcounter $\quad$ #2, label={#1}}

\newtcolorbox[auto counter,crefname={Box}{Boxes}]{floatpabox}[2][]{%
title=Box~\thetcbcounter $\quad$ #2, label={#1}}

\makeatletter
\newcommand{\altqedhere}{%
  \ifmeasuring@\else\sbox0{\popQED}\fi
  \tag*{\qedsymbol}%
}

\makeatletter
\let\orgdescriptionlabel\descriptionlabel
\newcommand*{\@restrictlabeltext}[1]{#1\protected@edef\@currentlabel{#1}}
\newcommand*{\nolabel}[1]{#1}%
\renewcommand*{\descriptionlabel}[1]{%
  \let\orglabel\label
  \let\label\@gobble
  \let\orig@hfil\hfil
  \def\hfil{}%
  \let\nolabel\@gobble
  \let\restrictlabeltext\@firstofone
  \phantomsection
  \protected@edef\@currentlabel{#1}%
  \let\hfil\orig@hfil
  \let\label\orglabel
  \let\restrictlabeltext\@restrictlabeltext
  \orgdescriptionlabel{#1}%
}
\makeatother

\title{Tensor Programs II:\\
Neural Tangent Kernel for Any Architecture}

\author{%
  Greg Yang\\
  Microsoft Research AI\\
  \texttt{gregyang@microsoft.com} \\
}
\newcommand{\NTK}{\Theta}
\begin{document}

\newcommand{\repo}{\url{https://github.com/thegregyang/NTK4A}}

\maketitle

\begin{abstract}

    We prove that a randomly initialized neural network of \emph{any architecture} has its Tangent Kernel (NTK) converge to a deterministic limit, as the network widths tend to infinity.
    We demonstrate how to calculate this limit.
    In prior literature, the heuristic study of neural network gradients often assumes every weight matrix used in forward propagation is independent from its transpose used in backpropagation \citep{schoenholz_deep_2017}.
    This is known as the \emph{gradient independence assumption (GIA)}.
    We identify a commonly satisfied condition, which we call \emph{Simple GIA Check}, such that the NTK limit calculation based on GIA is correct.
    Conversely, when Simple GIA Check fails, we show GIA can result in wrong answers.
    Our material here presents the NTK results of \citet{yangScalingLimitsWide2019arXiv.org} in a friendly manner and showcases the \emph{tensor programs} technique for understanding wide neural networks.
    We provide reference implementations of infinite-width NTKs of recurrent neural network, transformer, and batch normalization at \repo{}.
\end{abstract}

\vspace{-1em}

\section{Introduction}
\label{sec:Introduction}

\vspace{-0.5em}

\citet{jacot_neural_2018} showed that, in the limit of large width, a neural network undergoing training by gradient descent evolves like a linear model.
Their argument proceeds in two steps:
\begin{description}
    \item[\textsc{ntkInit}\label{NTKInit}]
        If $f(x; \theta)$ is the neural network (with parameters $\theta$ and input $x$), then we can define a kernel called \emph{Neural Tangent Kernel} by
        \vspace{-.5em}
         \begin{align*}
             \NTK(x, \bar x) \defeq \la \nabla_\theta f(x; \theta), \nabla_\theta f(\bar x; \theta) \ra,\quad\text{for any inputs $x, \bar x$.}
         \end{align*}
        \citet{jacot_neural_2018} showed that, if the parameters $\theta$ are appropriately randomized, then $\Theta$ converges to a deterministic kernel $\mathring \Theta$ as the widths of $f$ grow to infinity.
    \item[\textsc{ntkTrain}\label{NTKTrain}]
        In the limit of large width, the NTK in the course of gradient descent stays constant, and remarkably, the network evolves like a linear model under kernel gradient descent with this limiting NTK $\mathring \NTK$.
\end{description}

With recent experimental validation \citep{lee_wide_2019}, NTK promises to shed light on the training and generalization properties of overparametrized neural networks.
Yet, it's not clear whether NTK continues to be valid for modern deep learning, such as Faster R-CNN in image segmentation \citep{ren_faster_2015}, transformer in machine translation \citep{vaswani_attention_2017}, or generative adversarial networks for distribution learning \citep{goodfellow_generative_2014}.
In particular, we ask
\begin{equation*}
    \text{
    \it 
    Does every modern neural network have an infinite-width NTK? Can we compute it?
    }
\end{equation*}
\textbf{Our contributions.\ }
In this paper, we show that the NTK for any randomly initialized neural network of \emph{standard architecture}%
\footnote{In this work, architecture refers to the network topology along with the ratios of widths of hidden layers.}
converges almost surely to a deterministic limit, as the network widths%
\footnote{In fully-connected network, width is the number of neurons in a layer.
In a convolutional network, width is the number of channels in a layer.}
tend to infinity, and we show how to exactly compute this limit, i.e. we generalize \ref{NTKInit} to standard architectures.
\emph{By \textbf{standard architecture} we mean any architecture that is some composition of multilayer perceptrons (MLPs), recurrent neural networks (RNNs) (e.g., Long-Short Term Memory (LSTM) \cite{hochreiter_long_1997} or Gated Recurrent Unit (GRU) \cite{cho_learning_2014}), skip connections \mbox{\cite{he_deep_2016,huang_densely_2016}},
convolutions
\cite{fukushima_cognitron:_1975,fukushima_neocognitron:_1982,rumelhart_learning_1985,lecun_gradient-based_1998,lecun_object_1999} or graph convolutions \cite{
    bruna_spectral_2013,
    henaffDeepConvolutionalNetworks2015arXiv.org,
    duvenaudConvolutionalNetworksGraphs2015NeuralInformationProcessingSystems,
    liGatedGraphSequence2015arXiv.org,
    defferrardConvolutionalNeuralNetworks2016arXiv.org,
    kipfSemiSupervisedClassificationGraph2016arXiv.org},
pooling \cite{lecun_gradient-based_1998,lecun_object_1999}, batch normalization \cite{ioffe_batch_2015},
layer normalization \cite{ba_layer_2016} and/or attention \cite{bahdanau_neural_2014,vaswani_attention_2017}.}
More generally, our result applies to any architecture whose forward and backpropagation can be expressed via nonlinearities and matrix multiplication (\cref{defn:simpleNetsorT}).

\begin{wrapfigure}{r}{0.4\textwidth}
    \begin{center}
        \includegraphics[width=0.4\textwidth]{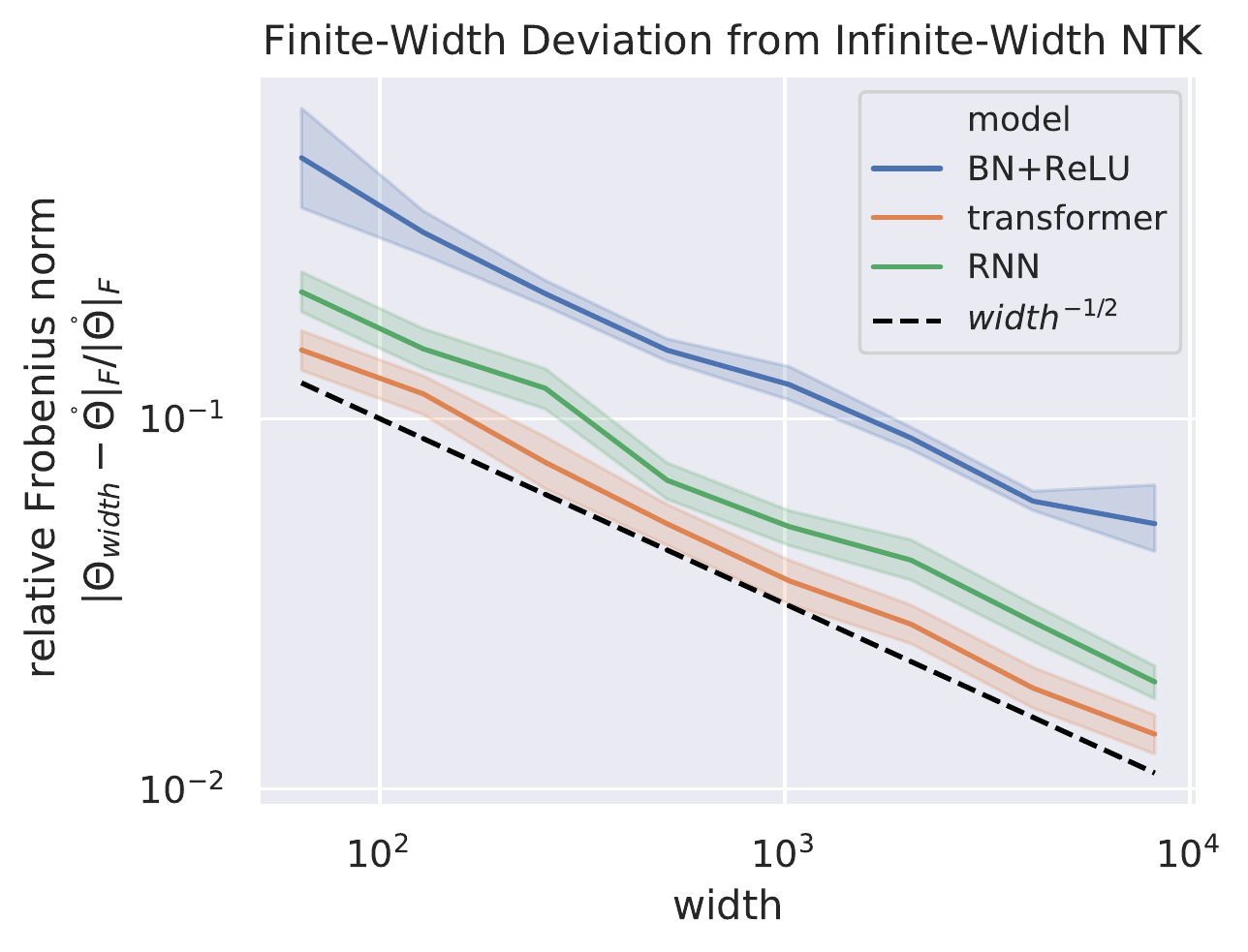}
    \end{center}
    \vspace{-1em}
\end{wrapfigure}
In the process, we identify a commonly satisfied condition (Simple GIA Check, \cref{assm:simpleGIACheck}) that rigorously justifies what is known as the \emph{Gradient Independence Assumption (GIA)} \citep{yang_mean_2017}.
This is the heuristic, used in calculating the statistics of neural network gradients at initialization, that $W^\trsp$ in backpropagation is independent from $W$ in forwardpropagation.
However, without \cref{assm:simpleGIACheck}, calculation based on GIA can be incorrect (\cref{sec:GIAbreaks}).

We give concrete algorithms to compute the infinite-width NTK for batchnorm-ReLU MLP, transformer, and RNN (\cref{sec:exampleNTKComputation}) and verify they agree with simulations.
In the plot on the right, we have computed the deviation of empirical NTKs $\NTK_{width}$ over widths $2^6, \ldots, 2^{13}$ from the corresponding limits $\mathring\NTK$.
The shade represents 95\% confidence interval of the mean over 100 seeds.

\paragraph{The \emph{Tensor Programs} Series}
This paper is the second in the \emph{Tensor Programs} series, following \citet{yang2019wide}.
Here we show \ref{NTKInit} holds for any standard architecture, which motivates \netsort{}, an extension of the tensor program language \netsor{} in \citet{yang2019wide} by matrix transposes.
Whereas \netsor{} can only express the forward propagation of a neural network, \netsort{} can also express its backpropagation.
This allows us to reason about the network's gradients in the infinite-width limit (and hence the NTK), which \netsor{} cannot do.
In a future paper, we will also generalize \ref{NTKTrain} for any standard architecture, which requires an even more expressive extension of \netsort{}.
The results in this paper supercede all results in \citet{yangScalingLimitsWide2019arXiv.org} regarding NTK.

Our results here imply a \emph{universal Neural Network-Tangent Kernel correspondence}.
It opens a way toward studying the inductive biases of a neural network of any architecture trained under SGD.
We hope this can enable theoretical understanding to catch up to practice even as neural networks manifest in increasingly many varied architectures in modern deep learning.

\section{Background}

Given a parametrized function $f(x;\theta)$ with parameter $\theta$ and with scalar output, we can naively expand $f$ in $\theta$ around a base point $\theta_{0}$ 
\begin{align}
f(x;\theta)-f(x;\theta_{0}) & \approx\langle\nabla_{\theta}f(x;\theta_{0}),\theta-\theta_{0}\rangle
\label{eqn:linearization}
\end{align}
for any input $x$, where $\la, \ra$ denotes inner product.
The RHS is a linear model, where $\nabla_{\theta}f(-;\theta_{0})$ acts as a input featurizer, and $\theta-\theta_{0}$ acts as the weights. This is a good approximation as long as $\theta$ is not too far from $\theta_{0}$ --- in particular, if $f$ is a neural network and we train it for a short amount of time under gradient descent with a small learning rate. However, at face value, it seems $f$ can never change --- and learn --- much under such training. Why would such a naive linearization of $f$ be helpful?

Counterintuively, \citet{jacot_neural_2018} showed that, as the network widths tend to infinity, $f$ can in fact fit any data \emph{perfectly} while \cref{eqn:linearization} remains an accurate description of the training dynamics!
An explanatory intuition is that, when $\theta$ is high dimensional, even a small change in $\theta$ can cause a large change in $f$.

Let's be a bit more precise.
Consider the $L$-hidden-layer MLP $f(x; \theta)$ described below in \cref{eqn:NTKparam}, with width $n^l$ in layer $l$.
Then \citet{jacot_neural_2018} showed that the finite-width NTK $\NTK(x,\bar{x})\defeq\langle\nabla_{\theta}f(x;\theta),\nabla_{\theta}f(\bar{x};\theta)\rangle$ converges in probability 
\begin{equation}
\NTK\probto\mathring{\NTK}\quad\text{as \ensuremath{n^{1},\ldots,n^{L}\to\infty} in that sequence,}\label{eqn:jacotNTKConvergence}\tag{\ref{NTKInit}}
\end{equation}
for some deterministic $\mathring{\NTK}$ to be described below (\cref{eqn:NTKlimit}), over the randomness induced by randomly initializing the parameters like $\omega_{\alpha\beta}^{l},b_{\alpha}^{l}\sim\Gaus(0,1),\forall\alpha,\beta.$
This means that the inner product between every pair of features $\nabla_\theta f(x; \theta_0), \nabla_\theta f(\bar x; \theta_0)$ of \cref{eqn:linearization} converges, as widths tend to infinity, even though the parameters $\theta$ are random.

Now consider the evolution of the MLP $f_t$ with time $t$, trained under continuous time gradient descent with loss function $\mathcal L$.
Let the initial function $f_0$ be obtained by standard Gaussian random initialization as above.
Then \citet{jacot_neural_2018} showed that, in the large width limit, for any fixed training time $T$,
\begin{align}
    f_t \to \mathring f_t \quad\text{for all $t < T$,}\quad\text{where}\quad
    \mathring{f}_{0}=f_{0},\ \pd_{t}\mathring{f}_{t}=-\eta\mathring{\NTK}\cdot\nabla_{f}\loss(\mathring{f}_{t}).
    \label{eqn:finfty}
    \tag{\ref{NTKTrain}}
\end{align}

Thus, somehow the hopelessly complicated optimization trajectory of an MLP has reduced to a kernel gradient descent with a fixed kernel $\mathring{\Theta}$. For square loss $\loss$, this equation further simplifies to a linear differential equation, allowing one to solve for $\mathring{f}_{t}$ explicitly for all $t$: if labels are provided by a ground truth function $f^{*}$, then
\[
\mathring{f}_{t}-f^{*}=e^{-\eta t\mathring{\Theta}}(f_{0}-f^{*}).
\]
Because one can show $\mathring{\NTK}$ is in general a non-singular kernel, this equation implies that $f$ can fit any training data given it is wide enough \citep{jacot_neural_2018}.

Thus, the infinite-width NTK $\mathring{\NTK}$ reflects an \emph{implicit prior} induced by gradient descent and the choices of architecture and initialization scheme. For example, its spectrum informs us the kind of functions that can be learned quickly and generalize well \citep{yang2019finegrained}. \citet{jacot_neural_2018} gave us a way into the blackbox of MLPs, and this paper tries to fill the gap for modern architectures. Here we describe a general, rigorous way of computing the infinite-width NTK of a network. In a future paper of this series, we will also show \ref{eqn:finfty} holds for any architecture as well.
We hope our work here can enable theoretical analyses of state-of-the-art neural networks that contribute to the practice of modern machine learning.

\section{Related Works}

\label{sec:RelatedWorks}

A much older literature of Gaussian process (GP) behavior of wide neural networks also associates a kernel to each network (the NN-GP correspondence) \citep{neal_bayesian_1995,{williams_computing_1997,le_roux_continuous_2007,hazan_steps_2015,daniely_toward_2016},lee_deep_2018,matthews_gaussian_2018_arxiv,novak_bayesian_2018}.
While the NTK can be thought of as characterizing the behavior of training the full network under gradient descent, the infinite-width GP of a network characterizes the same when training only the last layer.

After \citet{jacot_neural_2018} invented NTK, our original paper \citep{yangScalingLimitsWide2019arXiv.org} proved the architectural universality of NTK and NN-GP.
However, the results were written densely and in heavy programming language notation.
\citet{yang2019wide} simplified the writing and generalized the results for GP.
We do the same here for the NTK results.

After \citet{yangScalingLimitsWide2019arXiv.org}, several works dove into specific kinds of NTKs, such as convolutional \citep{arora_exact_2019}, graph \citep{du2019graph}, hypernetworks \citep{littwin2020optimization}, RNNs \citep{alemohammad2020recurrent}, attention \citep{hron2020infinite}, or NTK with orthogonal initialization \citep{huang2020neural}.
Other works studied the higher order terms in the Taylor expansion \citep{huang2019dynamics,dyer2019asymptotics}, ensembled NTK \citep{littwin2020collegial}, or finite width corrections \citep{hanin2019finite,littwin2020residual}.

Closely related is the signal propagation literature, which tries to understand how to prevent pathological behaviors in randomly initialized neural networks when they are deep \citep{poole_exponential_2016,schoenholz_deep_2017,yang_mean_2017,yangVarianceVariation,hanin_which_2018,hanin_how_2018,chen_dynamical_2018,yang_mean_2019,pennington_resurrecting_2017,hayou_selection_2018,philipp_nonlinearity_2018}.
The investigation of \emph{forward signal propagation} corresponds to studying the infinite-depth limit of the associated Gaussian process, and the investigation of \emph{backward signal propagation} corresponds to studying the infinite-depth limit of NTK.

Neural tangent kernel solved an age-old question of ``how does training of neural network work so well despite being highly nonconvex?'' \citep{jacot_neural_2018,allen-zhu_convergence_2018,allen-zhu_convergence_2018-1,allen-zhu_learning_2018,du_gradient_2018,zou_stochastic_2018}.
This in turn has been used for studying convergence questions in deep reinforcement learning \citep{achiam_towards_2019,cai2019neural}.
The spectrum of NTK has been analyzed to provide finer-grained answers to these problems~\citep{yang2019finegrained,basriConvergenceRateNeural2019arXiv.org,ghorbaniLinearizedTwolayersNeural2019arXiv.org}.

Compared to neural networks, kernel regression with the corresponding NTKs work better in the low data regime \citep{arora2019harnessing}, consistent with classical observations about kernel methods and previous works on NNGPs \citep{lee_deep_2018,novak_bayesian_2018}.
This can be valuable in important settings such as medical data that need to make decisions based on only a few data points.

\section{Warmup: Neural Tangent Kernel for a Multi-Layer Perceptron}

We first demonstrate the intuitions of our framework by redoing the MLP NTK limit computation.
Consider the MLP $f(\xi; \theta) = W^{L+1} x^L(\xi)$ with input $\xi \in \R^{n^0}$ and output dimension $n^{L+1} = 1$, where we recursively define, for $l = 2, \ldots , L$,
\begin{align*}
    h^{l}(\xi)=W^{l}x^{l-1}(\xi)+b^{l}\in\R^{n^{l}},\quad x^{l}(\xi)=\phi(h^{l}(\xi)),\quad h^{1}(\xi)=W^{1}\xi+b^{1}\in\R^{n^{1}}
    \numberthis\label{eqn:NTKparam}
\end{align*}
in which each $W^l$ is factorized as $W^l = \f 1 {\sqrt {n^{l-1}}} \omega^l$, and the MLP's parameters are $\theta = \{\omega^l \in \R^{n^l \times n^{l-1}}\}_{l=1}^{L+1}\cup \{b^l \in \R^{n^l}\}_{l=1}^L$.
This style of parametrization of weight matrices is known as the \emph{NTK parametrization.}
We shall sample $\omega_{\alpha\beta}^{l},b_{\alpha}^{l}\sim\Gaus(0,1),\forall\alpha,\beta.$
\citet{jacot_neural_2018}'s argument for \ref{eqn:jacotNTKConvergence} is inductive in the depth of the MLP, which would run into difficulty generalizing to other architectures with weight sharing, like RNNs.
Here we show a different technique based on decomposing the NTK into an explicit sum of products of terms whose limits we can evaluate.

\subsection{Decomposing NTK}

For simplicity, write $f(\xi)=f(\xi;\theta)$ and $\nabla_{p}f(\xi)$ will
denote gradient of the output $f(\xi)$ in some quantity $p$, given
$\xi$ and $\theta$. In the MLP (\cref{eqn:NTKparam})
above, we can decompose the NTK into contributions from weights and
biases: for inputs $\xi,\bar{\xi}\in\R^{n^{0}}$ (possibly $\xi = \bar \xi$),
\begin{equation}
\NTK(\xi,\bar{\xi})=\langle \nabla_\theta f(\xi), \nabla_\theta f(\bar \xi) \rangle = \sum_{l=1}^{L+1}\left\langle \nabla_{\omega^{l}}f(\xi),\nabla_{\omega^{l}}f(\bar{\xi})\right\rangle +\sum_{l=1}^{L}\left\langle \nabla_{b^{l}}f(\xi),\nabla_{b^{l}}f(\bar{\xi})\right\rangle, \label{eq:MLPNTKDecomp}
\end{equation}
where $\la, \ra$ denotes (trace) inner product.
To see this quantity converges as widths $n^{1},\ldots,n^{L}\to\infty$,
it suffices to show that each summand converges.
First note the $n^{l}\times n^{l-1}$
matrix $\nabla_{\omega^{l}}f(\xi)$ is the product of the $n^{l}\times1$
vector $\frac{1}{\sqrt{n^{l-1}}}\nabla_{h^{l}}f(\xi)$ and the $1\times n^{l-1}$
vector $x^{l-1}(\xi)^{\trsp}$, by chain rule.
Abbreviate $\bullet = \bullet(\xi), \bar \bullet = \bullet(\bar \xi)$ for different vectors $\bullet \in \{h^l, x^l\}_l$.
Set $dh^{l}=\sqrt{n^{l}}\nabla_{h^{l}}f(\xi)$ and $d\bar{h}^{l}=\sqrt{n^{l}}\nabla_{\bar{h}^{l}}f(\bar{\xi})$.
Then we have $\nabla_{\omega^{l}}f(\bar{\xi})=\f 1{\sqrt{n^{l}n^{l-1}}}d\bar{h}^{l}\bar{x}^{l-1\trsp}.$
Using the cyclic property of the trace inner product in the right equality,
\begin{align}
\left\langle \nabla_{\omega^{l}}f(\xi),\nabla_{\omega^{l}}f(\bar{\xi})\right\rangle  & =\f 1{n^{l}n^{l-1}}\left\langle dh^{l}x^{l-1\trsp},d\bar{h}^{l}\bar{x}^{l-1\trsp}\right\rangle
    =\left(\frac{dh^{l\trsp}dh^{l}}{n^{l}}\right)\left(\frac{x^{l-1\trsp}\bar{x}^{l-1}}{n^{l-1}}\right).\label{eq:MLPNTKSimplify}
\end{align}
In the rest of the section we seek to understand the two terms in this product in an intuitive way.
The main ingredients in our argument are a central limit heuristic (i.e. the sum of many roughly independent random variables looks like a Gaussian) and gradient independence assumption.

\subsection{Limits of Forward Quantities \texorpdfstring{$x^{l\trsp} \bar x^l/n^l$}{}}
\label{sec:limitForwardMLP}

By the randomness of the initial weight matrices and inductive applications of central limit arguments, $(x_{\alpha}^{l},\bar{x}_{\alpha}^{l})$
is intuitively correlated but roughly iid across $\alpha \in [n^l]$ \citep{poole_exponential_2016,schoenholz_deep_2017},
so
\begin{equation}
\frac{x^{l\trsp}\bar{x}^{l}}{n^{l}}\to C^{l}(\xi,\bar{\xi}),\label{eq:MLPactKernel}
\end{equation}
for some deterministic scalar $C^{l}(\xi,\bar{\xi})$.
Unpacking this a bit: for each $\alpha$, the coordinate $(W^{l}x^{l-1})_{\alpha}=\sum_{\beta=1}^{n}W_{\alpha\beta}^{l}x_{\beta}^{l-1}$ is a sum of a large number $n$ of roughly iid random variables $W_{\alpha\beta}^{l}x_{\beta}^{l-1}$.
Its variance is $\EV(W^{l}x^{l-1})_{\alpha}^{2}=\EV(\sum_{\beta=1}^{n}W_{\alpha\beta}^{l}x_{\beta}^{l-1})^{2}=\sum_{\beta=1}^{n}\EV(W_{\alpha\beta}^{l})^{2}\EV(x_{\beta}^{l-1})^{2}=\|x^{l-1}\|^{2}/n^{l-1}\approx C^{l-1}(\xi,\xi)$.
So by a central limit argument, $(W^{l}x^{l-1})_{\alpha}$ should look like $\Gaus(0,C^{l-1}(\xi,\xi))$. Similarly, $(W^{l}\bar{x}^{l-1})_{\alpha}$ should be roughly $\Gaus(0,C^{l-1}(\bar{\xi},\bar{\xi}))$ and the pair $((W^{l}x^{l-1})_{\alpha},(W^{l}\bar{x}^{l-1})_{\alpha})$ should be jointly Gaussian with covariance $C^{l-1}(\xi,\bar{\xi})$.
Then the pair $(x_{\alpha}^{l},\bar{x}_{\alpha}^{l})$
should be distributed like $(\phi(\zeta),\phi(\bar{\zeta}))$, and
$C^{l}$ satisfies the following recursion (here the $+1$ comes from the bias $b^l\sim \Gaus(0,1)$)
\begin{equation}
C^{l}(\xi,\bar{\xi})=\EV\phi(\zeta)\phi(\bar{\zeta}),\quad\text{where}\quad
(\zeta,\bar{\zeta})\sim\Gaus\left(0,\left(\begin{array}{cc}
    C^{l-1}(\xi,\xi) & C^{l-1}(\xi,\bar{\xi})\\
    C^{l-1}(\bar{\xi},\xi) & C^{l-1}(\bar{\xi},\bar{\xi})
    \end{array}\right)+1\right).
\label{eq:MLPActKernelRec}
\end{equation}

\subsection{Limits of Backward Quantities \texorpdfstring{$dh^{l\trsp} d\bar h^l/n^l$}{}}

For simplicity, assume $n^{1}=\cdots=n^{L}$.
Then like $h^l$, we can also expand $dx^l_\alpha \defeq (W^{l+1\trsp } dh^{l+1})_\alpha = (W^{l+1\trsp } (dx^{l+1} \odot \phi'(h^{l+1}))_\alpha
= \sum_\beta W^{l+1}_{\beta\alpha} dx^{l+1}_\beta \phi'(h^{l+1}_\beta)$.
We might hope to say that each term of this sum is roughly independent so we can apply a central limit heuristic, but $h^{l+1}_\beta$ actually depends on $W^{l+1}_{\beta \gamma}$ for all $\gamma$.
Interestingly, the signal propagation literature~\citep{schoenholz_deep_2017,yang_mean_2017,xiao_dynamical_2018,yang_mean_2019} has found \emph{it's fine to ignore} such dependences: If we adopt the following
\begin{heur}[gradient independence assumption, or GIA \citep{schoenholz_deep_2017,yang_mean_2017}]
    For any matrix $W$, we assume $W^{\trsp}$ used in backprop is independent from $W$ used in the forward pass.
\end{heur}
then the resulting calculation will still agree with simulation when $n^1, \ldots, n^L \gg 1$.
With this assumption, we can then proceed as in \cref{sec:limitForwardMLP} and argue $dx^l_\alpha$ is roughly distributed as $\Gaus(0, \|dh^{l+1}\|^2 / n^{l+1})$ and iid across $\alpha \in [n^l]$.
Likewise, we argue the pair $(dx_{\alpha}^{l},d\bar{x}_{\alpha}^{l})\defeq((W^{l+1\trsp}dh^{l+1})_{\alpha},(W^{l+1\trsp}d\bar{h}^{l+1})_{\alpha})$ is jointly Gaussian with zero mean and covariance $\|dh^{l+1\trsp} d \bar h^{l+1}\|^2/n^{l+1}$, and is iid across $\alpha$.
Since $(h^{l}_\alpha, \bar h^{l}_\alpha)$ is also roughly iid across $\alpha$, we expect $(dh_{\alpha}^{l},d\bar{h}_{\alpha}^{l})=(dx_{\alpha}^{l}\phi'(h_{\alpha}^{l}),d\bar{x}_{\alpha}^{l}\phi'(\bar{h}_{\alpha}^{l}))$ to be so as well, and
\begin{equation}
\frac{dh^{l\trsp}d\bar h^{l}}{n^{l}}\to D^{l}(\xi,\bar{\xi}),\label{eq:MLPgradKernel}
\end{equation}
for some deterministic scalar $D^{l}(\xi,\bar{\xi})$.
Combining our calculations here, we see $D^l$ satisfies the recurrence
\begin{align*}
D^{l}(\xi,\bar{\xi})&=\EV\eta\bar{\eta}\EV\phi'(\zeta)\phi'(\bar{\zeta})=D^{l+1}(\xi,\bar{\xi})\EV\phi'(\zeta)\phi'(\bar{\zeta})\numberthis\label{eq:MLPGradKernelRec}\\
\text{where}\quad (\eta,\bar{\eta})&\sim\Gaus\left(0,\left(\begin{array}{cc}
    D^{l+1}(\xi,\xi) & D^{l+1}(\xi,\bar{\xi})\\
    D^{l+1}(\bar{\xi},\xi) & D^{l+1}(\bar{\xi},\bar{\xi})
    \end{array}\right)\right),\\
(\zeta,\bar{\zeta})&\sim\Gaus\left(0,\left(\begin{array}{cc}
    C^{l}(\xi,\xi) & C^{l}(\xi,\bar{\xi})\\
    C^{l}(\bar{\xi},\xi) & C^{l}(\bar{\xi},\bar{\xi})
    \end{array}\right)+1\right)
\end{align*}
Together with \cref{eq:MLPactKernel} and \cref{eq:MLPgradKernel},
we have
\[
\left\langle \nabla_{\omega^{l}}f(\xi),\nabla_{\omega^{l}}f(\bar{\xi})\right\rangle \to C^{l-1}(\xi,\bar{\xi})D^{l}(\xi,\bar{\xi}),\quad\forall l\in[2,L].
\]
Similarly, because $\nabla_{b^{l}}f(\xi)=\nabla_{h^{l}}f(\xi)=dh^{l}/\sqrt{n^{l}}$,
we have
\[
\left\langle \nabla_{b^{l}}f(\xi),\nabla_{b^{l}}f(\bar{\xi})\right\rangle \to D^{l}(\xi,\bar{\xi}),\quad\forall l\in[2,L].
\]
So the NTK should converge like
\begin{equation}
\NTK(\xi,\bar{\xi})\to\sum_{l=1}^{L+1}C^{l-1}(\xi,\bar{\xi})D^{l}(\xi,\bar{\xi})+\sum_{l=1}^{L}D^{l}(\xi,\bar{\xi}).\label{eq:MLPNTKDecompLimit}
\end{equation}
Together with \cref{eq:MLPActKernelRec} and \cref{eq:MLPGradKernelRec},
this in fact recovers the NTK limit formula in \citet{jacot_neural_2018}.

\section{NTK for Any Architecture? The Issues and the Proposal}

The method presented in the last section for computing the MLP NTK already seem easier than that of \citet{jacot_neural_2018} to generalize to other architectures, but several thorny issues still remain.

\paragraph{Q1: Can we meaningfully generalize the NTK decomposition in \cref{eq:MLPNTKDecomp}?}
For example, for an MLP with weights tied across layers (i.e. $W^{l}=W^{l+1}$,
for all $l=1,\ldots,L-1$), we can generalize \cref{eq:MLPNTKDecomp} into a similar decomposition, but how do we know terms like $
\frac{dh^{l\trsp}d\bar h^{l}}{n^{l}}$ will converge or won't blow up to $\infty$ due to the extra correlations from weight tying?

\paragraph{Q2: Can we continue to assume gradient independence?}

GIA significantly simplified our calculation above for the MLP.
However, at a first glance it would still seem absurd to assume $W^{\trsp}$ is independent from $W$.
Now, for example, suppose we tie the weights across layers in the MLP.
The additional correlations then make GIA even more questionable.
Can we still assume GIA?

\paragraph{Q3: Can we uniformly handle the complexity of modern neural networks?}

Standard architectures like CNN, RNN, GRU, LSTM, transformer, ResNet, etc contain a wide variety of gadgets, and \emph{a priori} it's not clear there's a systematic way of handling all of them at once.

The techniques in this paper yield the following answers:

\paragraph{A1: Yes.}
We can generalize \cref{eq:MLPNTKDecomp} to decompose the NTK into a sum of products of inner products of the form $h^\trsp \bar h /n$ with $h, \bar h \in \R^n$, and importantly, each such term will turn out to tend to a deterministic finite constant as $n \to \infty$, implying NTK converges as well.
See \cref{eq:generalNTKDecomp,eqn:NTKlimit}.

\paragraph{A2: Conditional Yes.}
It turns out, somewhat counterintuitively, whether GIA works doesn't depend on the hidden-to-hidden weight matrices (which GIA concerns) so much as the output layer weights.
The following is a general but easily checkable condition that implies GIA:
\begin{cond}[Simple GIA Check]\label{assm:simpleGIACheck}
    \mylabel{text:simpleGIACheck}{Simple GIA Check}
    The output layer (like $W^{L+1}$ in the MLP example above) is sampled independently and with zero mean from all other parameters and is not used anywhere else in the interior of the network%
    \footnote{i.e.\ if the output weight is $v$ and the output is $v^\trsp x$, then $x$ does not depend on $v$.}.
\end{cond}
At a very high level, \cref{assm:simpleGIACheck} implies GIA because any weight matrix $W$ can only interact with its transpose $W^\trsp$ via a path that goes through the last layer weights.
If these weights are sampled independently and with zero mean, then such interactions are zeroed out as well.
See \cref{eqn:GIAIntuition} for a concrete explanation.
In \cref{sec:GIAbreaks}, we also show a counterexample where \cref{assm:simpleGIACheck} is violated and GIA doesn't work%
\footnote{Note that GIA means we can assume the backward weights are independent
from the forward weights but multiple usages of backward weights (e.g.
in an RNN backprop) are not assumed to be independent from each other.}.
For a more general condition guaranteeing GIA, see \cref{defn:BPlike}.

\paragraph{A3: Yes.}
We introduce a simple and general language, \netsort{} (extending \netsor{} from \citet{yang2019wide}), expressing compositions of matrix multiplication and nonlinearity application, such that if an NN satisfies \cref{assm:simpleGIACheck} and one can write down its forward and backward computations in \netsort{} (as can be done for standard architectures), then its NTK provably converges under mild regularity conditions (\cref{cor:NTKConvergence}).
This \netsort{} program can allow one to mechanistically compute the infinite-width NTK by recursively applying the Master Theorem (\cref{thm:netsorTMasterTheoremSimple}).

\section{Strategy for Computing the Infinite-Width NTK}

For general architectures, we can in fact compute the NTK with an
overall strategy very similar to \cref{eq:MLPNTKDecomp} and \cref{eq:MLPNTKDecompLimit}. 

\subsection{The Canonical Decomposition}

Consider a neural network%
\footnote{formally, we consider any neural network whose computation can be expressed in \netsort{} (\cref{defn:simpleNetsorT}); however, in this section, an intuitive understanding of ``neural network'' is enough.}
$f(\xi)$ with input $\xi \in \R^d$, scalar output, and with weights
$W$ and biases $b$ such that any weight $W\in\R^{n\times m}$ is
always used in the computation of $f(\xi)$ in the form $y(\xi)=Wz(\xi)$,
for possibly many different vectors $y(\xi)\in\R^{n},z(\xi)\in\R^{m}$.
For example, in the MLP example above, $W$ would be $W^{l}$ for
some $l$, and $y(\xi)=h^{l}(\xi),z(\xi)=x^{l-1}(\xi)$.
If the MLP weights are tied across layers with $W = W^2 = \cdots = W^L$, then $(y, z) \in \{(h^2, x^{1}), \ldots, (h^L, x^{L-1})\}$.

Suppose that we adopt the NTK parametrization where $W$ is factored
as $W=\f 1{\sqrt{m}}\omega$ for $\omega\in\R^{n\times m}$, and $\omega$, instead of $W$, is trained.
Then
the NTK $\NTK$ of $f$ is a sum 
\begin{equation}
\NTK(\xi,\bar{\xi})=\sum_{\omega}\left\langle \nabla_{\omega}f(\xi),\nabla_{\omega}f(\bar{\xi})\right\rangle +\sum_{b}\left\langle \nabla_{b}f(\xi),\nabla_{b}f(\bar{\xi})\right\rangle \label{eq:generalNTKDecomp}
\end{equation}
over biases $b$ and factorized weights $\omega$. 
In the MLP example with tied-weights $W = W^2 = \cdots = W^L \in \R^{n \times n}$ and $W = \f 1 {\sqrt n} \omega$, we can write $\nabla_\omega f(\xi) = \f 1 {n} \sum_{l=1}^{L-1}dh^{l+1}\:x^{l\trsp}$, and
\begin{align*}
    \left\langle \nabla_{\omega}f(\xi),\nabla_{\omega}f(\bar{\xi})\right\rangle 	&=\f 1{n^{2}}\left\langle \sum_{l=1}^{L-1}dh^{l+1}\:x^{l\trsp},\sum_{\ell=1}^{L-1}d\bar{h}^{\ell+1}\:\bar{x}^{\ell\trsp}\right\rangle \\
	&=\f 1{n^{2}}\sum_{l,\ell=1}^{L-1}\left\langle dh^{l+1}\:x^{l\trsp},d\bar{h}^{\ell+1}\:\bar{x}^{\ell\trsp}\right\rangle =\sum_{l,\ell=1}^{L-1}\frac{dh^{l+1\trsp}d\bar{h}^{\ell+1}}{n}\f{x^{l\trsp}\bar{x}^{\ell}}n.
\end{align*}
In the general case, consider any two
inputs $\xi,\bar{\xi}$ to $f$ (possibly equal).
If we abbreviate $\bar{y}=y(\bar{\xi}),\bar{z}=z(\bar{\xi}),dy=\sqrt{n}\nabla_{y}f(\xi),d\bar{y}=\sqrt{n}\nabla_{\bar{y}}f(\bar{\xi})$,
then we can express the contribution of $\nabla_{\omega}f$ to the NTK $\NTK$
of $f$ as
\begin{align}
\left\langle \nabla_{\omega}f(\xi),\nabla_{\omega}f(\bar{\xi})\right\rangle
 & =\f 1m\left\langle \nabla_{W}f(\xi),\nabla_{W}f(\bar{\xi})\right\rangle
 =\f 1{mn}\left\langle \sum_{y,z}dy\:z^{\trsp},\sum_{\bar{y},\bar{z}}d\bar{y\:}\bar{z}^{\trsp}\right\rangle \nonumber \\
 & =\f 1{mn}\sum_{\substack{y,z,\bar{y},\bar{z}}}\left\langle dy\:z^{\trsp},d\bar{y}\:\bar{z}^{\trsp}\right\rangle
 =\sum_{\substack{y,z,\bar{y},\bar{z}}}\frac{dy^{\trsp}d\bar{y}}{n}\f{z^{\trsp}\bar{z}}m\label{eqn:NTKsimplify}
\end{align}
where the sum is over all matrix multiplication of the
form $y=Wz$ (resp.\ $\bar y = W \bar z$) used in the computation of $f(\xi)$ (resp.\ $f(\bar \xi)$).
Notice how \cref{eq:generalNTKDecomp} generalizes \cref{eq:MLPNTKDecomp},
and \cref{eq:MLPNTKSimplify} is just \cref{eqn:NTKsimplify} where
the sum is over the singleton sets $\left\{ h^{l},x^{l-1}\right\} $
and $\left\{ \bar{h}^{l},\bar{x}^{l-1}\right\} $.

We will show below (\cref{thm:netsorTMasterTheoremSimple})
that $\frac{dy^{\trsp}d\bar{y}}{n}$ and $\f{z^{\trsp}\bar{z}}m$
both converge almost surely to some deterministic limits $D^{y,\bar{y}}(\xi,\bar{\xi})$
and $C^{z,\bar{z}}(\xi,\bar{\xi})$ if the factored weights and biases
$\omega,b$ are drawn from standard Gaussians (i.e.\ in the \emph{NTK
parametrization}), as widths tend to infinity.
Similarly, we will also show the convergence of $\nabla_{b}f(\xi)^{\trsp}\nabla_{b}f(\bar{\xi})$
for any bias $b$ of $f$ and compute its limiting value $D^{b}(\xi,\bar{\xi})$.
Then the limiting NTK is given by 
\begin{equation}
\mathring{\NTK}(\xi,\bar{\xi})=\sum_{\text{weight \ensuremath{W}}}\sum_{\substack{y,z:y=Wz\\\bar{y},\bar{z}:\bar{y}=W\bar{z}}}D^{y,\bar{y}}(\xi,\bar{\xi})C^{z,\bar{z}}(\xi,\bar{\xi})+\sum_{\text{bias \ensuremath{b}}}D^{b}(\xi,\bar{\xi}).\label{eqn:NTKlimit}
\end{equation}

\subsection{Intuitive Rules for Computing Intermediate Kernels \texorpdfstring{$C$ and $D$}{C and D}}

Here we present intuitive rules for computing $C$ and $D$, which would yield the NTK by \cref{eqn:NTKlimit}.
Their justifications will follow in the next section.
Consider the first forward and backward propagations of a neural network.
Assume for simplicity that the hidden layers all have the same width, denoted $n$, which tends to infinity.
Then under \cref{assm:simpleGIACheck}%
\footnote{or when the associated \netsort{} program is BP-like (\cref{defn:BPlike})},
the following is the key intuition for computing the kernels $C$ and $D$ for arbitrary architecture.

\begin{pabox}[keyIntuition]{Key Intuitions for Understanding a Wide Neural Network}
    When the width $n \gg 1$, every (pre-)activation vector $x\in\R^{n}$ has roughly iid coordinates distributed as some
    random variable denoted $Z^{x}$. The set of random variables $\{ Z^{x}\} _{x}$
    over $x \in \R^n$ in this computation is possibly correlated, as $\{ x_{\alpha}\} _{x}$ is possibly correlated for each
    $\alpha\in[n]$, but is roughly iid across $\alpha$.
    Thus, for any vectors $x, y \in \R^n$, as $n \to \infty$,%
    \[{x^\trsp y}/n \to \EV Z^x Z^y,\]
    which is the form of the limit (kernels $C$ and $D$) we want.
    We can use the following rules to compute $Z^{x}$ recursively.
    \begin{enumerate}
    \item \textbf{(Nonlin)} For any fixed (i.e. constant as $n \to\infty$) $k$ and $\phi:\R^{k}\to\R$, we have\footnote{here $\phi$ is applied coordinatewise to $x^1, \ldots, x^k$, i.e.\ $\phi(x^1, \ldots, x^k)_\alpha = \phi(x^1_\alpha, \ldots, x^k_\alpha)$}
        \[Z^{\phi(x^{1},\ldots,x^{k})}=\phi(Z^{x^{1}},\ldots,Z^{x^{k}}).\]
    \item \textbf{(MatMul)} For any set of $\R^n$ vectors $\mathcal{X}$ and a matrix $W\in\R^{n\times n}$
    with $W_{\alpha\beta}\sim\Gaus(0,\sigma_{W}^{2}/n)$, the set of random
    variables $\{ Z^{Wx}:x\in\mathcal{X}\} $ is jointly Gaussian
    with zero mean and covariance
    \[
    \Cov(Z^{Wx},Z^{W\bar{x}})=\sigma_{W}^{2}\EV Z^{x}Z^{\bar{x}},\quad
    \text{for any }x, \bar x \in \mathcal X.
    \]
    If $\mathcal{Y}$ is any set of $\R^n$ vectors and $\bar W \ne W$, then
    $\{ Z^{Wx}:x\in\mathcal{X}\} $
    is independent from $\{ Z^{\bar W y}:y\in\mathcal{Y}\} $.
    \end{enumerate}
\end{pabox}

\begin{remk}
    Rule 2 applies even if $W$ is correlated with vectors in $\mathcal{X}$,
    for example if $x=W\bar{x}$ or $x=W^{\trsp}\bar{x}$ for some $x,\bar{x}\in\mathcal{X}$.
\end{remk}

\begin{remk}
    In Rule 2, if we set $\bar W = W^\trsp$, then the rule implies
    $\left\{ Z^{W^\trsp y}:y\in\mathcal{Y}\right\} $ is independent from $\left\{ Z^{Wx}:x\in\mathcal{X}\right\} $.
    This is how we use GIA implied by \cref{assm:simpleGIACheck}.
\end{remk}

\begin{remk}
    To reason about the computation of a wide neural network on the input $\xi$ of fixed dimension, we apply the above rules to the first layer embedding $W\xi$ into $\R^n$, but not $\xi$ itself.
\end{remk}

These rules largely generalize our intuitive treatment of the MLP example above:
The ``iid coordinates'' intuition suggests the limits in \cref{eq:MLPgradKernel,eq:MLPactKernel}.
The recursive relations in \cref{eq:MLPGradKernelRec,eq:MLPActKernelRec} of $C$ and $D$ are then given by Rule 1 and 2.
Now let us examine the power of these rules by looking at more advanced weight sharing inside an RNN.

\subsubsection{Example: RNN}
\label{sec:RNNIntuition}

Consider the RNN with state $s^t$ at time $t$ evolving according to
\begin{equation}
s^{t}(\xi)=\phi(g^{t}(\xi)+u^{t}(\xi)+b),\quad g^{t}(\xi)=Ws^{t-1}(\xi),\quad u^{t}(\xi)=U \xi^{t}
\label{eqn:RNNevolution}
\end{equation}
with input sequence $\xi=\left\{ \xi^{1},\ldots,\xi^{t},\ldots, \xi^T \in\R^{d}\right\} $, nonlinearity $\phi$, weights $W\in\R^{n\times n},U\in\R^{n\times d},$ and bias $b\in\R^{n}$.
The RNN outputs $v^\trsp s^T(\xi)/\sqrt{n} \in \R$ for some output weights $v \in \R^n$ and the last state $s^T(\xi)$.
We shall sample $W_{\alpha\beta}\sim\Gaus(0,1/n),U_{\alpha\beta}\sim\Gaus(0,1/d),b_{\alpha}\sim\Gaus(0,1), v_\alpha \sim \Gaus(0, 1)$.
Then \cref{assm:simpleGIACheck} becomes true automatically, and we may use the rules in \cref{keyIntuition}.

As in \cref{eqn:NTKsimplify}, we shall consider a second input sequence
$\bar{\xi}=\{ \bar{\xi}^{1},\ldots,\bar{\xi}^{t},\ldots\in\R^{d}\} $, possibly with $\bar \xi = \xi$.
There are two weight matrices in this network, $W$ and $U$. For
$W$, the double sum in \cref{eqn:NTKsimplify} is over $\{ g^{t},s^{t-1}\} _{t}$
and $\{ \bar{g}^{t},\bar{s}^{t-1}\} _{t}$. Thus we seek
to calculate the limits of $\frac{s^{t\trsp}\bar{s}^{r}}{n}$ and
$\frac{dg^{t\trsp}d\bar{g}^{r}}{n}$ for all $t$ and $r$. Similarly,
for $U$, the double sum in \cref{eqn:NTKsimplify} is over $\left\{ u^{t},\xi^{t}\right\} _{t}$
and $\left\{ \bar{u}^{t},\bar{\xi}^{t}\right\} _{t}$.
Thus we also seek to calculate the limits of $\frac{du^{t\trsp}d\bar{u}^{r}}{n}$,
whereas we are already given $\frac{\xi^{t\trsp}\bar{\xi}^{r}}{d}$, which
is constant in $n$ for each $t$ and $r$.

\subparagraph*{Forward}

As width $n\to\infty$ (but input dimension $d$ fixed),
\cref{keyIntuition} says we can think of $g^{t},u^{t},s^{t},b$
as having iid coordinates distributed resp.\ as some random variables
$Z^{g^{t}},Z^{u^{t}},Z^{s^{t}},Z^{b}$. Of course, $Z^{b}=\Gaus\left(0,1\right)$
and $\{ Z^{u^{t}},Z^{\bar{u}^{t}}\} _{t}$ is jointly Gaussian
with mean zero and covariance $\Cov(Z^{u^{t}},Z^{\bar{u}^{r}})=\xi^{t\trsp}\bar{\xi}^{r}/d$.
By Rule 2, $\{ Z^{g^{t}},Z^{\bar{g}^{t}}\} _{t}$ is also
jointly Gaussian with mean zero, and it has covariance $\Cov(Z^{g^{t}},Z^{\bar{g}^{r}})=\EV Z^{s^{t-1}}Z^{\bar{s}^{r-1}}$.
Stringing them together gives us the following recursion
\begin{align*}
\EV Z^{s^{t}}Z^{\bar{s}^{r}}
    &=\EV\phi(Z^{g^{t}}+Z^{u^{t}}+Z^{b})\phi(Z^{\bar{g}^{r}}+Z^{\bar{u}^{r}}+Z^{b})=\EV\phi(\zeta_{1})\phi(\zeta_{2}),\\
\text{where}\quad
    &(\zeta_{1},\zeta_{2})\sim\Gaus\left(0,\EV\left(\begin{array}{cc}
    \left(Z^{s^{t-1}}\right)^{2} & Z^{s^{t-1}}Z^{\bar{s}^{r-1}}\\
    Z^{\bar{s}^{r-1}}Z^{s^{t-1}} & \left(Z^{\bar{s}^{r-1}}\right)^{2}
    \end{array}\right)+\frac{\xi^{t\trsp}\xi^{r}}{d}+1\right).
\end{align*}
This recursion yields the desired limit
\begin{equation}
    C^{s^t, \bar s^r}(\xi, \bar \xi) = \lim_{n\to\infty}\frac{s^{t\trsp}\bar{s}^{r}}{n} = \EV Z^{s^{t}}Z^{\bar{s}^{r}}.
\label{eqn:RNNActKer}
\end{equation}
\subparagraph*{Backward}

The backward equation is given by 
\begin{equation}
ds^{t-1}=W^{\trsp}dg^{t},\quad dg^{t}=du^t=\phi'\left(g^{t}+u^{t}+b\right)\odot ds^{t}.
\label{eqn:RNNbackprop}
\end{equation}
By \cref{keyIntuition}, we should think of $ds^{t}$ as having iid coordinates
distributed like some random variable $Z^{ds^{t}}$ which satisfy
\begin{align*}
\EV Z^{ds^{t}}Z^{d\bar{s}^{r}} & =\EV Z^{du^{t+1}}Z^{du^{r+1}}\\
 & =\EV\phi'(Z^{g^{t+1}}+Z^{u^{t+1}}+Z^{b})Z^{ds^{t+1}}\phi'(Z^{\bar{g}^{r+1}}+Z^{\bar{u}^{r+1}}+Z^{b})Z^{d\bar{s}^{r+1}}\\
 & =\EV Z^{ds^{t+1}}Z^{d\bar{s}^{r+1}}\EV\phi'(Z^{g^{t+1}}+Z^{u^{t+1}}+Z^{b})\phi'(Z^{\bar{g}^{r+1}}+Z^{\bar{u}^{r+1}}+Z^{b})\\
 & =\EV Z^{ds^{t+1}}Z^{d\bar{s}^{r+1}}\EV\phi'(\zeta_{1})\phi'(\zeta_{2}),
\end{align*}
where $(\zeta_{1},\zeta_{2})\sim\Gaus\left(0,\EV\left(\begin{array}{cc}
\left(Z^{s^{t}}\right)^{2} & Z^{s^{t}}Z^{\bar{s}^{r}}\\
Z^{\bar{s}^{r}}Z^{s^{t}} & \left(Z^{\bar{s}^{r}}\right)^{2}
\end{array}\right)+\frac{\xi^{t\trsp}\xi^{r}}{d}+1\right)$. This recursion yields the desired limit
\begin{align*}
&\phantomeq D^{s^{t},\bar s^{r}}(\xi, \bar \xi)
=\lim_{n\to\infty}\frac{ds^{t\trsp}d\bar{s}^{r}}{n}
=\EV Z^{ds^{t}}Z^{d\bar{s}^{r}}\\
&=D^{u^{t+1},\bar u^{r+1}}(\xi, \bar \xi)
=\lim_{n\to\infty}\frac{du^{t+1\trsp}d\bar{u}^{r+1}}{n}
=\EV Z^{du^{t+1}}Z^{du^{r+1}}.
\end{align*}
Combined with \cref{{eqn:NTKlimit},{eqn:RNNActKer}}, we can compute the infinite-width NTK.
See \cref{sec:simpleRNNCalc} also for generalization to RNN with average pooling.

\paragraph*{Other standard architectures}
follow a similar scheme; see \cref{sec:BackpropInNetsorT,sec:exampleNTKComputation}.

\subsection{GIA Makes or Breaks the Intuitive Rules of \texorpdfstring{\cref{keyIntuition}}{Box 6.1}}
\label{sec:GIAbreaks}

\paragraph{Without \cref{assm:simpleGIACheck}, rules of \cref{keyIntuition} may not work}

When the last layer outputs the average of the final embedding, \cref{assm:simpleGIACheck} doesn't hold anymore.
Let us see how this means we can't treat $W^{\trsp}$ as independent from $W$.
Suppose we have a 2-hidden-layer network
\[
x^{1}=W^{1}\xi+1,\quad h^{2}=W^{2}x^{1},\quad x^{2}=\phi(h^{2}),\quad y=1^{\trsp}x^{2}/n
\]
with $\phi(z)=z^{2}$ being the square function, $\xi=0\in\R^{d},y\in \R, x^{1},h^{2},x^{2}\in\R^{n},W^{1}\in\R^{n\times d},W^{2}\in\R^{n\times n},W_{\alpha\beta}^{1}\sim\Gaus(0,1/d),W_{\alpha\beta}^{2}\sim\Gaus(0,1/n)$.
If we set $dx^2 = n\pdf{y}{x^2}$, then backprop yields
\[
dx^{2}=1,\quad dh^{2}=2h^{2}\odot1=2h^{2},\quad dx^{1}=W^{2\trsp}dh^{2}=2W^{2\trsp}h^{2}=2W^{2\trsp}W^{2}x^{1}
\]
By Rule 2, $h^{2}$ should have coordinates distributed like $Z^{h^{2}}=\Gaus(0,1)$
and likewise $dh^{2}$ has coordinates distributed like $Z^{dh^{2}}=2Z^{h^{2}}=\Gaus(0,4)$.

If we assumed that $W^{2\trsp}$ is independent from $W^{2}$, then
this would imply $dx^{1}$ also has coordinates distributed like $\Gaus(0,4).$
But a simple calculation shows its mean cannot be 0 in reality:
\begin{align*}
\EV dx_{\alpha}^{1} & =2\EV\sum_{\beta,\gamma}W_{\beta\alpha}^{2}W_{\beta\gamma}^{2}x_{\gamma}^{1}
  =2\sum_{\beta}\EV(W_{\beta\alpha}^{2})^2 x_{\alpha}^{1}+2\sum_\beta \sum_{\gamma\ne\alpha}\EV W_{\beta\alpha}^{2}W_{\beta\gamma}^{2}x_{\gamma}^{1}
  =2\EV x_{\alpha}^{1}=2
\end{align*}
where the second sum vanishes because the terms $W_{\beta\alpha}^{2},W_{\beta\gamma}^{2}, x_{\gamma}^{1}$
in the product are independent, while in the first sum we have $\sum_{\beta}\EV(W_{\beta\alpha}^{2})^2=1$.

\paragraph{Intuition for why \cref{assm:simpleGIACheck} implies GIA}

On the other hand, if the last layer is $y=v^\trsp x^2 / \sqrt n$ for $v_\alpha \sim \Gaus(0, 1)$ so that \cref{assm:simpleGIACheck} holds, then a similar calculation with $dx^2 = \sqrt n\pdf{y}{x^2}$ yields
\begin{align}
    \EV dx^1_\alpha
    = 2\sum_{\beta}\EV v_\beta (W_{\beta\alpha}^{2})^2 x_{\alpha}^{1}
      + 2\sum_\beta \sum_{\gamma\ne\alpha}\EV v_\beta W_{\beta\alpha}^{2}W_{\beta\gamma}^{2}x_{\gamma}^{1}
    = 0
    \label{eqn:GIAIntuition}
\end{align}
which now vanishes because $v_\beta$ appears unpaired in the expectation of the first sum and it is independent from everything else.
This illustrates an intuition for \emph{why \cref{assm:simpleGIACheck} implies GIA}: the last layer weights zero out all potential pathways through which $W$ and $W^\trsp$ can correlate.

We have demonstrated our intuitive rules for calculating the kernels that combine to form the NTK.
Now let us rigorously justify these rules.

\section{\texorpdfstring{\netsort{}}{NetsorT}}

To justify our intuitive calculations, we need to pin down the range of architectures they are valid for, and also the precise regularity conditions for the corresponding limits to hold. Here 1) we introduce the \netsort{} language such that an architecture is covered if its forward and backward propagations are expressible in \netsort{}, and 2) we prove a Master Theorem for \netsort{} programs that allows us to justify the intuitions of \cref{keyIntuition} rigorously.

\begin{defn}[Simplified \netsort{}]\label{defn:simpleNetsorT}
For simplicity's sake%
\footnote{See \cref{sec:netsorAppendix} for the formal description of the general notion of \netsort{}; for variable dimension generalization, see \cref{sec:VariableDim}.},
in this section, a \netsort{} program is just a sequence of $\R^n$ vectors inductively generated via one of the following ways from an initial set $\mathcal{V}$ of random $\R^n$ vectors and a set $\mathcal{W}$ of random $n\times n$ matrices 
\begin{description}
    \item [Nonlin\label{instr:nonlin}] Given $\phi:\R^{k}\to\R$ and $x^{1},\ldots,x^{k}\in\R^{n}$, we can generate $\phi(x^{1},\ldots,x^{k})\in\R^{n}$
    \item [{MatMul}\label{instr:matmul}] Given $W\in\R^{n\times n}$ and $x\in\R^{n}$, we can generate $Wx\in\R^{n}$ or $W^{\trsp}x\in\R^{n}$
    \end{description}
\end{defn}

Note that $\phi$ in \ref{instr:nonlin} is applied coordinatewise.
Here, $n$ should be thought of as the width, $\mathcal W$ the weight matrices, and $\mathcal V$ the biases and the first layer \emph{embeddings} of inputs.
Note that $\phi$ in \ref{instr:nonlin} can also be linear, e.g.\ $x,y\mapsto x+y$ in a skip connection.
For example, the RNN equations (\cref{eqn:RNNevolution,eqn:RNNbackprop}) form a natural \netsort{} program:
the initial vectors are $\mathcal V = \{u^{t}(\xi)=U \xi^{t}\}_{t=1}^T \cup \{v = ds^T, b\}$ and the initial matrix is $\mathcal W = \{W\}$, and new vectors $\{g^t, s^t, ds^t, dg^t\}_t$ are formed inductively according to
\begin{align*}
    g^{t}(\xi)&=Ws^{t-1}(\xi)&
    ds^{t-1}&=W^{\trsp}dg^{t}&\textbf{\ref{instr:matmul}}\\
    s^{t}(\xi)&=\phi(g^{t}(\xi)+u^{t}(\xi)+b)&
    dg^{t}&=\phi'\left(g^{t}+u^{t}+b\right)\odot ds^{t}.&\textbf{\ref{instr:nonlin}}
\end{align*}
Like \netsor{} in \citet{yang2019wide} for forward propagation, \netsort{} can express all standard architectures.
For example, in a program expressing convolution neural network with width $n$, the activation vector for each pixel across all channels is represented by an $\R^n$ vector.
See \cref{sec:BackpropInNetsorT} for more details and other examples of modern deep learning layers.
We state the Master Theorem below assuming a generalization of \cref{assm:simpleGIACheck} to a condition called \emph{BP-like} (short for ``backpropagation-like'') for \netsort{} programs; see \cref{defn:BPlike}. On the first read-through, we recommend the reader to mentally replace \emph{BP-like} with \cref{assm:simpleGIACheck} which covers most of the cases we are interested in practice with regard to NTK calculations.
In previous sections, we cared about limits of the form $x^\trsp y / n = \f 1 n \sum_{\alpha=1}^n \psi(x_\alpha, y_\alpha)$ where $\psi$ is the product function.
The Master Theorem tells us how to compute this for almost any function $\psi$.
\begin{thm}[BP-like \netsort{} Master Theorem]\label{thm:netsorTMasterTheoremSimple}
Consider a \netsort{} program.
Suppose: 1) for each initial $W\in\mathcal{W}$, $W_{\alpha\beta}\sim\Gaus(0,\sigma_{W}^{2}/n)$ for an associated variance $\sigma_{W}^{2}$; 2) there is a multivariate Gaussian $Z^{\mathcal{V}}=\left\{ Z^{g}:g\in\mathcal{V}\right\} \in\R^{|\mathcal{V}|}$ such that the initial set of vectors $\mathcal{V}$ are sampled like $\left\{ g_{\alpha}:g\in\mathcal{V}\right\} \sim Z^{\mathcal{V}}$ iid for each $\alpha\in[n]$.
If the program is BP-like and all $\phi$ used in \textbf{\ref{instr:nonlin}}
are polynomially bounded%
\footnote{We say a function $\phi: \R^k \to \R$ is \emph{polynomially-bounded} if $|\phi(x)| \le C\|x\|^p + c$ for some $p, C, c > 0$, for all $x \in \R^k$.}, then
\begin{equation}
\f 1n\sum_{\alpha=1}^{n}\psi(h_{\alpha}^{1},\ldots,h_{\alpha}^{k})\asto\EV\psi(Z^{h^{1}},\ldots,Z^{h^{k}}),\quad\text{as}\quad n\to\infty,
\label{eqn:netsortConvergence}
\end{equation}
for any collection of vectors $h^{1},\ldots,h^{k}$ in the program and any polynomially bounded $\psi:\R^{k}\to\R$, where $Z^{h^{i}}$ are defined in \cref{keyIntuition}.%
\footnote{Difference with \cite[Thm 5.1]{yangScalingLimitsWide2019arXiv.org}: We have gotten rid of the ``rank convergence'' assumption by showing that it comes for free.
See \ref{IH:coreSet} and \cref{lemma:rankStability} in \cref{sec:ProofMainTheorem}.}
\end{thm}

This rigorously justifies the intuitions in the previous section (after checking the regularity conditions).

\paragraph{Back to the MLP example \cref{eqn:NTKparam}}
Assuming $n^{1}=\cdots=n^{L}$, we'd have $\mathcal{W}=\left\{ W^{2},\ldots,W^{L}\right\} $ and $\mathcal{V}=\left\{ b^{l}\right\} _{l}\cup\left\{ W^{1}\xi,W^{1}\bar{\xi}\right\} \cup \{dx^L = \sqrt n W^{L+1}\}$. The weight matrices are by default sampled like in \cref{thm:netsorTMasterTheoremSimple}, and $\mathcal{V}$ is distributed as in \cref{thm:netsorTMasterTheoremSimple} with
$\left\{ Z^{W^{1}\xi},Z^{W^{1}\bar{\xi}}\right\} \sim \Gaus\lp 0, \frac{\sigma_{w}^{2}}{\dim(\xi)}\left(\begin{array}{cc}
\|\xi\|^{2} & \xi^{\trsp}\bar{\xi}\\
\bar{\xi}^{\trsp}\xi & \|\bar{\xi}\|^{2}
\end{array}\right) \rp$ and with $Z^{b^{l}}, Z^{dx^L} \sim\Gaus(0,\sigma_{b}^{2})$ independently.
The forward and backpropagation of the MLP (\cref{eqn:NTKparam}) form a natural \netsort{} program, once we unwind it a little:
We set $g^1(\xi) = W^1 \xi$ via \textbf{\ref{instr:nonlin}} (with identity as the nonlinearity), and
\begin{align*}
    g^l(\xi) &= W^{l}x^{l-1}(\xi)&
    dx^{l-1}(\xi) &= W^{l\trsp} dg^l(\xi)
    &\textbf{\ref{instr:matmul}}\\
    x^{l}(\xi)&=\phi(g^{l}(\xi) + b^l)&
    dg^l(\xi) &= \phi'(g^{l}(\xi) + b^l) \odot dx^l(\xi)
    &\textbf{\ref{instr:nonlin}}
\end{align*}
and similarly for computations on $\bar \xi$.
This program is BP-like because the MLP satisfies \cref{assm:simpleGIACheck}.
For typical activation function $\phi$ like ReLU, $\phi$ and its derivative are both polynomially bounded.
Therefore \cref{thm:netsorTMasterTheoremSimple} applies: for example, with $\psi(x,y)=xy$ applied to the vectors $dh^{l},d\bar{h}^{l}$, \cref{eqn:netsortConvergence} recovers \cref{{eq:MLPgradKernel}} rigorously.

\paragraph*{Summary}
So a formal proof of the NTK convergence proceeds as follows:
\begin{enumerate}
\item Express the network in \netsort{}
\item Check the network satisfies \cref{assm:simpleGIACheck} or more generally the program is BP-like
\item Check that the $\phi$s of the program (which correspond to both the activation functions in the original network and their derivatives) are all polynomially bounded
\end{enumerate}
This is sufficient to show that the NTK converges almost surely as width goes to infinity.
To further compute this limit, follow \cref{eqn:NTKlimit} and \cref{keyIntuition} as in the RNN example in \cref{sec:RNNIntuition}.
As a summary:
\begin{cor}
\label{cor:NTKConvergence} Let $f$ be a (possibly recurrent) neural
network of standard architecture with scalar output and satisfying \cref{assm:simpleGIACheck}. If its nonlinearities
have polynomially bounded weak derivatives, then its NTK $\NTK$ converges almost surely, over any finite set of inputs, to a deterministic kernel $\mathring{\NTK}$
\[
\NTK\asto\mathring{\NTK}
\]
as its widths go to infinity and each of its factored weights $\omega$
and biases $b$ are randomly initialized as $\omega_{\alpha\beta}\sim\Gaus(0,\sigma_\omega^2),b_{\alpha}\sim\Gaus(0,\sigma_b^2)$ for some $\sigma_\omega, \sigma_b \ge 0$.
\end{cor}

See more examples of NTK computations and proofs of convergence (\cref{sec:exampleNTKComputation}) in the appendix.

\begin{remk}[Importance of BP-like Condition]
    Recall the counterexample for GIA in \cref{sec:GIAbreaks}, which can be expressed in a valid but not BP-like \netsort{} program.
    Therefore \cref{thm:netsorTMasterTheoremSimple} is not true when the BP-like condition does not hold.
    We will extend \cref{thm:netsorTMasterTheoremSimple} to cover the non-BP-like cases in a future paper, which requires much more machinery.
\end{remk}

\paragraph{Generalizations}
All results in this section can be generalized to the case where the dimensions in the \netsor{} program are not all equal (such as when an NN has varying widths across layers); see \cref{sec:VariableDim}.
It is easy to show \cref{cor:NTKConvergence} also holds when the output is multidimensional (possibly variable-dimensional, like in a language model).
Architectural blocks like layernorm or attention requires extending \netsort{} to a more powerful language \netsortplus{} (like how \netsorplus{}{} extends \netsor{} in \citet{yang2019wide}), which is discussed in \cref{sec:netsortplus}.

The \netsort{} Master Theorem has implications outside of NTK as well.
For example, most of the semirigorous computations made in the signal propagation literature \citep{poole_exponential_2016,schoenholz_deep_2017,yang_mean_2017,xiao_dynamical_2018,yang_mean_2019} can now be justified rigorously.
See \citet{yangScalingLimitsWide2019arXiv.org} for more discussions.

\paragraph{Guide to the Appendix}
\begin{description}
    \item[\cref{sec:netsorAppendix}] Treats \netsort{} from a formal perspective (similar to the style of \citet{yang2019wide}).
    \item[\cref{sec:netsortplus}] Introduces \netsortplus{} and proves its Master Theorem.
    \item[\cref{sec:VariableDim}] Extends \netsor{} and \netsortplus{} to allow matrices, vectors of variable dimensions.
    \item[\cref{sec:BackpropInNetsorT}] Examples writing forward and backprop of standard architectures in \netsort{}.
    \item[\cref{sec:exampleNTKComputation}] Examples calculating the limiting NTKs for RNN, CNN, transformer, and batchnorm.
    \item[\cref{sec:proofs}] Theoretical tools for our main proof.
    \item[\cref{sec:ProofMainTheorem}] Proof of our main theorem \cref{thm:netsorTMasterTheoremSimple} 
\end{description}

\section{Conclusion}

We showed that for any randomly initialized feedforward or recurrent neural network of standard architecture, its NTK converges almost surely to a deterministic kernel.
We did so by introducing \netsort{}, a language capable of expressing both forward and backward propagation of NNs, along with a tool (\cref{thm:netsorTMasterTheoremSimple}) for understanding the behavior of such computations.
We hope our work lays the foundation for understanding modern overparametrized neural networks.

\section*{Acknowledgements}

We thank Edward Hu, Judy Shen,  Zhiyuan Li, Ilya Razenshteyn, Jason Lee, Huishuai Zhang, Simon Du, Suriya Gunasekar, Etai Littwin, Roman Novak, Jaehoon Lee, Sam Schoenholz, Jascha Sohl-Dickstein, Tomer Galanti, Janardhan Kulkarni, Zeyuan Allen-Zhu, and Jeffrey Pennington for feedback and discussions.

\bibliography{references}
\bibliographystyle{plainnat}
\newpage

\appendix

\section{\texorpdfstring{\netsort{}}{NetsorT}: The Formal Version}
\label{sec:netsorAppendix}

While we recommend using \netsort{} defined in \cref{defn:simpleNetsorT} in practice, we give a formal treatment of the \netsort{} language here and formulate its corresponding Master Theorem (\cref{thm:netsorTMasterTheoremFormal}), which is equivalent to \cref{thm:netsorTMasterTheoremSimple} and which we will prove instead.
The type system of this formal \netsort{} allows us to express the proof of \cref{thm:netsorTMasterTheoremFormal} more easily.

The formal syntax of \netsort{} extends \netsor{} \citep{yang2019wide} by a new transpose (\ref{linetype:Trsp}) instruction.
Compared to \cref{defn:simpleNetsorT}, we allow matrices and vectors to have varying dimension, and explicitly single out the vectors produced by \textbf{MatMul} via an elementary type system.
\begin{defn}\label{defn:netsort}
    \textit{\netsort{} programs} are straightline programs, where each variable follows one of three types, $\Gtype, \Htype$, or $\Atype$ (such variables are called \emph{G-vars}, \emph{H-vars}, and \emph{A-vars}), and after input variables, new variables can be introduced by one of the rules \ref{linetype:MatMul} or
    \ref{linetype:nonlin} to be discussed shortly.
    Variables of $\Gtype$ and $\Htype$ types are vectors, while variables of $\Atype$ type are matrices.
    Each type is annotated by dimensionality information \footnote{formally, we are dealing with dependent types $\Gtype$ and $\Htype$ indexed by $\N$ and $\Atype$ indexed by $\N^2$}:
    \begin{itemize}
        \item If $x$ is a (vector) variable of type $\Gtype$ (or $\Htype$) and has dimension $n$, we write $x: \Gtype(n)$ (or $x: \Htype(n)$).
        \item If $A$ is a (matrix) variable of type $\Atype$ and has size $n_1 \times n_2$, we write $A: \Atype(n_1, n_2)$.
    \end{itemize}
    $\Gtype$ is a \emph{subtype} of $\Htype$, which means that $x: \Gtype(n)$ implies $x: \Htype(n)$.
    A \netsort{} program consists of the following two parts.
    \begin{description}
        \item[Input]
            A set of input G- or A-vars (corresponding to the initial set of matrices and vectors in \cref{defn:simpleNetsorT}).
        \item[Body]
            New variables can be introduced and assigned via the following rules%
        \begin{description}
            \item[\texttt{Trsp}\label{linetype:Trsp}]
                if $A: \Atype(n_1, n_2)$ is an A-var, then we can form its transpose as an A-var:
                \[A^\trsp: \Atype(n_2, n_1)
                \]
                Naturally we identify $(A^\trsp)^\trsp$ with $A$.
            \item[\texttt{MatMul}\label{linetype:MatMul}] if $A: \Atype(n_1, n_2)$ and $x: \Htype(n_2)$, then we can form a G-var via matrix-vector product:
            \[A x : \Gtype(n_1)\]
            \item[\texttt{Nonlin}\label{linetype:nonlin}] If $x^1, \ldots, x^k: \Gtype(n)$ are G-vars with the same dimension $n$ and $\phi: \R^k \to \R$, then we can form an H-var by coordinatewise application of $\phi$
            \[\phi(x^1, \ldots, x^k): \Htype(n)
            \]
        \end{description}
        \item[Output]
        For the purpose of this paper\footnote{In general, the output of a tensor program need not be defined, as most of the time we are concerned with how the H-vars produced over the course of the program interact with each other.}, the output of a \netsort{} program is any function of scalars of the form
        \[\f 1 n \sum_{\alpha=1}^n \psi(h^1_\alpha, \ldots, h^k_\alpha)\]
        for some function $\psi$ and collection of H-vars $h^1, \ldots, h^k$.
    \end{description}
\end{defn}

\begin{remk}
    In comparison with \netsor{} introduced in \cite{yang2019wide}, the language \netsort{} defined here has additionally the transpose (\ref{linetype:Trsp}) instruction and drops the \texttt{LinComb} instruction%
    \footnote{\netsort{} is exactly the \netsormin{} language, introduced in the appendix of \citet{yang2019wide}, augmented with \ref{linetype:Trsp}.}%
    , and additionally has no output.
    While the lack of \texttt{LinComb} and output is just taking away some syntactic sugar, the \ref{linetype:Trsp} instruction significantly expands the expressivity of the language.
    Importantly, it allows us to express backpropagation.
\end{remk}

\begin{algorithm}[tb]
    \caption{MLP Forward and Backward Computation on Network Input $x$}
    \label{tp:MLP}
    \begin{algorithmic}[1]
      \Require $W^1 x: \Gtype(n^1)$ \Comment{layer 1 embedding of input}
      \Require $b^1: \Gtype(n^1)$ \Comment{layer 1 bias}
      \Require $W^2: \Atype(n^2, n^1)$ \Comment{layer 2 weights}
      \Require $b^2: \Gtype(n^2)$ \Comment{layer 2 bias}
      \Require $v: \Gtype(n^2)$ \Comment{readout layer weights}
      \State $x^1 := \phi(W^1 x + b^1): \Htype(n^1)$ \Comment{layer 1 activation; \ref{linetype:nonlin} with $(u, v) \mapsto \phi(u+v)$}
      \State $\tilde h^2 := W^2 x^1: \Gtype(n^2)$ \Comment{\ref{linetype:MatMul}}
      \State $x^2 := \phi(\tilde h^2 + b^2): \Htype(n^2)$ \Comment{layer 2 activation; \ref{linetype:nonlin} with $(u, v) \mapsto \phi(u+v)$}
      \State \Comment{output is $v^\trsp x^2/\sqrt{n^2}$, but this does not need to be expressed in the program}
      \State \Comment{begin backprop}
      \State $W^2{}^\trsp := (W^2)^\trsp : \Atype(n^1, n^2)$
        \Comment{\ref{linetype:Trsp}}
      \State $d x^2 := v: \Gtype(n^2)$
        \Comment{gradient wrt $x^2$ equals the last layer weights, scaled up by $\sqrt{n^2}$}
      \State $d\tilde h^2 := \phi'(\tilde h^2 + b^2) \odot d x^2: \Htype(n^2)$
        \Comment{gradient wrt $\tilde h^2$, scaled up by $\sqrt{n^2}$; \ref{linetype:nonlin}}
      \State $d x^1 := W^2{}^\trsp d \tilde h^2: \Gtype(n^1)$
        \Comment{gradient wrt $x^1$, scaled up by $\sqrt{n^2}$; \ref{linetype:MatMul}}
      \State $d(W^1 x) := \phi'(W^1 x + b^1) \odot d x^1$
        \Comment{gradient wrt the vector $W^1x$, scaled up by $\sqrt{n^2}$; \ref{linetype:nonlin}}
      \State \Comment{Return the NTK value $\NTK(x, x)$; see \cref{eq:MLPNTKDecompLimit}}
      \Ensure
        $\f{\|x^{2}\|^2}{n^2} + $
        $\f{\|d\tilde h^{2}\|^2}{n^2}
        \lp 1 + \f{\|x^1\|^2}{n^1}\rp +$
        $\f{\|d(W^1 x)\|^2}{n^1}
        \lp 1 + \f{\|x\|^2}{\dim(x)}\rp$
    \end{algorithmic}
\end{algorithm}

\begin{algorithm}[t]
    \caption{Simple RNN Forward and Backward Computation on Two Input Sequences}
    \label{tp:RNN}
    \begin{multicols}{2}
    \begin{algorithmic}
      \State {\it // Embeddings of sequence 1 tokens}
      \Require $U x^{11}, \ldots, U x^{T_1 1}: \Gtype(n)$
      \State {\it // Embeddings of sequence 2 tokens}
      \Require $U x^{12}, \ldots, U x^{T_2 2}: \Gtype(n)$
      \State {\it // Weight and bias}
      \Require $W: \Atype(n, n)$
      \Require $b: \Gtype(n)$
      \State {\it // Readout weights}
      \Require $v: \Gtype(n)$
      \State \textit{// The FOR loop is a shorthand for the unrolled straight-line program}
      \For{$a = 1, 2$}
          \State $s^{1a} := \phi(U x^{1a} + b): \Htype(n)$
          \State $\tilde h^{2a} := W s^{1a}: \Gtype(n)$
          \State $s^{2a} := \phi(\tilde h^{2a} + U x^{2a} + b): \Htype(n)$
          \State $\vdots$
          \State $\tilde h^{T_a a} := W s^{T_a-1, a}: \Gtype(n)$
          \State $s^{T_a a} := \phi(\tilde h^{T_a a} + U x^{T_a a} + b): \Htype(n)$
          \State \textit{// Output is $v^\trsp s^{T_a a} / \sqrt n$, but}
          \State \textit{// we don't express this in the program}
          \State \textit{// --- Backprop ---}
          \State {\it // $\forall$ variable $u$, $du$ represents $\sqrt n\nabla_u \text{out}$}
          \State $ds^{T_a a} := v: \Gtype(n)$
          \State {\it // $\phi'$ is derivative of $\phi$}
          \State $d\tilde h^{T_a a} := \phi'(\tilde h^{T_a a} + Ux^{T_a a} + b) \odot ds^{T_a a}: \Htype(n)$
          \State $ds^{T_a-1, a} := W^\trsp d\tilde h^{T_a a}: \Gtype(n)$
          \State $d\tilde h^{T_a-1, a} := \phi'(\tilde h^{T_a-1, a} + U x^{T_a-1, a} + b) \odot ds^{T_a-1, a}: \Htype(n)$
          \State $ds^{T_a-2, a} := W^\trsp d\tilde h^{T_a-1, a}: \Gtype(n)$
          \State $\vdots$
          \State $d s^{1 a} := W^\trsp d\tilde h^{2a}: \Gtype(n)$
          \State $d \tilde h^{1a} := \phi'(Ux^{1a} + b) \odot ds^{1a}: \Htype(n)$
      \EndFor
      \State \textit{// Return NTK evaluated on sequences $x^1, x^2$}
      \State \textit{// See \cref{eqn:NTKlimit}}
      \Ensure
        $
        \f{s^{T_1 1}{}^\trsp s^{T_2 2}}n +$
      \State $\ \ 
        \sum_{i=1}^{T_1} \sum_{j=1}^{T_2}
        \f{d \tilde h^{i1}{}^\trsp d \tilde h^{j2}} n \times$
      \State $\quad\quad
        \lp 1 + \f{s^{i-1, 1}{}^\trsp s^{j-1, 2}} n
        + \f{x^{i1}{}^\trsp x^{j2}}{\dim(x^{i1})}
        \rp
        $
    \end{algorithmic}
    \end{multicols}
\end{algorithm}

\paragraph{Examples}
    \cref{tp:MLP,tp:RNN} write out the forward and backward computation of resp.\ an MLP and a simple RNN.
    We remark on a few things:
    First, notice that the new transpose instruction \ref{linetype:Trsp} allows us to express backpropagation.
    Second, as in \citet{yang2019wide}, \emph{we account for the input $x$ through its embedding $W^1 x$, not $x$ itself.}
    This is because 1) our theorems concern the case where all input G-vars are random;
    in the context of expressing neural network computation, $x$ is a deterministic input, while
    $W^1x$ is a Gaussian vector when $W^1$ has iid Gaussian entries; 2) $x$ has a fixed dimension, while we intend all dimensions (like $n^1, n^2$) in the \netsor{} program to tend to infinity, as we'll describe shortly.
    Third, weight-sharing is easily expressed because we can arbitrarily re-use A-vars.

Programs expressing backpropagation have a special property that we would like to isolate, and which will reduce the complexity the \netsort{} master theorem we need to prove.
It is a tensor program generalization of \cref{assm:simpleGIACheck} for neural networks.
\begin{defn}\label{defn:BPlike}
A \netsort{} program is said to be \emph{BP-like} if there is a special nonempty set of input G-vars $v^1, \ldots, v^k$ (intuitively, these should be thought of as the readout weights of the forward computation) such that
\begin{enumerate}
    \item If $W^\trsp z$ is used in the program for some H-var $z$, and $W$ is an input A-var, then $z$ must be an odd function of $v^1, \ldots, v^k$, in the sense that, fixing all other G-vars, if $v^1, \ldots, v^k$ are negated simultaneously, then $z$ is negated as well:
    \[z(-v^1, \ldots, -v^k, \text{all other G-vars}) = -z(v^1, \ldots, v^k, \text{all other G-vars}).\]
    \item If $W z$ is used in the program for some H-var $z$, and $W$ is an input A-var, then $z$ cannot depend on any of $v^1, \ldots, v^k$.
    \item $v^1, \ldots, v^k$ are sampled with zero mean (but possibly with nontrivial covariances) and independently from all other G-vars.
\end{enumerate}
\end{defn}

\begin{remk}\label{remk:backpropIsBPLike}
    A \netsort{} program expressing backpropagation of a network satisfying \cref{assm:simpleGIACheck} can be seen to be BP-like as follows:
    We can always write the program such that the ``forward computation'', up to but before applying readout weights, appears first, followed by the ``backward computation'' (scaled up by $\sqrt{\text{width}}$, as exemplified by \cref{tp:MLP,tp:RNN}).
    If the network output is a scalar $\text{out} = v{}^\trsp x/\sqrt{n}$ with readout weights $v$, then $v$ is not used until the backward computation, where we have $\sqrt{n} \pd \text{out} / \pd x = v$.
    This fulfills condition 2 in \cref{defn:BPlike}.
    In the backward computation, only transposed matrices (A-vars) appear in \ref{linetype:MatMul} lines, and all vectors (H-vars) are linear functions of (and thus are odd in) $v$, because backpropagation is linear in output gradients.
    This fulfills condition 1 in \cref{defn:BPlike}.
\end{remk}

Like in \citet{yang2019wide}, the G-vars in a BP-like \netsort{} program will roughly jointly Gaussian in each coordinate slice.
We keep track of their mean and covariance using the usual recursive equations.

\begin{align}
\tmu(g) & =\begin{cases}
\muin(g) & \text{if \ensuremath{g} is input}\\
0 & \text{otherwise}
\end{cases},\nonumber \\
\tSigma(g,\bar{g}) & =\begin{cases}
\Sigmain(g,g') & \text{if \ensuremath{g,g'} are inputs}\\
\sigma_{W}^{2}\EV_{Z}\phi(Z)\bar{\phi}(Z) & \text{if \ensuremath{g=Wh,\bar{g}=W\bar{h}}, }\\
0 & \text{otherwise}
\end{cases}\label{eqn:extendedMuSigma}
\end{align}

\begin{setup}
    \label{assm:equalDimSampling}
    For \netsort{} program:
    For simplicity, assume all dimensions in the program are equal to $n$.
    Suppose for each A-var $W: \Atype(n, n)$, we sample $W_{\alpha \beta} \sim \Gaus(0, \sigma_W^2/n)$ for some $\sigma_W^2 > 0$, and for each
    $\alpha \in [n]$, we sample, i.i.d., $\{x_\alpha: x \text{ is input G-var}\} \sim \Gaus(\muin, \Sigmain)$ for some mean $\muin$ and (possibly singular) covariance $\Sigmain$ over input G-vars.
\end{setup}

Finally, we have the BP-like \netsort{} Master theorem.

\begin{restatable}[BP-like \netsort{} Master Theorem]{thm}{NetsorTMasterTheoremFormal}
    \label{thm:netsorTMasterTheoremFormal}
    Fix any BP-like \netsort{} program satisfying \cref{assm:equalDimSampling} and with all nonlinearities polynomially-bounded.
    If $g^1, \ldots, g^M$ are all of the G-vars in the entire program, including all input G-vars, then for any polynomially-bounded $\psi: \R^M \to \R$, as $n \to \infty$,
    \begin{align*}
        \f 1 n \sum_{\alpha=1}^n \psi(g^1_\alpha, \ldots, g^M_\alpha) \asto 
        \EV_{Z \sim \Gaus(\tmu, \tSigma)}\psi(Z)
        =
        \EV_{Z \sim \Gaus(\tmu, \tSigma)}\psi(Z^{g^1}, \ldots, Z^{g^M}),
    \end{align*}
    where $\asto$ means almost sure convergence,
    $Z = (Z^{g^1}, \ldots, Z^{g^M}) \in \R^M$, and $\tmu = \{\tmu(g^i)\}_{i=1}^M \in \R^M$ and $\tSigma = \{\tSigma(g^i, g^j)\}_{i,j=1}^M \in \R^{M \times M}$ are given in \cref{eqn:extendedMuSigma}.
    See \cref{fig:mastertheoremIllustration} for an illustration.
\end{restatable}

\begin{figure}[t]
    \centering
    \includegraphics[width=\textwidth]{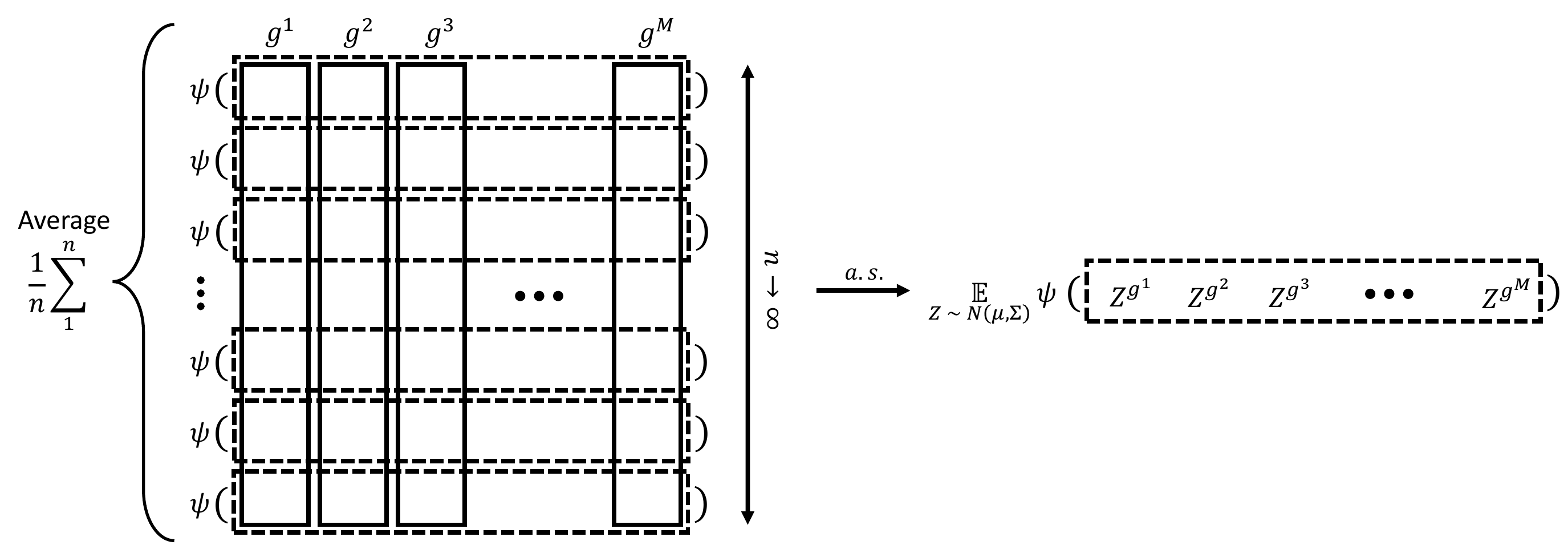}
    \caption{An illustration of the \netsort{} Master Theorem \cref{thm:netsorTMasterTheoremFormal}.}
    \label{fig:mastertheoremIllustration}
\end{figure}

This is equivalent to \cref{thm:netsorTMasterTheoremSimple}, but emphasizes that the main source of randomness comes from the G-vars (i.e. approximately Gaussian vectors) $g^1, \ldots, g^M$ whose distribution can be explicitly computed via \cref{eqn:extendedMuSigma}.

\section{\texorpdfstring{\netsortplus{}}{NetsorT+}}
\label{sec:netsortplus}

In this section we augment \netsort{} with a constant type to form \netsortplus{}, like how \netsor{} is augmented likewise to form self-parametrized \netsorplus{} in \citet{yang2019wide}.
The main result here is the \emph{BP-like \netsortplus{} Master Theorem} (\cref{thm:selfParamNetsorplusMasterTheorem}).

We first formally define the \netsortplus{} language.

\newcommand{\Ctype}{\mathsf{C}}
\begin{defn}
\label{defn:selfParam}
A \emph{\netsortplus{} program}%
\footnote{
    In the language of \citet{yang2019wide}, \netsortplus{} actually corresponds to self-parametrized \netsorplus{} with \ref{linetype:Trsp}, but we omit ``self-parametrized'' in the name because this is the primary form of \netsortplus{} we will use in practice.
    What would be the plain \netsortplus{} in the language of \citet{yang2019wide} would just be a fragment of of the \netsortplus{} in \cref{defn:selfParam}, so our Master Theorem will also cover that case.
    }
is a \netsort{} program where we have an additional scalar type, called $\Ctype$, which should intuitively be thought of as a random variable that tends to a deterministic limit (i.e. a \emph{$\Ctype$onstant}) almost surely.
Colloquially, we will call variables of type $\Ctype$ ``C-vars.''
C-vars can be used as parameters of nonlinearities.

For completeness, we specify a \netsorplus{} program as follows:
\begin{description}
    \item[Input]
        A set of input C-vars, in addition to the G- and A-vars allowed in \cref{defn:netsort}.
    \item[Body]
        New variables can be introduced and assigned via the following rules
    \begin{description}
        \item[\texttt{MatMul}] Same as in \cref{defn:netsort}.
        \item[\texttt{Trsp}] Same as in \cref{defn:netsort}.
        \item[\texttt{Nonlin$^+$}\label{linetype:nonlin+}] If $x^1, \ldots, x^k: \Gtype(n)$ are G-vars with the same dimension $n$, $\theta_1, \ldots, \theta_l: \Ctype$ are C-vars, and $\phi(-; -): \R^k \times \R^l \to \R$ is a parametrized function, then we may create an H-var
        \[\phi(x^1, \ldots, x^k; \theta_1, \ldots, \theta_l): \Htype(n)\]
        where $\phi(- ; \theta_1, \ldots, \theta_l)$ acts coordinatewise.
        \item[\texttt{Moment}\label{linetype:moment}]
        If $x^1, \ldots, x^k: \Gtype(n)$ are G-vars with the same dimension $n$, $\theta_1, \ldots, \theta_l: \Ctype$ are C-vars, and $\phi(-; -): \R^k \times \R^l \to \R$ is a parametrized function, then we may create a C-var
        \[\f 1 n \sum_{\alpha=1}^n\phi(x^1_\alpha, \ldots, x^k_\alpha; \theta_1, \ldots, \theta_l): \Ctype.\]
    \end{description}
    \item[Output]
        Same as in \cref{defn:netsort}.
\end{description}
\end{defn}
See \cref{sec:BackpropInNetsorT} for examples of layernorm and attention written in \netsortplus{}.

The following gives the \netsortplus{} Master Theorem, which first lists the regularity conditions needed (\emph{rank stability} and \emph{parameter-control}), along with some natural notations defined later, before stating the main convergence results.
\begin{thm}[BP-like \netsortplus{} Master Theorem]
    \label{thm:selfParamNetsorplusMasterTheorem}
    Fix any BP-like \netsortplus{} program sampled in the natural way as in \cref{assm:netsorplusScalarLimit} and also satisfying rank stability (\cref{assm:asRankStab}).
    Let $\varphi^\bullet, \bigtheta^\bullet, \tmu, \tSigma, \mathring{(\phantom{\theta})}$ be as in \cref{defn:netsorplusMuSigma}.    
    Suppose for every H-var or C-var $u$, $\varphi^u(-; \bigvtheta^u)$ is parameter-controlled at $\mathring{\bigvtheta}^u$.
    Then the following hold.
    \begin{enumerate}
        \item
        Let $g^1, \ldots, g^M$ be all of the G-vars in the program (including all input G-vars).
        Then for any polynomially bounded $\psi: \R^M \to \R$, we have
        \begin{align*}
            \f 1 n \sum_{\alpha=1}^n \psi(g^1_\alpha, \ldots, g^M_\alpha) \asto \EV_{Z \sim \Gaus(\tmu, \tSigma)}\psi(Z).
        \end{align*}
        More generally, for any $l$, for any random vector $\bigtheta \in \R^l$ that converges almost surely to a deterministic vector $\mathring{\bigtheta}$, as $n \to \infty$, and for any $\psi(-; -): \R^M\times \R^l \to \R$ parameter-controlled at $\mathring{\bigtheta}$,
        \begin{align*}
            \f 1 n \sum_{\alpha=1}^n \psi(g^1_\alpha, \ldots, g^M_\alpha; \bigtheta) \asto \EV_{Z \sim \Gaus(\tmu, \tSigma)}\psi(Z; \mathring{\bigtheta}).
        \end{align*}
        \item
        Each C-var $\theta$ converges to its natural limit $\mathring \theta$:
        \begin{align*}
        \theta \asto \mathring \theta.
        \end{align*}
    \end{enumerate}
    
\end{thm}

We now expand on the assumptions and definitions used in the Master Theorem.

\begin{assm}\label{assm:netsorplusScalarLimit}
    Fix a self-parametrized \netsortplus{} program satisfying \cref{assm:equalDimSampling}.
    Assume each input C-var $\theta$ is sampled in a way such that $\theta \asto \mathring \theta$ as $n \to \infty$ for some deterministic scalar $\mathring \theta \in \R$.
\end{assm}

\begin{defn}\label{defn:netsorplusMuSigma}
    Fix a \netsortplus{} program with scalar variables satisfying \cref{assm:netsorplusScalarLimit}.
    For the purpose of this definition, write $g^1, \ldots, g^M$ for the entirety of the G-vars in the program, including input G-vars.
    
    \emph{New Notations: $\varphi^\bullet, \bigtheta^\bullet$}\quad
    For each H-var $h = \phi(g^1, \ldots, g^M; \theta_1, \ldots, \theta_l)$ introduced by \ref{linetype:nonlin+}, set $\varphi^h \defeq \phi$ and $\bigtheta^h \defeq (\theta_1, \ldots, \theta_l)$.
    For each G-var $g^i$, this means that $\varphi^{g^i}(x^1, \ldots, x^M) = x^i$ and $\bigtheta^{g^i} = () \in \R^0$ is the empty vector.
    Likewise, for each C-var $c = \f 1 n \sum_{\alpha=1}^n \phi(g^1_\alpha, \ldots, g^M_\alpha; \theta_1, \ldots, \theta_l)$ introduced by \ref{linetype:moment}, set $\varphi^c \defeq \phi$ and $\bigtheta^c \defeq (\theta_1, \ldots, \theta_l)$.
    
    \emph{Extending the $\mathring{(\phantom{\theta})}$ notation from \cref{assm:netsorplusScalarLimit} and the
    Recursive Definition of $\tmu$ and $\tSigma$}\quad
    Given $\muin$ and $\Sigmain$ as in \cref{assm:equalDimSampling}, we define $\mu$ and $\Sigma$ on G-vars, along with ``limit scalars'' $\mathring{\theta}$ for each C-var $\theta$ (extending $\mathring \theta$ given by \cref{assm:netsorplusScalarLimit} for input $\theta$), as follows:
    For any pair of G-vars $g, \bar g$, we define recursively
    \begin{align*}
        \tmu(g)
            &\defeq
                \begin{cases}
                \muin(g)  &   \text{if $g$ is input}\\
                0   &   \text{otherwise}
                \end{cases}
                \\
        \tSigma(g, \bar g)
            &\defeq
                \begin{cases}
                \Sigmain(g, \bar g)   &   \text{if $g, g'$ are inputs}\\
                \sigma^2_W \EV_Z \varphi^h(Z; \mathring{\bigvtheta}^h) \varphi^{\bar h}(Z; \mathring{\bigvtheta}^{\bar h}) &   \text{if $g = Wh, \bar g=W\bar h$}\\
                0   &   \text{otherwise}
                \end{cases}
                \numberthis\label{eqn:selfParamExtendedMuSigma}
    \end{align*}
    where $W$ is any A-var, transposed or not%
    \footnote{
        In \cref{eqn:selfParamExtendedMuSigma}, if $W$ is an input A-var, then $\sigma_W$ is as in \cref{assm:equalDimSampling}; if $W = \bar W^\trsp$ for some input G-var $\bar W$, then $\sigma_{W} = \sigma_{\bar W}$.
        Note: when we allow variable dimensions in the program, this is defined slightly differently; see \cref{eqn:extendedMuSigmaCDC}.
    };
    and for each C-var $\theta$ introduced by \ref{linetype:moment},
    \begin{align*}
    \mathring \theta \defeq
        \EV_Z \varphi^\theta(Z; \mathring {\bigtheta}^\theta).
        \numberthis\label{eqn:mathringtheta}
    \end{align*}
    In all of the equations above, $Z \sim \Gaus(\tmu, \tSigma)$ is a random Gaussian vector with an entry for each G-var in the program.
\end{defn}

\paragraph{Parameter-Control}
We adapt the definition of \emph{parameter-control} from \citet{yang2019wide} to our setting.
\begin{defn}\label{defn:parameterControlled}
    We say a parametrized function $\phi(-; -): \R^k \times \R^l \to \R$ is \emph{polynomially parameter-controlled} or just \emph{parameter-controlled} for short%
    \footnote{This overloads the meaning of \emph{parameter-controlled} from \citet{yang2019wide}, where the definition replaces the ``polynomially bounded'' in the definition here with ``bounded by $e^{C\|\cdot\|^{2-\epsilon} + c}$ for some $C, c, \epsilon > 0$.''
    In this paper, we shall never be concerned with the latter (more generous) notion of boundedness, so there should be no risk of confusion.}
    , at $\mathring {\bigtheta} \in \R^l $ if
    \begin{enumerate}
        \item $\phi(-; \mathring{\bigtheta})$ is polynomially bounded, and\label{item:parameterControlled1}
        \item there are some polynomially bounded $\bar \phi:\R^k \to \R$ and some function $f: \R^l \to \R^{\ge 0} \cup \{\infty\}$ that has $f(\mathring{\bigtheta}) = 0$ and that is continuous at $\mathring \bigtheta$, such that, for all $x^1, \ldots, x^k \in \R$ and $\bigtheta \in \R^l$,
    \[
    |\phi(x^1, \ldots, x^k; \bigtheta) - \phi(x^1, \ldots, x^k; \mathring{\bigtheta})|
    \le
    f(\bigtheta) \bar \phi(x^1, \ldots, x^k)
    .
    \]
    \label{item:parameterControlled2}
    \end{enumerate}
    \end{defn}
    Note that $f$ and $\bar\phi$ here can depend on $\mathring{\bigtheta}$.
    The following examples come from \citet{yang2019wide}.
    \begin{exmp}\label{exmp:parameterControl}
    Any function that is (pseudo-)Lipschitz\footnote{A pseudo-Lipschitz function $\phi: \R^r \to \R$ is one that satisfies
    \[
    |\phi(x) - \phi(y)| \le C\|x - y\|(\|x\|^p + \|y\|^q + 1)
    \]
    for some constants $C, p, q \ge 0$.
    Roughly speaking, pseudo-Lipschitz functions are those that have polynomially bounded weak derivatives.
    }
    in $x^1, \ldots, x^k$ and $\bigtheta$ is polynomially parameter-controlled.
    An example of a discontinuous function that is polynomially parameter-controlled is $\phi(x; \theta) = \mathrm{step}(\theta x)$.
    Then for $\mathring \theta \ne 0$, 
    \[
    |\phi(x; \theta) - \phi(x; \mathring \theta)| \le
    \f{|\mathring \theta - \theta|}{|\mathring \theta|}
    ,
    \]
    so we can set $f(\theta) = \f{|\mathring \theta - \theta|}{|\mathring \theta|}$ and $\bar \phi = 1$ in \cref{defn:parameterControlled}.
\end{exmp}

\paragraph{Rank Stability}
The following assumption says that the vectors in a program should not change any linear dependence relations abruptly in the infinite $n$ limit.
\begin{assm}[Rank Stability]\label{assm:asRankStab}
    For any $W: \Atype(n, m)$ and any collection $\Ss \sbe \{(h: \Htype(m)) \mid \exists (g: \Gtype(n)), g := W h\}$,
    let $H \in \R^{m \times |\Ss|}$ be the matrix whose columns are $h \in \Ss$.
    If $\f 1 m H^\trsp H \in \R^{|\Ss| \times |\Ss|}$ converges almost surely to some $\mathring C$ as $n, m \to \infty$ with convergent ratio $n/m \to \alpha$, then almost surely $\rank H = \rank \mathring C$ for all large $n$ and $m$.
\end{assm}
Some remarks
\begin{itemize}
    \item An example violating rank stability would be $\Ss = \{1, 1+1/n\}$, vectors with constant entries 1 and $1+1/n$, which are linearly independent for any finite $n$ but their kernel matrix becomes singular in the limit $n \to \infty$.
    \item Note that a common situation where rank stability holds is when all limit $\mathring C$ matrices are full rank.
    By the lower semi-continuity of rank, $\rank H = \rank \mathring C$ must hold asymptotically.
    \item There are counterexamples to the \netsortplus{} Master Theorem if rank stability is not assumed \citep{yang2019wide}.
    \item Note that we did not assume \cref{assm:asRankStab} explicitly in \cref{thm:netsorTMasterTheoremSimple,thm:netsorTMasterTheoremFormal} because we in fact get it for free (\cref{lemma:rankStability}).
\end{itemize}
See \citet{yang2019wide} for further discussions of the rank stability assumption.

\subsection{The Simplified \texorpdfstring{\netsortplus{}}{NetsorT+}}

We can simplify the above formal description of \netsortplus{} like how we simplified \netsort{} in the main text.

\begin{defn}[Simplified \netsortplus{}]\label{defn:simpleNetsorTPlus}
    A simplified \netsortplus{} program is just a sequence of vectors and scalars recursively generated from an initial set of random $n\times n$ matrices $\mathcal{W}$, random size $n$ vectors $\mathcal{V}$, and random scalars $\mathcal C$ via one of the following ways
    \begin{description}
        \item [{Nonlin}\label{instr:nonlin+}] Given $\phi:\R^{k} \times \R^l \to\R$, previous scalars $\theta_1, \ldots, \theta_l \in \R$ and vectors $x^{1},\ldots,x^{k}\in\R^{n}$, we can generate a new vector
        \[\phi(x^{1},\ldots,x^{k}; \theta_1, \ldots, \theta_l)\in\R^{n}\]
        where $\phi(-; \theta_1, \ldots, \theta_l)$ applies coordinatewise to each $\alpha$-slice $(x^1_\alpha,\ldots, x^k_\alpha)$.
        \item [{Moment}\label{instr:moment}] Given same setup as above, we can also generate a new scalar
        \[\f 1 n \sum_{\alpha=1}^n \phi(x^{1}_\alpha,\ldots,x^{k}_\alpha; \theta_1, \ldots, \theta_l)\in\R\]
        \item [{MatMul}] Given $W\in\R^{n\times n}$ and $x\in\R^{n}$, we can generate $Wx\in\R^{n}$ or $W^{\trsp}x\in\R^{n}$
        \end{description}
\end{defn}
Here the (initial) matrices, vectors, and scalars correspond to the (input) A-, H-, and C-vars in \cref{defn:selfParam}, and the rules with the same name mirror one another.
The main difference between this version of \netsortplus{} and \cref{defn:selfParam} is that we are implicitly allowing \textbf{Nonlin} and \textbf{Moment} to take in H-vars here instead of only G-vars.
Nevertheless, their expressive powers are equivalent, since any vector generated with \cref{defn:simpleNetsorTPlus} is generated from a chain of \textbf{Nonlin} that ends up in G-vars (those vectors created by \textbf{MatMul}), and we can just collapse parametrized nonlinearities into a single parametrized nonlinearity that takes in G-vars only.
For example, if $z = \phi(x^1, x^2; \theta_1), x^1 = W v, x^2 = \psi(y; \theta_2), y = W u$, then $z$ can be directly expressed in terms of G-vars: $z = \overline\phi(W v, W u; \theta_1, \theta_2) \defeq \phi(Wv, \psi(W u; \theta_2); \theta_1)$.
Therefore, we make the following definition
\begin{defn}[$\varphi^\bullet, \bigtheta^\bullet$ Notation]
    For any size $n$ vector $x$ in \cref{defn:selfParam}, let $\varphi^x$ and $\bigtheta^x$ be the parametrized nonlinearity and the scalars such that $x = \varphi^x(z^1, \ldots, z^k; \bigtheta^x)$ for some G-vars $z^1, \ldots, z^k$.
    Likewise, for any scalar $c$ in \cref{defn:selfParam}, let $\varphi^c$ and $\bigtheta^c$ be the parametrized nonlinearity and the scalars such that $c = \f 1 n \sum_{\alpha=1}^n \varphi^x(z^1_\alpha, \ldots, z^k_\alpha; \bigtheta^c)$ for some G-vars $z^1, \ldots, z^k$.
\end{defn}

Then we can write down a Master Theorem in the style of \cref{thm:netsorTMasterTheoremSimple} for \cref{defn:simpleNetsorTPlus}-style \netsortplus{} programs.
\begin{pabox}[box:netsortPlusKeyIntuition]{How to Intuitively Understand a Simplified \netsortplus{} Program}
    Consider a \netsortplus{} program sampled as in \cref{thm:netsorTPlusMasterTheoremSimple}.
    When $n\gg 1$, each vector $x \in \R^n$ in the program has roughly iid coordinates distributed like a random variable $Z^x$, and each scalar $\theta \in \R$ is close to a deterministic scalar $\mathring \theta$, with $Z^x$ and $\mathring \theta$ defined recursively as below.
    \begin{description}
        \item[Nonlin] If $y = \phi(x^1, \ldots, x^k; \theta_1, \ldots, \theta_l)$, then
        \[Z^y = \phi(Z^{x^1}, \ldots, Z^{x^k}; \mathring \theta_1, \ldots, \mathring \theta_l).\]
        \item[Moment] If $\theta = \f 1 n \sum_{\alpha=1}^n \phi(x^{1}_\alpha,\ldots,x^{k}_\alpha; \theta_1, \ldots, \theta_l)$, then
        \[\mathring \theta = \EV_Z \phi(Z^{x^1}, \ldots, Z^{x^k}; \mathring \theta_1, \ldots, \mathring \theta_l).\]
        \item[MatMul]
        For any set of infinite vectors $\mathcal{X}$ and matrix $W\in\mathcal W$, the set of random
        variables $\left\{ Z^{Wx}:x\in\mathcal{X}\right\} $ is jointly Gaussian
        with zero mean and covariance
        \[
        \Cov\left(Z^{Wx},Z^{W\bar{x}}\right)=\sigma_{W}^{2}\EV Z^{x}Z^{\bar{x}}.
        \]
        If $\mathcal{Y}$ is any set of $\R^n$ vectors and $\bar W \ne W$, then
        $\{ Z^{Wx}:x\in\mathcal{X}\} $
        is independent from $\{ Z^{\bar W y}:y\in\mathcal{Y}\} $.
    \end{description}
\end{pabox}
\begin{thm}[Simplified BP-like \netsort{} Master Theorem]\label{thm:netsorTPlusMasterTheoremSimple}
    Consider a \netsortplus{} program in the style of \cref{defn:simpleNetsorTPlus}.
    Suppose: 1) for each initial $W\in\mathcal{W}$, $W_{\alpha\beta}\sim\Gaus(0,\sigma_{W}^{2}/n)$ for an associated variance $\sigma_{W}^{2}$; 2) there is a multivariate Gaussian $Z^{\mathcal{V}}=\left\{ Z^{g}:g\in\mathcal{V}\right\} \in\R^{|\mathcal{V}|}$ such that the initial set of vectors $\mathcal{V}$ are sampled like $\left\{ g_{\alpha}:g\in\mathcal{V}\right\} \sim Z^{\mathcal{V}}$ iid for each $\alpha\in[n]$; 3) each initial scalar $\theta$ tends to a deterministic constant $\mathring \theta$ as $n \to\infty$.
    Suppose the program is BP-like and $\varphi^u(-;-)$ is parameter-controlled at $\mathring \bigtheta^u$ for all vectors and scalars $u$.
    Assume the program satisfies rank stability (\cref{assm:asRankStab}).

    Recursively define $Z^h$ for each vector $h$ and $\mathring \theta$ for each scalar $\theta$ in the program as in \cref{box:netsortPlusKeyIntuition}.
    Then the following hold.
    \begin{enumerate}
        \item
        For any $l$, for any random vector $\bigtheta \in \R^l$ that converges almost surely to a deterministic vector $\mathring{\bigtheta}$ as $n \to \infty$, for any vectors $x^1, \ldots, x^M \in \R^n$ in the program, and for any $\psi(-; -): \R^M\times \R^l \to \R$ parameter-controlled at $\mathring{\bigtheta}$,
        \begin{align*}
            \f 1 n \sum_{\alpha=1}^n \psi(x^1_\alpha, \ldots, x^M_\alpha; \bigtheta) \asto \EV\psi(Z^{x^1}, \ldots, Z^{x^M}; \mathring{\bigtheta}).
        \end{align*}
        \item
        Each C-var $\theta$ converges to its natural limit $\mathring \theta$:
        \begin{align*}
        \theta \asto \mathring \theta.
        \end{align*}
    \end{enumerate}

\end{thm}

\section{Programs with Variable Dimensions}
\label{sec:VariableDim}

\paragraph{Notation}
In this section, we let $\dim(x)$ denote the dimension of an H-var $x$.

As in \citet{yang2019wide}, we focused in the main text on the case where all dimensions of vectors in a \netsort{} program are equal, but this does not have to be the case.
In this section, we state the Master Theorem for BP-like \netsort{} and \netsortplus{} programs with variable dimensions.
The main idea is exactly the same as before, but we require some more notation to describe how the limit is taken when all the widths vary.

First, there are some obvious dimensionality constraints even when they do vary, induced by the rules we apply to introduce variables:
\begin{equation}
\begin{cases}
\text{If $y = \phi(x^1, \ldots, x^k)$, then $\dim(y) = \dim(x^i), \forall i$; similarly for \ref{linetype:nonlin+} and \ref{linetype:moment}.}
\\
\text{If $y = W x$ and $\bar y = W \bar x$, then $\dim(x) = \dim(\bar x)$ and $\dim(y) = \dim(\bar y)$.}
\end{cases}
\label{eqn:dimconstraint}
\end{equation}

\begin{defn}
Given an equivalence relation $\simeq$ on the input G-vars of a program, we extend this to an equivalence relation on all H-vars of the program by
\begin{equation}
h \equiv h' \iff h \simeq h' \text{ OR $h$ and $h'$ are constrained to have the same dimension by (\ref{eqn:dimconstraint})}.
\end{equation}
We call any such equivalence class a \emph{Common Dimension Class}, or CDC.
\end{defn}
Intuitively, the dimensions of H-vars in each CDC are all the same but can be different in different CDCs.

\begin{exmp}
In \cref{tp:MLP}, if we let different layers have different widths, then the CDCs are $\{W^1 x, b^1, x^1, dx^1, d(W^1x)\}$ and $\{b^2, v, \tilde h^2, x^2, dx^2, d\tilde h^2\}$.
If we tie the widths, then all of these H-vars are in the same CDC.
In \cref{tp:RNN}, all G-vars are in the same CDC, and given the body of the program, this is the only way to partition the H-vars into CDCs, because the reuse of $W$ across time step ties all H-var dimensions to be equal.
\end{exmp}

The following describes the sampling and CDCs of \netsort{} programs we are interested in.

\begin{assm}\label{assm:samplingVariableDim}
Fix a \netsort{} or \netsortplus{} program with some equivalence relation on the input G-vars, and thus with induced CDCs over its H-vars.
Assume the dimensions in each CDC are the same, but the dimensions of different CDCs can vary.
Suppose for each input A-var $W: \Atype(m', m)$, we sample $W_{\alpha \beta} \sim \Gaus(\sigma_W^2/m)$ for some $\sigma_W^2 > 0$.
For each transpose A-var $W^\trsp: \Atype(m, m')$, we also set $\sigma_{W^\trsp}^2 = \f{m'}{m} \sigma_W^2$.
Suppose further for each CDC $\cdc$ with dimension $n$, for each $\alpha \in [n]$, we sample, i.i.d., $\{x_\alpha: x \in \cdc \text{ and }x\text{ is input G-var}\} \sim \Gaus(\mu^\cdc, \Sigma^\cdc)$ for some mean $\mu^\cdc$ and covariance $\Sigma^\cdc$ over input G-vars in $\cdc$.
\end{assm}

Then the following result is an easy extension of \cref{thm:netsorTMasterTheoremFormal}.
\begin{restatable}[BP-like \netsort{} Master Theorem; Variable Dimensions]{thm}{netsorMasterTheoremVarDim}
\label{thm:netsorMasterTheoremVarDim}
Fix any BP-like \netsort{} program satisfying \cref{assm:samplingVariableDim} and with all nonlinearities polynomially bounded.
Consider the limit where all dimensions go to infinity, with their pairwise ratios tending to finite but nonzero values:
\[\forall W : \Atype(n', n), \quad\text{we have}\quad n'/n \to \rho
\quad\text{for some $\rho \in (0, \infty)$}.\]
Then for each transpose $W^\trsp$ of an input A-var $W: \Atype(n', n)$, we have
\[\sigma_{W^\trsp}^2 = \f{n'}{n} \sigma_W^2 \to \mathring \sigma_{W^\trsp}^2\]
for some limit $\mathring \sigma_{W^\trsp}^2$.
For each input A-var $W$, we also set $\mathring \sigma_W = \sigma_W$.

For any CDC $\cdc$, if $g^1, \ldots, g^M: \Gtype(n)$ are all of the G-vars in $\cdc$ (including all input G-vars), then for any polynomially bounded $\psi: \R^M \to \R$, as all dimensions in the program tend to infinity (not just the dimension of $\cdc$) in the manner above, we have
\begin{align}
    \f 1 n \sum_{\alpha=1}^n \psi(g^1_\alpha, \ldots, g^M_\alpha) \asto 
    \EV_{Z \sim \Gaus(\tmu^\cdc, \tSigma^\cdc)}\psi(Z)
    =
    \EV_{Z \sim \Gaus(\tmu^\cdc, \tSigma^\cdc)}\psi(Z^{g^1}, \ldots, Z^{g^M}),
    \label{eqn:netsorTVarDimLimit}
\end{align}
where $\asto$ means almost sure convergence,
$Z = (Z^{g^1}, \ldots, Z^{g^M}) \in \R^M$, and $\tmu^\cdc = \{\tmu^\cdc(g^i)\}_{i=1}^M \in \R^M$ and $\tSigma^\cdc = \{\tSigma^\cdc(g^i, g^j)\}_{i,j=1}^M \in \R^{M \times M}$ are given in \cref{eqn:extendedMuSigmaCDC}.
\end{restatable}

The mean $\tmu^\cdc$ and covariance $\tSigma^\cdc$ used in \cref{thm:netsorMasterTheoremVarDim} are defined as follows.
\begin{defn}
For any CDC $\cdc$ and G-vars $g, \bar g$ in $\cdc$, define recursively
\begin{align*}
    \tmu^\cdc(g)
        &=
            \begin{cases}
            \mu^\cdc(g)  &   \text{if $g$ is input}\\
            0   &   \text{otherwise}
            \end{cases},
            \\
    \tSigma^\cdc(g, \bar g)
        &=
            \begin{cases}
            \Sigma^\cdc(g, \bar g)   &   \text{if $g, \bar g$ are inputs}\\
            \mathring \sigma^2_W \EV_Z \varphi^h(Z) \varphi^{\bar h}(Z) &   \text{if $g = Wh, \bar g=W\bar h$}\\
            0   &   \text{otherwise}
            \end{cases}
            \numberthis\label{eqn:extendedMuSigmaCDC}
\end{align*}
where $W$ can be either an input or a transposed A-var (with $\mathring\sigma_W^2$ defined in \cref{thm:netsorMasterTheoremVarDim}) and $Z \sim \Gaus(\tmu^{\cdc'}, \tSigma^{\cdc'})$ with $\cdc'$ denoting the CDC of $h$ and $\bar h$.
\end{defn}

\begin{proof}
    Trivial adaptation of the proof of \cref{thm:netsorTMasterTheoremFormal}.
\end{proof}

\begin{remk}
    \cref{eqn:netsorTVarDimLimit} only concerns G-vars of the same CDC; what about G-vars of different CDCs?
    One can quickly see that a convergence statement like in \cref{eqn:netsorTVarDimLimit} cannot be made naively just because the dimensions are not equal: we cannot even write down the ``empirical average'' that should converge.
    In fact, one can intuitively think of vectors of different CDCs as roughly ``independent'' because their ``main source of randomness'' must come from different A-vars.
\end{remk}

Likewise we can extend the BP-like \netsortplus{} Master Theorem to the Variable Dimension case.
\begin{thm}[BP-like \netsortplus{} Master Theorem; Variable Dimensions]
    \label{thm:selfParamNetsorplusMasterTheoremVarDim}
    Fix any BP-like \netsortplus{} program sampled in the natural way as in \cref{assm:netsorplusScalarLimit,assm:samplingVariableDim} and also satisfying rank stability (\cref{assm:asRankStab}).
    Let $\varphi^\bullet, \bigtheta^\bullet$ be as in \cref{defn:netsorplusMuSigma}.    
    Suppose for every H-var or C-var $u$, $\varphi^u(-; \bigvtheta^u)$ is parameter-controlled at $\mathring{\bigvtheta}^u$.

    Consider the limit where all dimensions go to infinity, with their pairwise ratios tending to finite but nonzero values:
    \[\text{for all }W : \Atype(n', n), \quad\text{we have}\quad n'/n \to \rho
    \quad\text{for some $\rho \in (0, \infty)$}.\]
    Then for each transpose $W^\trsp$ of an input A-var $W: \Atype(n', n)$, we have
    \[\sigma_{W^\trsp}^2 = \f{n'}{n} \sigma_W^2 \to \mathring \sigma_{W^\trsp}^2\]
    for some limit $\mathring \sigma_{W^\trsp}^2$.
    For each input A-var $W$, we also set $\mathring \sigma_W = \sigma_W$.

    For any pair of G-vars $g, \bar g$, we define recursively
    \begin{align*}
        \tmu(g)
            &\defeq
                \begin{cases}
                \muin(g)  &   \text{if $g$ is input}\\
                0   &   \text{otherwise}
                \end{cases}
                \\
        \tSigma(g, \bar g)
            &\defeq
                \begin{cases}
                \Sigmain(g, \bar g)   &   \text{if $g, g'$ are inputs}\\
                \mathring\sigma^2_W \EV_Z \varphi^h(Z; \mathring{\bigvtheta}^h) \varphi^{\bar h}(Z; \mathring{\bigvtheta}^{\bar h}) &   \text{if $g = Wh, \bar g=W\bar h$}\\
                0   &   \text{otherwise}
                \end{cases}
    \end{align*}
    where $W$ is any A-var, transposed or not;
    and for each C-var $\theta$ introduced by \ref{linetype:moment},
    \begin{align*}
    \mathring \theta \defeq
        \EV_Z \varphi^\theta(Z; \mathring {\bigtheta}^\theta).
        \numberthis\label{eqn:mathringthetaVarDim}
    \end{align*}

    Then the following hold.
    \begin{enumerate}
        \item
        For any CDC $\cdc$, let $g^1, \ldots, g^M$ be all of the G-vars in $\cdc$ (including all input G-vars).
        Then for any polynomially bounded $\psi: \R^M \to \R$, we have
        \begin{align*}
            \f 1 n \sum_{\alpha=1}^n \psi(g^1_\alpha, \ldots, g^M_\alpha) \asto \EV_{Z \sim \Gaus(\tmu^\cdc, \tSigma^\cdc)}\psi(Z).
        \end{align*}
        More generally, for any $l$, for any random vector $\bigtheta \in \R^l$ that converges almost surely to a deterministic vector $\mathring{\bigtheta}$, as $n \to \infty$, and for any $\psi(-; -): \R^M\times \R^l \to \R$ parameter-controlled at $\mathring{\bigtheta}$,
        \begin{align*}
            \f 1 n \sum_{\alpha=1}^n \psi(g^1_\alpha, \ldots, g^M_\alpha; \bigtheta) \asto \EV_{Z \sim \Gaus(\tmu^\cdc, \tSigma^\cdc)}\psi(Z; \mathring{\bigtheta}).
        \end{align*}
        \item
        Each C-var $\theta$ converges to its natural limit $\mathring \theta$:
        \begin{align*}
        \theta \asto \mathring \theta.
        \end{align*}
    \end{enumerate}
    
\end{thm}

\section{Writing Backpropagation of Standard Architectures in \texorpdfstring{\netsort{}}{NetsorT}}
\label{sec:BackpropInNetsorT}

In general, one can observe that, if the forward propagation can be written down in \netsor{} (which can be done for all standard architectures as noted in \citet{yang2019wide}), then the backprop can be written down in \netsort{}. Here we give some explicit examples of this.

\paragraph{Notation}
If $x \in \R^n$ is an (pre-)activation vector, then $dx$ denotes the gradient of the network output at $x$.

\paragraph*{Dense Matrix Multiplication}

If $y=Wx$, then $dx=W^{\trsp}dy$.

\paragraph*{Skip Connection}

If $z=x+y$, then $dx=dy=dz$.

\paragraph*{(Graph) Convolution}

A convolution can be decomposed as a sum of many weight-shared dense matrix multiplications, as observed in \citet{yang2019wide}. Combining the above, we can also express convolution and its backpropagation in \netsort{}.

Let $x=\{x_{s}\in\R^{n}:s\in P\}$ be the feature maps of a convolutional neural network, where $n$ is the number of channels, $P$ is the set of pixel positions (e.g. $P=[32]\times[32]$), and $x_{s}$ is the vector of activations at pixel $s$ across all channels. A convolutional layer is given by a set of weights $W=\{W_{\kappa}\in\R^{n\times n}:\kappa\in K\}$: a dense (\#channel-by-\#channel) matrix $W_{\kappa}$ for each kernel position $\kappa\in K$ (e.g. $K=\{-1,0,1\}\times\{-1,0,1\}$ for a $3\times3$ kernel or $K=\{-2,0,2\}\times\{-2,0,2\}$ for the same with dilation 2), where for simplicity we assume $n$ is also the number of output channels. Assume $W_{\kappa}\sim\Gaus(0,\sigma_{w}^{2}/n)$.
A kernel position $\kappa$ acts on a pixel position $s$ to obtain another pixel position $s+\kappa$.
The convolution of $x$ by $W$ (that maintains the pixel positions) can then be written via a combination of \textbf{\ref{instr:matmul}} and \textbf{\ref{instr:nonlin}} as $\{h_{s}\in\R^{n}:s\in P\}$ where
\[
h_{s}=\sum_{\kappa}W_{\kappa}x_{s+\kappa}
\]
where the range of $\kappa$ depends on the padding property of the convolution. Here we will assume the sum ranges over all $\kappa$ such that $s+\kappa\in P$, which corresponds to the most common zero padding.

Similarly, during backpropagation, given gradients $\{dh_{t}\in\R^{n}:t\in P\}$, the gradients $\{dx_{t}\in\R^{n}:t\in P\}$ can be computed by
\[
dx_{t}=\sum_{\kappa}W_{\kappa}^{\trsp}dh_{t-\kappa}
\]
where again the sum is over $\kappa$ such that $t+\kappa\in P$. Both equations are valid \netsort{} snippets and we may form the associated random variables by
\begin{align*}
Z^{h_{s}} & =\sum_{\kappa}Z^{W_{\kappa}x_{s+\kappa}}\\
Z^{dx_{t}} & =\sum_{\kappa}Z^{W_{\kappa}^{\trsp}dh_{t-\kappa}}
\end{align*}
where each $Z^{W_{\kappa}x_{s+\kappa}},Z^{W_{\kappa}^{\trsp}dh_{t-\kappa}}$ is Gaussian. 

Let $\bar{x}$ be any second set of input feature maps (possibly $x=\bar{x}$), and $\bar{h}$ is the convolution of $W$ with $\bar{x}$. Then $\{Z^{h_{s}}\}_{s}\cup\{Z^{\bar{h}_{s}}\}_{s}$ are jointly Gaussian, with
\[
\EV Z^{h_{s}}Z^{\bar{h}_{t}}=\sum_{\kappa,\tau}\EV Z^{W_{\kappa}x_{s+\kappa}}Z^{W_{\tau}\bar{x}_{t+\tau}}.
\]
But by Rule 2,
\[
\EV Z^{W_{\kappa}x_{s+\kappa}}Z^{W_{\tau}\bar{x}_{t+\tau}}=\begin{cases}
0 & \text{if \ensuremath{\kappa\ne\tau}}\\
\sigma_{w}^{2}\EV Z^{x_{s+\kappa}}Z^{\bar{x}_{t+\kappa}} & \text{if \ensuremath{\kappa=\tau}}
\end{cases}
\]
so we can simplify
\begin{equation}
\EV Z^{h_{s}}Z^{\bar{h}_{t}}=\sigma_{w}^{2}\sum_{\kappa}\EV Z^{x_{s+\kappa}}Z^{\bar{x}_{t+\kappa}}.\label{eqn:convForwardCov}
\end{equation}
Similarly, $\{Z^{dx_{t}}\}_{t}\cup\{Z^{d\bar{x}_{t}}\}_{t}$ is jointly Gaussian with
\begin{equation}
\EV Z^{dx_{s}}Z^{d\bar{x}_{t}}=\sigma_{w}^{2}\sum_{\kappa}\EV Z^{dh_{s+\kappa}}Z^{d\bar{h}_{t+\kappa}}.\label{eqn:convBackwardCov}
\end{equation}
If we apply nonlinearity to $h$, then the usual V-transform calculations apply. This routine can be easily generalized to different strides, paddings, dilations, and also to graph convolutions.

\paragraph{Pooling}
Continuing the notation from convolution above, global average pooling (GAP) can be expressed via \textbf{\ref{instr:nonlin}} as
\[\mathrm{GAP}(x) = \f 1 {|P|} \sum_{s\in P} x_s \in \R^n.\]
Likewise, (local) maxpool with kernel positions $K$ can be expressed via \textbf{\ref{instr:nonlin}} as 
\[\mathrm{Maxpool}(x)_s = \max\{x_{s+\kappa}: \kappa \in K, s+\kappa \in P\} \in \R^n,\]
where $\max$ is applied coordinatewise.

\paragraph*{Batchnorm and Pooling}

For $\epsilon>0$%
\footnote{
    If $\epsilon =0$ here, then the batchnorm jacobian has a singularity, so the BP-like Master Theorem does not cover it.
}, $\zeta=(\zeta^{1},\ldots,\zeta^{B})\in\R^{B}$, let $\tilde{\phi}:\R^{B}\to\R^{B},$ 
\begin{align}
\tilde{\phi}(\zeta) & \defeq\phi\left(\tilde{\zeta}\right),\tilde{\zeta}\defeq\f{\hat{\zeta}}{\sigma(\hat{\zeta})},\hat{\zeta}\defeq\zeta-\nu(\zeta) & \text{where}\quad\nu(\zeta)\defeq\f 1B\sum_{i=1}^{B}\zeta^{i},\quad\sigma(\hat{\zeta})^{2}\defeq\f 1B\|\hat{\zeta}\|^{2}+\epsilon,\numberthis\label{eqn:BN}
\end{align}
be batchnorm followed by coordinatewise nonlinearity $\phi$, where $\zeta\in\R^{B}$ should be interpreted as a single neuron across a batch, and $\nu$ and $\sigma$ are the \emph{batch} mean and standard deviations. Here, $B$ should be thought of as fixed while $n\to\infty$. 

If $\gamma=\tilde{\phi}(\zeta)\in\R^{B}$, and $d\gamma\in\R^{B}$ is a gradient of some loss wrt $\gamma$, then the gradient wrt $d\zeta\in\R^{B}$ can be written as
\[
d\zeta=d\tilde{\phi}(d\gamma \mid \zeta)\defeq\left(I-\frac{1}{B}\right)\left(I-\frac{\tilde{\zeta}\tilde{\zeta}^{\trsp}}{B}\right)\frac{d\gamma\odot\phi'(\tilde{\zeta})}{\sigma(\hat{\zeta})}.
\]
Then, given a batch of vectors $z^{1},\ldots,z^{B}\in\R^{n}$ (for example, they could be the preactivations after applying a linear layer), we can express batchnorm via \textbf{\ref{instr:nonlin}} as coordinatewise applications of $\tilde{\phi}$:
\begin{align}
y^{i} & :=\tilde{\phi}_{i}(z^{1},\ldots,z^{B})\in\R^{n},\quad i=1,\ldots,B.\label{eqn:BNnetsor}
\end{align}
Given gradients $dy^{1},\ldots,dy^{B}\in\R^{n}$, backpropagation can similarly be expressed via \textbf{\ref{instr:nonlin}} as coordinatewise applications of $d\tilde{\phi}$:
\[
dz^{i}=d\tilde{\phi}_{i}(dy^{1},\ldots,dy^{B} \mid z^{1},\ldots,z^{B})\in\R^{n},\quad i=1,\ldots,B.
\]

\paragraph*{GRU and LSTM}

Since GRU and LSTMs are just recurrent dense matrix mutliplication and coordinatewise nonlinearities, we can naturally write their forward and backprop in \netsort{}.
Here we do so concretely for GRU.

GRU evolves according to:
\begin{align*}
z^{t} & =\sigma(\zeta^{t}),\quad\zeta^{t}=U_{z}x^{t}+W_{z}h^{t-1}+b_{z}\\
r^{t} & =\sigma(\rho^{t}),\quad\rho^{t}=U_{r}x^{t}+W_{r}h^{t-1}+b_{r}\\
h^{t} & =z^{t}\odot h^{t-1}+(1-z^{t})\odot\phi(\gamma^{t}),\quad\gamma^{t}=U_{h}x^{t}+W_{h}(r^{t}\odot h^{t-1})+b_{h}
\end{align*}
where $\sigma$ is sigmoid and $\phi$ is tanh, $x^{t},h^{t},z^{t},r^{t}$ are resp. the input, state, update gate, and reset gate vectors, and $W_{\bullet},U_{\bullet},b_{\bullet}$ are the weights and biases going into vector $\bullet$. Since these equations only involve \textbf{\ref{instr:matmul}} and \textbf{\ref{instr:nonlin}}, they can be expressed in \netsort{}.

If the output is $v^{\trsp}h^{T}$ on the final time step $T$ for some weights $v\in\R^{n}$, and $d\bullet$ denotes the gradient of this output against $\bullet$, then we can write the backprop as follows
\begin{align*}
dh^{T} & =v\\
dh^{t-1} & =z^{t}\odot dh^{t}+W_{z}^{\trsp}d\zeta^{t}+r^{t}\odot W_{h}^{\trsp}d\gamma^{t}+W_{r}^{\trsp}\left(\sigma'(\rho^{t})\odot d\rho^{t}\right)\\
dz^{t} & =dh^{t}\odot\left(h^{t-1}-\phi(\gamma^{t})\right)\\
d\zeta^{t} & =dz^{t}\odot\sigma'(\zeta^{t})\\
d\gamma^{t} & =dh^{t}\odot(1-z^{t})\odot\phi'(\gamma^{t})\\
dr^{t} & =h^{t-1}\odot W_{h}^{\trsp}d\gamma^{t}\\
d\rho^{t} & =dr^{t}\odot\sigma'(\rho^{t})
\end{align*}
which involves only \textbf{\ref{instr:matmul}} and \textbf{\ref{instr:nonlin}} and so is expressible in \netsort{}.

To express layernorm and attention, we need the extension \netsortplus{} to \netsort{} so we can express scalars such as the mean and variance of a layer.
We will use the simplified version of \netsortplus{} given in \cref{defn:simpleNetsorTPlus}.

\newcommand{\Layernorm}{\operatorname{Layernorm}}
\paragraph*{Layernorm}

Given a layer's pre-activation $x\in\R^{n}$, we can use \textbf{\ref{instr:moment}} to compute its mean and variance, where we introduce nonlinearities $\phi_\bullet$ to put the expressions in the form of \textbf{\ref{instr:moment}}:
\begin{align*}
\nu & =\f 1n\sum_{\alpha=1}^{n}x_{\alpha}=\f 1n\sum_{\alpha=1}^{n}\phi_{id}(x_{\alpha})\in\R\\
\sigma^{2} & =\f 1n\sum_{\alpha=1}^{n}(x_{\alpha}-\nu)^{2}=\f 1n\sum_{\alpha=1}^{n}\phi_{sq}(x_{\alpha};\nu)\in\R.
\end{align*}
With these scalars defined, we can then express layernorm of $x$ as
\[
y=\Layernorm(x)=\text{\ensuremath{\frac{x-\nu}{\sqrt{\sigma^{2}+\epsilon}}}}=\phi_{LN}(x;\nu,\sigma^{2},\epsilon)\in\R^{n}.
\]
Now suppose we have a gradient $dy\in\R^{n}$ at $y$. Then backpropagating to $x$ through the layernorm yields
\begin{align*}
dx
&=d\Layernorm(dy \mid x) \in \R^n \numberthis\label{eqn:dLayernorm}\\
&\defeq \left(I_{n}-\frac{1}{n}\right)\left(I_{n}-\frac{yy^{\trsp}}{n}\right)\frac{dy}{\sqrt{\sigma^{2}+\epsilon}}\in\R^{n}\\
 & =\frac{1}{\sqrt{\sigma^{2}+\epsilon}}\left(I_{n}-\frac{1}{n}\right)\left(dy-c\cdot y\right)\\
 & =\frac{1}{\sqrt{\sigma^{2}+\epsilon}}\left(dy-c\cdot y-a\right)\\
 & =\phi_{bp}(dy,y;c,a,\sigma^{2},\epsilon) & \textbf{\ref{instr:nonlin+}}\\
\text{where}\quad c & =\frac{y^{\trsp}dy}{n}=\f 1n\sum_{\alpha=1}^{n}y_{\alpha}dy_{\alpha} \in \R & \textbf{\ref{instr:moment}}\\
a & =\f 1n\sum_{\alpha=1}^{n}dy_{\alpha}-c\cdot y_{\alpha} \in \R & \textbf{\ref{instr:moment}}
\end{align*}
In terms of the corresponding random variables, we have
\begin{align*}
    Z^y &= \mathring\Layernorm(x) 
        \defeq \f{Z^x - \EV Z^x}{\sqrt{\Var(Z^x) + \epsilon}}
        \numberthis\label{eqn:LayernormLimit}
        \\
    Z^{dx}
        &= d\mathring\Layernorm(dy \mid x)
        \defeq
            \f{\mathrm{Center}(Z^{dy} - Z^y \EV Z^{dy} Z^y)}{\sqrt{\Var(Z^x) + \epsilon}}
        \numberthis\label{eqn:dLayernormLimit}
\end{align*}
where $\mathrm{Center}(X) \defeq X - \EV X$.

\newcommand{\Attention}{\mathrm{Attn}}

\newcommand{\dvAttnLim}[1]{d_{\mathrm{v}#1}\mathring{\operatorname{Attn}}}
\newcommand{\dkAttnLim}[1]{d_{\mathrm{k}#1}\mathring{\operatorname{Attn}}}
\newcommand{\dqAttnLim}[1]{d_{\mathrm{q}#1}\mathring{\operatorname{Attn}}}

\paragraph*{Attention}

Given keys, queries, and values for $T$ tokens, $k^{1},\ldots,k^{T},q^{1},\ldots,q^{T},v^{1},\ldots,v^{T}\in\R^{n}$, attention yields
\begin{align*}
y^{i} 
  &= \Attention(q^i, \{k^i\}_i, \{v^i\}_i)\\
  &=a_{1}^{i}v^{1}+\cdots a_{T}^{i}v^{T}\in\R^{n}&\textbf{\ref{instr:nonlin+}}\\
\text{where \ensuremath{\quad}}(a_{1}^{i},\ldots,a_{T}^{i}) &=\mathrm{SoftMax}(c_{1}^{i},\ldots,c_{T}^{i})\\
a_{j}^{i} & =\mathrm{SoftMax}(c_{1}^{i},\ldots,c_{T}^{i})_{j}\in\R&\textbf{\ref{instr:moment}}\\
c_{j}^{i} & =q^{i\trsp}k^{j}/n=\frac{1}{n}\sum_{\alpha=1}^{n}q_{\alpha}^{i}k_{\alpha}^{j}\in\R. & \textbf{\ref{instr:moment}}
\end{align*}
In terms of corresponding random variables, we have
\begin{align*}
    Z^{y^i}
    &=
        \mathring a^i_1 Z^{v^1} + \cdots + \mathring a^i_T Z^{v^T}\\
    (\mathring a_{1}^{i},\ldots,\mathring a_{T}^{i})
    &=
        \mathrm{SoftMax}\lp \EV Z^{q^i}Z^{k^1}, \ldots, \EV Z^{q^i}Z^{k^T}\rp
        .
\end{align*}
If $dy^{1},\ldots,dy^{T}\in\R^{n}$ are gradients, and we abbreviate $dy = \{dy^i\}_i, k = \{k^i\}_i, q = \{q^i\}_i, v = \{v^i\}_i$, then backpropagating through the attention yields
\newcommand{\dvAttn}[1]{d_{\mathrm{v}#1}\operatorname{Attn}}
\newcommand{\dkAttn}[1]{d_{\mathrm{k}#1}\operatorname{Attn}}
\newcommand{\dqAttn}[1]{d_{\mathrm{q}#1}\operatorname{Attn}}
\begin{align*}
dv^{j}
    &= \dvAttn{j}(dy \mid k, q, v)
    \defeq a_{j}^{1}dy^{1}+\cdots+a_{j}^{T}dy^{T}
    &\text{\textbf{\ref{instr:nonlin+}}}\\
dq^{i} &= \dqAttn{i}(dy \mid k, q, v)
    \defeq\sum_{j,l}e_{j}^{i}f_{jl}^{i}k^{l}\in\R^{n}
    &\text{\textbf{\ref{instr:nonlin+}}}\\
dk^{i} &= \dkAttn{i}(dy \mid k, q, v)
    \defeq\sum_{j,l}e_{j}^{l}f_{ji}^{l}q^{l}\in\R^{n}
    &\text{\textbf{\ref{instr:nonlin+}}} \numberthis\label{dAttn}\\
\text{with}\quad e_{j}^{i} & =dy^{i\trsp}v^{j}/n=\frac{1}{n}\sum_{\alpha=1}^{n}dy_{\alpha}^{i}v_{\alpha}^{j}\in\R
    &\text{\textbf{\ref{instr:moment}}}\\
f_{jl}^{i} & =\partial a_{j}^{i}/\partial c_{l}^{i}=\f 1n\sum_{\alpha=1}^{n}\psi_{jl}(;c_{1}^{i},\ldots,c_{T}^{i})\in\R
    &\text{\textbf{\ref{instr:moment}}}
\end{align*}
where $\psi_{jl}$ is a ``parametrized nonlinearity'' that depends only on the parameters:
\begin{align*}
    \psi_{jl}(;c_1, \ldots, c_T) \defeq
        \partial \mathrm{SoftMax}_j(c_1, \ldots, c_T)
        /
        \partial c_l.
\end{align*}
If we abbreviate $Z^{dy} = \{Z^{dy^i}\}_i, Z^{k} = \{Z^{k^i}\}_i, Z^q = \{Z^{q^i}\}_i, Z^v=\{Z^{v^i}\}_i$, then
in terms of the corresponding random variables, we have
\begin{align*}
    Z^{dv^j}
    &=
        \dvAttnLim{j}(Z^{dy} \mid Z^k, Z^q, Z^v)
    \defeq
        \mathring a^1_j Z^{dy^1} + \cdots + \mathring a^T_{j} Z^{dy^T}\\
    Z^{dq^i}
    &=\dqAttnLim{i}(Z^{dy} \mid Z^k, Z^q, Z^v)
    \defeq
        \sum_{j,l} \mathring e^i_j \mathring f^i_{jl} Z^{k^l}
        \\
    Z^{dk^i}
    &=\dkAttnLim{i}(Z^{dy} \mid Z^k, Z^q, Z^v)
    \defeq
        \sum_{j,l} \mathring e^l_j \mathring f^l_{ji} Z^{q^l}
        \numberthis\label{eqn:dAttnLimit}
        \\
    \mathring e^i_j
    &=
        \EV Z^{dy^i} Z^{v^j}
        \\
    \mathring f^i_{jl}
    &=
        \psi_{jl}(;\EV Z^{q^i}Z^{k^1}, \ldots, \EV Z^{q^i}Z^{k^T})
        \\
\end{align*}

\section{Example NTK Computations}
\label{sec:exampleNTKComputation}

\newcommand{\relu}{\mathrm{relu}}
\newcommand{\step}{\mathrm{step}}
\newcommand{\erf}{\mathrm{erf}}

In this section, we show how to compute the NTK of different architecture.

First, we review the \emph{V-transform} of a nonlinearity.
\begin{defn}\label{defn:Vtransform}
Given a multivariate nonlinearity $\Phi: \R^B \to \R^B$, its \emph{V-transform} $\Vt\Phi$ is a function taking $B \times B$ positive semidefinite matrices to $B \times B$ positive semidefinite matrices, and is given by the following formula
\begin{equation*}
\Vt\Phi(K) \defeq \EV_{z \sim \Gaus(0, K)} \Phi(z) \Phi(z)^\trsp.
\end{equation*}
When $\phi: \R \to \R$, we take $\Vt\phi$ to be V-transform of the $\R^B \to \R^B$ function that applies $\phi$ to each coordinate.
\end{defn}

We collect below some of the common V-transforms.
Here we describe the V-transforms using the function notation of kernels, but we shall freely switch between the function notation and the matrix notation in what follows.

\begin{fact}[\cite{cho_kernel_2009}]
\label{fact:Vrelu}
For any kernel $K$,
\begin{align*}
\Vt\relu(K)(x, x')
    &=
        \f 1 {2\pi} (\sqrt{1-c^2} + (\pi - \arccos c) c) \sqrt{K(x, x) K(x', x')}
        \\
\Vt{\relu'}(K)(x, x')
    &=
        \f 1 {2\pi} (\pi - \arccos c)
        \\
\end{align*}
where $c = K(x, x')/\sqrt{K(x, x)K(x', x')}$.
\end{fact}

\begin{fact}[\cite{neal_bayesian_1995}]
\label{fact:Verf}
For any kernel $K$,
\begin{align*}
\Vt\erf(K)(x, x')
    &=
        \f 2 {\pi} \arcsin
        \f{K(x, x')}{\sqrt{(K(x, x)+ 0.5)(K(x', x')+0.5)}}
        \\
\Vt{\erf'}(K)(x, x')
    &=
        \f 4 { 
        \pi
        \sqrt{(1 + 2 K(x, x))(1 + 2 K(x', x')) - 4 K(x, x')^2}}
        .
\end{align*}
\end{fact}

\begin{fact}
\label{fact:Vexp}
Let $\phi(x) = \exp(x/\sigma)$ for some $\sigma > 0$.
For any kernel $K$,
\begin{align*}
\Vt\phi(K)(x, x')
    &=
        \exp\lp \f{K(x, x) + 2K(x, x') + K(x', x')}{2\sigma^2}\rp.
\end{align*}
\end{fact}

\subsection{MLP}

In the main text, we showed how to compute infinite-width NTK of an MLP using the simplified \netsort{} (\cref{defn:simpleNetsorT}).
While this is the recommended way of performing the calculations,
here we demonstrate formal \netsort{} (\cref{defn:netsort}) calculation of the infinite-width NTK $\mathring \NTK(x, x)$ of the MLP described in \cref{tp:MLP}, using \cref{thm:netsorTMasterTheoremFormal}.
By its nature, this calculation will be more verbose, so the meaning of \cref{thm:netsorTMasterTheoremFormal} can be seen concretely.
We hope this can help readers who find the main text calculations too dense.

For simplicity, let $\phi = \mathrm{ReLU}$, assume the hidden layer widths are equal to a common integer $n$, $n^1 = n^2 = n$, and suppose $x \in \R^m$.
This MLP has 5 parameters: $W^1 \in \R^{n \times m}, W^2 \in \R^{n \times n}, v \in \R^{n}, b^1 \in \R^{n}, b^2 \in \R^{n}$.
In the NTK parametrization, we factor $W^1 = \f 1 {\sqrt m} \omega^1$ and $W^2 = \f 1 {\sqrt {n}} \omega^2$, and we sample $\omega^1_{\alpha \beta}, \omega^2_{\alpha \beta}, v_{\alpha}, b^1_{\alpha}, b^2_{\alpha} \sim \Gaus(0, 1)$, iid, for any $\alpha, \beta$.
This implies that $\sigma_{W^2} = 1$ in \cref{assm:equalDimSampling}.

This implies that each coordinate of the G-var vector $W^1 x$ is distributed as $\Gaus(0, \|x\|^2/m)$.
Thus, $\muin$ is identically 0, and $\Sigmain$ takes the following values over pairs of G-vars
\begin{align*}
\Sigmain(W^1 x, W^1 x) = \|x\|^2/m,\quad
\Sigmain(b^1, b^1) = \Sigmain(b^2, b^2) = \Sigmain(v, v) = 1,
\end{align*}
and $\Sigmain(g, g') = 0$ for all other pairs of G-vars.

If we let $f(x)$ denote the network output $v^\trsp x^2 / \sqrt{n}$, then by \cref{eqn:NTKsimplify}, the contribution of $\omega^1$'s gradient to the NTK is
\begin{align*}
\|\nabla_{\omega^1} f(x)\|^2
    &=
        \|\nabla_{W^1 x}f(x)\|^2 \f{\|x\|^2}{m}
        .
\end{align*}
In \cref{tp:MLP}, the G-var $d(W^1 x)$ corresponds to $\sqrt{n} \nabla_{W^1 x}f(x)$.
Therefore, we can rewrite the above as
\begin{align*}
\|\nabla_{\omega^1} f(x)\|^2
    = \f{\|d(W^1 x)\|^2}{n} \f{\|x\|^2}{m}.
\end{align*}

Similarly, the contribution of $\omega^2$'s and $v$'s gradients to the NTK is
\begin{align*}
\|\nabla_{\omega^2} f(x)\|^2
    = \f{\|d\tilde h^2\|^2}{n} \f{\|x^1\|^2}{n},\quad
\|\nabla_{v} f(x)\|^2
    = \f{\|x^2\|^2}{n}.
\end{align*}

Likewise, the contributions of the bias gradients are
\begin{align*}
\|\nabla_{b^1} f(x)\|^2
    = \f{\|d(W^1 x)\|^2}{n},\quad
\|\nabla_{b^2} f(x)\|^2
    = \f{\|d\tilde h^2\|^2}{n}.
\end{align*}

Since the NTK can be expressed as
\begin{align*}
\NTK(x, x)
    &=
        \|\nabla_{\omega^1} f(x)\|^2 + \|\nabla_{\omega^2} f(x)\|^2 + \|\nabla_{v} f(x)\|^2 + \|\nabla_{b^1} f(x)\|^2 + \|\nabla_{b^2} f(x)\|^2\\
    &=
        \f{\|d(W^1 x)\|^2}{n} \f{\|x\|^2}{m}
        + \f{\|d\tilde h^2\|^2}{n} \f{\|x^1\|^2}{n}
        + \f{\|x^2\|^2}{n}
        + \f{\|d(W^1 x)\|^2}{n}
        + \f{\|d\tilde h^2\|^2}{n}\\
    &=
        \f{\|d(W^1 x)\|^2}{n} \lp \f{\|x\|^2}{m} + 1 \rp
        + \f{\|d\tilde h^2\|^2}{n} \lp \f{\|x^1\|^2}{n} + 1 \rp
        + \f{\|x^2\|^2}{n}
        \numberthis
        \label{eqn:MLPNTKExpression}
        ,
\end{align*}
it suffices to compute the limits of the following squared norms, as $n \to \infty$:
\begin{align*}
\f{\|d(W^1 x)\|^2}{n},
\f{\|d\tilde h^2\|^2}{n},
\f{\|x^1\|^2}{n},
\f{\|x^2\|^2}{n},
\f{\|x\|^2}{m}.
\end{align*}
\cref{thm:netsorTMasterTheoremFormal} provides exactly the tool needed for this purpose.
Here the last squared norm $\|x\|^2/m$ is constant in $n$ so we will focus on the other ones.

\paragraph{Checking the conditions of \cref{thm:netsorTMasterTheoremFormal}}
In order to apply \cref{thm:netsorTMasterTheoremFormal}, we first need to check that 1) \cref{tp:MLP} is BP-like, and 2) its nonlinearities are polynomially-bounded.
The latter assumption is obvious since both ReLU and its derivative, the step function, are polynomially-bounded (note that we don't require these functions to be smooth at all).
The former condition is already shown in \cref{remk:backpropIsBPLike} to be true for any program expressing backpropagation, but we can also reason explicitly as follows:
We can take the ``special set of input G-vars'' in \cref{defn:BPlike} to be the G-var $v$ in \cref{tp:MLP}.
Note that the only input A-var in \cref{tp:MLP} is $W^2$.
Then condition 2 of \cref{defn:BPlike} is satisfied because the only usage of $W^2$ in \cref{tp:MLP} is in the line $\tilde h^2 := W^2 x^1$, and here $x^1$ does not depend on $v$.
Likewise, condition 1 is satisfied because the only usage of $W^2{}^\trsp$ in \cref{tp:MLP} is in the line $d x^1 := W^2{}^\trsp d \tilde h^2$, and $d\tilde h^2$ depends linearly on, and is thus odd in $v$.

\paragraph{Limits of $x^1$ and $x^2$}
In fact, the limits of $\f{\|x^1\|^2}{n},
\f{\|x^2\|^2}{n}$ can be computed already with the \netsor{} Master Theorem of \citet{yang2019wide}, but for completeness we will present the calculation of their limits.

The variable $x^1$ has type H but it can be expressed as a function of G-vars as $\phi(W^1 x + b^1)$, so by \cref{thm:netsorTMasterTheoremFormal},
\begin{align*}
\f{\|x^1\|^2}{n} = \f 1 n \sum_{\alpha=1}^n \phi((W^1 x)_\alpha + b^1_\alpha)^2
\asto
    \EV_{Z^{W^1 x}, Z^{b^1}} \phi(Z^{W^1 x} + Z^{b^1})^2\\
\quad\text{where}\quad(Z^{W^1 x}, Z^{b^1}) \sim \Gaus\left(0, \begin{pmatrix}
\Sigma(W^1 x, W^1 x) & \Sigma(W^1 x, b^1)\\
\Sigma(W^1 x, b^1) & \Sigma(b^1, b^1)
\end{pmatrix}
\right).
\end{align*}
Because $W^1 x$ and $b^1$ are both input G-vars, this covariance matrix is just
\begin{align*}
\begin{pmatrix}
\Sigmain(W^1 x, W^1 x) & \Sigmain(W^1 x, b^1)\\
\Sigmain(W^1 x, b^1) & \Sigmain(b^1, b^1)
\end{pmatrix}
=
\begin{pmatrix}
\|x\|^2/m & 0\\
0 & 1
\end{pmatrix}.
\end{align*}
Furthermore, by linearity of Gaussian variables, we can simplify this expectation to
\begin{align*}
\f{\|x^1\|^2}{n} \asto \EV_\zeta \phi(\zeta)^2, \quad\text{where}\quad \zeta &\sim \Gaus(0, \Sigma(W^1 x, W^1 x)+ 2\Sigma(W^1 x, b^1) + \Sigma(b^1, b^1))\\
&= \Gaus\left(0, \f{\|x\|^2}{m}+1\right)
.
\end{align*}
Since we have assumed $\phi$ is ReLU, this expectation is just
\begin{align*}
\f{\|x^1\|^2}{n} \asto \f 1 2 \lp \f{\|x\|^2}{m}+1\rp = \f 1 2 \f{\|x\|^2}{m} + \f 1 2.
\end{align*}

To compute the the next limit $\lim_{n \to \infty} \|x^2\|^2/n$, we first need to compute $\Sigma(\tilde h^2,\tilde h^2)$ and $\Sigma(\tilde h^2, b^2)$.
By ``otherwise'' case of \cref{eqn:extendedMuSigma}, $\Sigma(\tilde h^2, b^2) = 0$, and by the \ref{linetype:MatMul} case of \cref{eqn:extendedMuSigma},
\begin{align*}
\Sigma(\tilde h^2,\tilde h^2) = \sigma_{W^2}\EV_{Z^{W^1 x}, Z^{b^1}} \phi(Z^{W^1 x} + Z^{b^1})^2 = \f 1 2 \f{\|x\|^2}{m} + \f 1 2,
\end{align*}
as we have computed above already.

Therefore, by \cref{fig:mastertheoremIllustration},
\begin{align*}
\|x^2\|^2/n &= \f 1 n \sum_{\alpha=1}^n \phi(\tilde h^2_\alpha + b^2_\alpha)^2
\asto
\EV \phi(Z^{\tilde h^2} + Z^{b^2})^2,\\
\quad\text{where}\quad
(Z^{\tilde h^2}, Z^{b^2})
&\sim
\Gaus\left(
0,
\begin{pmatrix}
\Sigma(\tilde h^2, \tilde h^2)
    & \Sigma(\tilde h^2, b^2)\\
\Sigma(\tilde h^2, b^2)
    & \Sigma(b^2, b^2)
\end{pmatrix}
\right)
= \Gaus\left(0,
\begin{pmatrix}
\f 1 2 \f{\|x\|^2}{m} + \f 1 2
    & 0\\
0
    & 1
\end{pmatrix}
\right).
\end{align*}
Again, by linearity of Gaussians, we have
\begin{align*}
\|x^2\|^2/n
    \asto
        \EV_{\zeta \sim \Gaus(0, \|x\|^2/2m + 3/2)} \phi(\zeta)
        = \f{\|x\|^2}{4m} + \f 3 4.
\end{align*}

\paragraph{Limits of 
$\f{\|d(W^1 x)\|^2}{n}$ and $\f{\|d\tilde h^2\|^2}{n}$}

Whereas the limits computed above could already be done using the \netsor{} Master Theorem of \citet{yang2019wide}, the limits we will compute here necessarily involve matrix transposes and thus can only be computed using \cref{thm:netsorTMasterTheoremFormal}.

By \cref{thm:netsorTMasterTheoremFormal},
\begin{align*}
\|d \tilde h^2\|^2/n = \f 1 n \sum_{\alpha=1}^n (\phi'(\tilde h^2_\alpha + b^2_\alpha) dx^2_\alpha)^2
    \asto
        \EV \phi'(Z^{\tilde h^2} + Z^{b^2})^2 (Z^{dx^2})^2
\end{align*}
where
\begin{align*}
(Z^{\tilde h^2}, Z^{b^2}, Z^{dx^2})
&\sim
    \Gaus\left(0,
    \begin{pmatrix}
    \Sigma(\tilde h^2, \tilde h^2)
        & \Sigma(\tilde h^2, b^2)
            & \Sigma(\tilde h^2, dx^2) \\
    \Sigma(b^2, \tilde h^2)
        & \Sigma(b^2, b^2)
            & \Sigma(b^2, dx^2) \\
    \Sigma(dx^2, \tilde h^2)
        & \Sigma(dx^2, b^2)
            & \Sigma(dx^2, dx^2)
    \end{pmatrix}
    \right)\\
&=
    \Gaus\left(0,
        \begin{pmatrix}
        \f 1 2 \|x\|^2/m + \f 1 2
            & 0
                & 0 \\
        0
            & 1
                & 0 \\
        0
            & 0
                & 1
        \end{pmatrix}
    \right).
\end{align*}

Since $Z^{dx^2}$ is independent from $Z^{\tilde h^2}$ and $Z^{b^2}$, we have
\begin{align*}
\|d \tilde h^2\|^2/n
&\asto
    \EV \phi'(\zeta_1)^2 (\zeta_2)^2,
    \quad\text{with}\quad
    \zeta_1 \sim \Gaus\left(0, \f 1 2 \|x\|^2/m + \f 3 2\right),\quad
    \zeta_2 \sim \Gaus(0, 1)\\
&=
    \EV \phi'(\zeta_1)^2 \EV (\zeta_2)^2
=
    \f 1 2 \cdot 1
=
    \f 1 2.
\end{align*}

Next, notice that by \cref{eqn:extendedMuSigma}, for the same $\zeta_1, \zeta_2$ above, we have
\begin{align*}
\Sigma(dx^1, dx^1)
=
    \sigma_{W^2{}^\trsp} \EV \phi'(\zeta_1)^2 (\zeta_2)^2
=
    1 \cdot \f 1 2
=
    \f 1 2 
\end{align*}
as before (in this calculation, the \ref{linetype:MatMul} case of \cref{eqn:extendedMuSigma} essentially ``forgets'' the correlation between $W^2$ and $W^2{}^\trsp$ and treats $W^2{}^\trsp$ as just another independently sampled matrix).
In addition, $\Sigma(dx^1, b^1) = \Sigma(dx^1, W^1 x) = 0$ by the ``otherwise'' case of \cref{eqn:extendedMuSigma}.
Consequently, by \cref{thm:netsorTMasterTheoremFormal},
\begin{align*}
\|d(W^1 x)\|^2/n
&=
    \f 1 n \sum_{\alpha=1}^n
        \phi'((W^1 x)_\alpha + b^1_\alpha)^2 (dx^1)_\alpha^2
 \asto
    \EV \phi'(Z^{W^1 x} + Z^{b^1})^2 (Z^{dx^1})^2,\\
\quad\text{where}\quad
(Z^{W^1x}, Z^{b^1}, Z^{dx^2})
&\sim
\Gaus\lp
0,
\begin{pmatrix}
\Sigma(W^1 x, W^1 x)
    & \Sigma(W^1 x, b^1)
        & \Sigma(W^1 x, dx^2)\\
\Sigma(b^1, W^1 x)
    & \Sigma(b^1, b^1)
        & \Sigma(b^1, dx^2)\\
\Sigma(dx^2, W^1 x)
    & \Sigma(dx^2, b^1)
        & \Sigma(dx^2, dx^2)
\end{pmatrix}
\rp\\
&=
\Gaus\lp 0,
\begin{pmatrix}
\|x\|^2/m
    & 0
        & 0\\
0
    & 1
        & 0\\
0
    & 0
        & \f 1 2
\end{pmatrix}
\rp.
\end{align*}
This expectation is easily evaluated, and we have
\begin{align*}
    \|d(W^1 x)\|^2/n \asto \f 1 2 \cdot \f 1 2 = \f 1 4
\end{align*}

\paragraph{Finish computing the infinite-width NTK $\mathring \NTK$}

In summary, we have
\begin{align*}
\f{\|x^1\|^2}n \asto \f 1 2 \f{\|x\|^2}m + \f 1 2,\quad
\f{\|x^2\|^2}n \asto \f 1 4 \f{\|x\|^2}m + \f 3 4,\quad
\f{\|d\tilde h^2\|^2}n \asto \f 1 2,\quad
\f{\|d(W^1 x)\|^2}n \asto \f 1 4.
\end{align*}
Thus, by \cref{eqn:MLPNTKExpression}, we have
\begin{align*}
\NTK(x, x) \asto \mathring \NTK(x, x) &=
\f 1 4 \lp \f{\|x\|^2}m + 1 \rp
+ \f 1 2 \lp \f 1 2 \f{\|x\|^2}m + \f 1 2 + 1 \rp
+ \lp\f 1 4 \f{\|x\|^2}m + \f 3 4\rp\\
&=
    \f 3 4 \f{\|x\|^2}m + \f 7 4.
\end{align*}

\paragraph{Generalization to multiple inputs}
This example only computed $\mathring \NTK(x, x)$.
The same reasoning can be easily applied to multiple inputs by writing down a program expressing the forward and backward computation of the MLP on two inputs.

\subsection{Simple Recurrent Neural Network and Average Pooling}
\label{sec:simpleRNNCalc}

We complete the NTK limit computation from the main text and also generalize it to the case where the output is a projection of the average state instead of just the last state.
Recall the RNN we consider is given by the following forward and backward equations
\begin{align*}
s^{t}(\xi) & =\phi(g^{t}(\xi)+u^{t}(\xi)+b),\quad g^{t}(\xi)=Ws^{t-1}(\xi),\quad u^{t}(\xi)=U\xi^{t}\\
ds^{t-1} & =W^{\trsp}dg^{t},\quad du^t = dg^{t}=\phi'\left(g^{t}+u^{t}+b\right)\odot ds^{t}.
\end{align*}
For an input sequence $\xi=\left\{ \xi^{1},\ldots,\xi^{t},\ldots,\xi^{T}\in\R^{d}\right\} $, we will consider both
\begin{description}
    \item[LastState] the case where the output of the RNN is a projection of the last state
    \[
    f(\xi)=v^{\trsp}s^{T}/\sqrt{n}
    \]
    as in main text and
    \item[AvgPool] the case where the output of the RNN is a projection of the average state
    \[
    f(\xi)=\frac{1}{T}\sum_{t=1}^{T}v^{\trsp}s^{t}/\sqrt{n}.
    \]
\end{description}
If we sample
\[
W_{\alpha\beta}\sim\Gaus(0,\sigma_{W}^{2}/n),U_{\alpha\beta}\sim\Gaus(0,\sigma_{U}^{2}/d),b_{\alpha}\sim\Gaus(0,\sigma_{b}^{2}),v_{\alpha}\sim\Gaus(0,\sigma_{v}^{2})
\]
then the recursion equations in the main text can be generalized straightforwardly to the following
\begin{align*}
C^{s^{t},\bar{s}^{r}} & =\EV\phi(\zeta_{1})\phi(\zeta_{2}),\\
D^{s^{t},\bar{s}^{r}} & =D^{s^{t+1},\bar{s}^{r+1}}\EV\phi'(\zeta_{1})\phi'(\zeta_{2})\\
D^{g^{t},\bar{g}^{r}} & =D^{u^{t},\bar{u}^{r}}=\sigma_{W}^{-2}D^{s^{t-1},\bar{s}^{r-1}}
\end{align*}
where $(\zeta_{1},\zeta_{2})\sim\Gaus\left(\sigma_{W}^{2}\begin{pmatrix}C^{s^{t},s^{t}} & C^{s^{t},\bar{s}^{r}}\\
C^{s^{t},\bar{s}^{r}} & C^{\bar{s}^{r},\bar{s}^{r}}
\end{pmatrix}+\sigma_{U}^{2}\frac{\xi^{t\trsp}\xi^{r}}{d}+\sigma_{b}^{2}\right)$, and we have abbreviated $C^{s^{t},\bar{s}^{r}}=C^{s^{t},\bar{s}^{r}}(\xi,\bar{\xi})= \EV Z^{s^t}Z^{\bar s^r} = \lim_{n\to\infty} \inv n s^{t\trsp} \bar s^r$, and so on.

\paragraph*{Initial condition for LastState}

If we use the last state for output, then the initial conditions are 
\begin{align*}
C^{s^{0},\bar{s}^{r}} & =C^{s^{t},\bar{s}^{0}}=0\\
D^{s^{T},\bar{s}^{\bar{T}}} & =\sigma_{v}^{2}\quad\quad\text{but}\\
D^{s^{T},\bar{s}^{r}} =D^{s^{t},\bar{s}^{\bar{T}}}&=0,\quad\quad\text{for all other $r,t$.}
\end{align*}
Here, the initial condition for $D^{s^{t},\bar{s}^{r}}$ reflects the fact that only the last state is used for output. This initial condition in fact implies that most $D^{s^{t},\bar{s}^{r}}$ are 0 by a simple induction:
\[
i\ne j\implies D^{s^{-i},\bar{s}^{-j}}=0
\]
where $s^{-i}=s^{T-i},\bar{s}^{-j}=\bar{s}^{\bar{T}-j}$.

\paragraph*{Initial condition for AvgPool}

If instead of projecting the last state, we project the average of all states to get the output, then the initial condition for $D$ is
\[
D^{s^{T},\bar{s}^{r}}=D^{s^{t},\bar{s}^{\bar{T}}}=\sigma_{v}^{2},\quad\forall r,t
\]
In this case we can't zero out the majority of $D$ like the above.

\paragraph*{NTK}

To compute the NTK, we apply \cref{eqn:NTKlimit} to get
\[
\mathring{\NTK}(\xi, \bar \xi)=
\sum_{t=1}^{T-1}\sum_{r=1}^{\bar T-1}
    D^{g^{t+1},\bar{g}^{r+1}}C^{s^{t},\bar{s}^{r}}
+\sum_{t=1}^{T}\sum_{r=1}^{\bar T}
    D^{g^{t},\bar{g}^{r}}\frac{\xi^{t}\bar{\xi}^{r}}{d}
+\sum_{t=1}^{T}\sum_{r=1}^{\bar T}D^{g^{t},\bar{g}^{r}}+C^{s^{T},\bar{s}^{T}}
\]
where the terms in the sum are resp. contributions from $W,U,b,$ and $v$.
As noted above, if the output depends only on the last state, then the double sum above can be replaced with a single sum over the diagonal $D^{s^{-i}, \bar s^{-i}}$.

\subsection{Convolution Neural Network}

We continue the notation of the \textbf{Convolution} section of \cref{sec:BackpropInNetsorT}. We consider a convolutional neural network with width $n$, with feature map positions given by $P$, and with convolutional kernel positions $K$, throughout the network. For example, $K$ can be $\{-1,0,1\}\times\{-1,0,1\}$, and $P$ can be $[32]\times[32]$. For simplicity, we forgo bias. Let $x^{l}=\{x_{s}^{l}\in\R^{n}:s\in P\}$ and $h^{l}=\{h_{s}^{l}\in\R^{n}:s\in P\}$ be the activations and preactivations of layer $l$. Let $W^{l}=\{W_{\kappa}^{l}:\kappa\in K\}$ be the layer $l$ weights. The input $\xi=\{\xi_{s}\in\R^{d}:s\in P\}$ to the network is an image with $d=3$ channels and has pixel positions $P$ . Then the network computation proceeds by
\begin{align*}
x_{s}^{0}(\xi) & =\xi_{s},&
h_{s}^{l}(\xi) & =\sum_{\kappa}W_{\kappa}^{l}x_{s+\kappa}^{l-1}(\xi),&
x_{s}^{l}(\xi) & =\phi(h_{s}^{l}(\xi)),
\end{align*}
for $l=1,\ldots,L$, where the sum is over $\kappa$ such that $s+\kappa\in P$. We sample $W_{\kappa\alpha\beta}^{l}\sim\Gaus(0,\sigma_{w}^{2}/n)$ for $l=2,\ldots,L$, and $W_{\kappa\alpha\beta}^{1}\sim\Gaus(0,\sigma_{w}^{2}/d)$. For the output, we will consider both
\begin{description}
    \item[Global Average Pooling (GAP)] where output of network is given by 
    \[
    f(\xi)=\frac{1}{|P|}\sum_{s\in P}\frac{1}{\sqrt{n}}v^{\trsp}x_{s}^{L}
    \]
    for $v_{\alpha}\sim\Gaus(0,\sigma_{v}^{2})$ and
    \item[Vectorization]    where output of network is given by
    \[
    f(\xi)=\frac{1}{\sqrt{|P|}}\sum_{s\in P}\frac{1}{\sqrt{n}}v_{s}^{\trsp}x_{s}^{L}
    \]
    for output weights $v=\{v_{s}\in\R^{n}:s\in P\}$ sampled like $v_{s\alpha}\sim\Gaus(0,\sigma_{v}^{2})$. This is called ``vectorization'' because it's equivalent to a linear readout of the flattening (vectorization) of final layer embeddings $x^{L}=\{x_{s}^{L}:s\in P\}.$
\end{description}

\paragraph*{Forward Propagation}

Given another input $\bar{\xi}=\{\bar{\xi}_{s}\in\R^{d}:s\in P\}$, we define $\bar{h}_{s}^{l}=h_{s}^{l}(\bar{\xi}),\bar{x}_{s}^{l}=x_{s}^{l}(\bar{\xi})$ similarly. Then by the \netsort{} Master Theorem (\cref{thm:netsorTMasterTheoremSimple}), there exist some deterministic scalars $C^{h_{s}^{l},\bar{h}_{t}^{l}}(\xi,\bar{\xi}),C^{x_{s}^{l},\bar{x}_{t}^{l}}(\xi,\bar{\xi})$, such that
\begin{align*}
h_{s}^{l\trsp}\bar{h}_{t}^{l}/n & \asto\EV Z^{h_{s}^{l}}Z^{\bar{h}_{t}^{l}}\defeq C^{h_{s}^{l},\bar{h}_{t}^{l}}(\xi,\bar{\xi})\\
x_{s}^{l\trsp}\bar{x}_{t}^{l}/n & \asto\EV Z^{x_{s}^{l}}Z^{\bar{x}_{t}^{l}}\defeq C^{x_{s}^{l},\bar{x}_{t}^{l}}(\xi,\bar{\xi}),
\end{align*}
for all $s,t\in P,l=1,\ldots,L$.
For convenience, we also define $C^{x_{s+\kappa}^{0},\bar{x}_{t+\kappa}^{0}}(\xi,\bar{\xi})\defeq{x_{s+\kappa}^{0\trsp}\bar{x}_{t+\kappa}^{0}}/{d}$. 
As in \cref{eqn:convForwardCov} and in the MLP case, these scalars are related to each other through a recurrence: for $l=1,\ldots,L$,
\begin{align*}
    C^{x_{s}^{l},\bar{x}_{t}^{l}}(\xi,\bar{\xi})
    &=
        \EV\phi(\zeta)\phi(\bar{\zeta}),\quad(\zeta,\bar{\zeta})\sim\begin{pmatrix}C^{h_{s}^{l},h_{s}^{l}}(\xi,\xi) & C^{h_{s}^{l},\bar{h}_{t}^{l}}(\xi,\bar{\xi})\\
        C^{h_{s}^{l},\bar{h}_{t}^{l}}(\xi,\bar{\xi}) & C^{\bar{h}_{t}^{l},\bar{h}_{t}^{l}}(\bar{\xi},\bar{\xi})
        \end{pmatrix}\\
    C^{h_{s}^{l},\bar{h}_{t}^{l}}(\xi,\bar{\xi})
    &=
        \sigma_{w}^{2}\sum_{\kappa}C^{x_{s+\kappa}^{l-1},\bar{x}_{t+\kappa}^{l-1}}(\xi,\bar{\xi}).
\end{align*}

\paragraph*{Backpropagation}

Define $dh_{s}^{l}\defeq\sqrt{n}\nabla_{h_{s}^{l}}f(\xi),d\bar{h}_{s}^{l}\defeq\sqrt{n}\nabla_{\bar{h}_{s}^{l}}f(\bar{\xi})$. Then, the backpropagation of $f$ proceed as follows.
\begin{align*}
dh_{s}^{l} & =dx_{s}^{l}\odot\phi'(h_{s}^{l})\\
dx_{s}^{l-1} & =\sum_{\kappa}W_{\kappa}^{l\trsp}dh_{s-\kappa}^{l}.
\end{align*}
By the \netsort{} Master Theorem (\cref{thm:netsorTMasterTheoremSimple}), there exist some deterministic scalars $D^{h_{s}^{l},\bar{h}_{t}^{l}}(\xi,\bar{\xi}),D^{x_{s}^{l},\bar{x}_{t}^{l}}(\xi,\bar{\xi})$ such that
\begin{align*}
dh_{s}^{l\trsp}d\bar{h}_{t}^{l}/n & \asto\EV Z^{dh_{s}^{l}}Z^{d\bar{h}_{t}^{l}}\defeq D^{h_{s}^{l},\bar{h}_{t}^{l}}(\xi,\bar{\xi})\\
dx_{s}^{l\trsp}d\bar{x}_{t}^{l}/n & \asto\EV Z^{dx_{s}^{l}}Z^{d\bar{x}_{t}^{l}}\defeq D^{x_{s}^{l},\bar{x}_{t}^{l}}(\xi,\bar{\xi}),
\end{align*}
for $l=L,\ldots,1$.
These scalars are related to each other through a recurrence:
As in the MLP case, we have, for $l=L,\ldots,1$,
\[
D^{h_{s}^{l},\bar{h}_{t}^{l}}(\xi,\bar{\xi})=D^{x_{s}^{l},\bar{x}_{t}^{l}}(\xi,\bar{\xi})\EV\phi'(\zeta)\phi'(\bar{\zeta}),\quad(\zeta,\bar{\zeta})\sim\begin{pmatrix}C^{h_{s}^{l},h_{s}^{l}}(\xi,\xi) & C^{h_{s}^{l},\bar{h}_{t}^{l}}(\xi,\bar{\xi})\\
C^{h_{s}^{l},\bar{h}_{t}^{l}}(\xi,\bar{\xi}) & C^{\bar{h}_{t}^{l},\bar{h}_{t}^{l}}(\bar{\xi},\bar{\xi})
\end{pmatrix}.
\]
As in \cref{eqn:convBackwardCov}, we also have, for $l=L,\ldots,2$,
\[
D^{x_{s}^{l-1},\bar{x}_{t}^{l-1}}(\xi,\bar{\xi})=\sigma_{w}^{2}\sum_{\kappa}D^{h_{s-\kappa}^{l},\bar{h}_{t-\kappa}^{l}}.
\]

Now the initial condition for this recurrence depends on whether the network has global average pooling or not:
\[
D^{x_{s}^{L},\bar{x}_{t}^{L}}(\xi,\bar{\xi})=\begin{cases}
\sigma_{v}^{2}/|P|^{2} & \text{if GAP}\\
\sigma_{v}^{2}\ind(s=t)/|P| & \text{if vectorization.}
\end{cases}
\]
Note in the vectorization case, this initial condition implies that, for all $l$, $D^{x_{s}^{l},\bar{x}_{t}^{l}}(\xi,\bar{\xi}) = D^{x_{s}^{l},\bar{x}_{t}^{l}}(\xi,\bar{\xi}) = 0$ if $s\ne t$.

\paragraph*{NTK}

Finally, we can decompose the NTK into contributions from $\nabla_{v}f$ and from $\{\nabla_{\omega_{\kappa}^{l}}f\}_{\kappa,l}$.
If the last layer involves GAP, then the contribution of $\nabla_v f$ is
\begin{align*}
\langle\nabla_{v}f(\xi),\nabla_{v}f(\bar{\xi})\rangle & =\frac{1}{n}\left\langle |P|^{-1}\sum_{s\in P}x_{s}^{L},|P|^{-1}\sum_{t\in P}\bar{x}_{t}^{L}\right\rangle
 =|P|^{-2}\sum_{s,t\in P}\frac{x_{s}^{L\trsp}\bar{x}_{t}^{L}}{n}\\
 & \asto|P|^{-2}\sum_{s,t\in P}C^{x_{s}^{L},\bar{x}_{t}^{L}}(\xi,\bar{\xi}).
\end{align*}
If the last layer involves vectorization, then
\begin{align*}
\langle\nabla_{v}f(\xi),\nabla_{v}f(\bar{\xi})\rangle & =\sum_{s\in P}\left\langle \nabla_{v_{s}}f(\xi),\nabla_{v_{s}}f(\bar{\xi})\right\rangle
 =|P|^{-1}\sum_{s\in P}\frac{x_{s}^{L\trsp}\bar{x}_{s}^{L}}{n}\\
 & \asto|P|^{-1}\sum_{s\in P}C^{x_{s}^{L},\bar{x}_{s}^{L}}(\xi,\bar{\xi}).
\end{align*}
For $l>1$, the contribution of $\nabla_{\omega_\kappa^l} f$ is
\begin{align*}
\langle\nabla_{\omega_{\kappa}^{l}}f(\xi),\nabla_{\omega_{\kappa}^{l}}f(\bar{\xi})\rangle & =n^{-2}\left\langle \sum_{s}dh_{s}^{l}x_{s+\kappa}^{l-1},\sum_{s}d\bar{h}_{s}^{l}\bar{x}_{s+\kappa}^{l-1}\right\rangle
 =\sum_{s,t}\frac{dh_{s}^{l\trsp}d\bar{h}_{t}^{l}}{n}\frac{x_{s+\kappa}^{l-1\trsp}\bar{x}_{t+\kappa}^{l-1}}{n}\\
 & \asto D^{h_{s}^{l},\bar{h}_{t}^{l}}(\xi,\bar{\xi})C^{x_{s+\kappa}^{l-1},\bar{x}_{t+\kappa}^{l-1}}(\xi,\bar{\xi}).
\end{align*}
For $l=1$, if we define $C^{x_{s+\kappa}^{0},\bar{x}_{t+\kappa}^{0}}(\xi,\bar{\xi})\defeq\frac{x_{s+\kappa}^{0\trsp}\bar{x}_{t+\kappa}^{0}}{d}$, we similarly have
\begin{align*}
\langle\nabla_{\omega_{\kappa}^{1}}f(\xi),\nabla_{\omega_{\kappa}^{1}}f(\bar{\xi})\rangle & =n^{-1}d^{-1}\left\langle \sum_{s}dh_{s}^{1}x_{s+\kappa}^{0},\sum_{s}d\bar{h}_{s}^{1}\bar{x}_{s+\kappa}^{0}\right\rangle
 =\sum_{s,t}\frac{dh_{s}^{1\trsp}d\bar{h}_{t}^{1}}{n}\frac{x_{s+\kappa}^{0\trsp}\bar{x}_{t+\kappa}^{0}}{d}\\
 & \asto D^{h_{s}^{1},\bar{h}_{t}^{1}}(\xi,\bar{\xi})C^{x_{s+\kappa}^{0},\bar{x}_{t+\kappa}^{0}}(\xi,\bar{\xi}).
\end{align*}

Altogether, the NTK is
\begin{align*}
\NTK(\xi,\bar{\xi}) & =\sum_{l=1}^{L}\sum_{\kappa\in K}\langle\nabla_{\omega_{\kappa}^{l}}f(\xi),\nabla_{\omega_{\kappa}^{l}}f(\bar{\xi})\rangle+\langle\nabla_{v}f(\xi),\nabla_{v}f(\bar{\xi})\rangle\\
 & \asto\sum_{l=1}^{L}\sum_{\kappa}\sum_{s,t}D^{h_{s}^{l},\bar{h}_{t}^{l}}(\xi,\bar{\xi})C^{x_{s+\kappa}^{l-1},\bar{x}_{t+\kappa}^{l-1}}(\xi,\bar{\xi})+\begin{cases}
|P|^{-2}\sum_{s,t\in P}C^{x_{s}^{L},\bar{x}_{t}^{L}}(\xi,\bar{\xi}) & \text{if GAP}\\
|P|^{-1}\sum_{s\in P}C^{x_{s}^{L},\bar{x}_{s}^{L}}(\xi,\bar{\xi}) & \text{else.}
\end{cases}
\end{align*}
Here the sum is over $\kappa\in K$ and $s,t\in P$ such that $s+\kappa,t+\kappa\in P$.

\subsubsection{Vectorized NTK Formula}

\global\long\def\conv{\mathcal{T}}%

Let $\{\xi_{i}\}_{i=1}^{k}$ be a set of inputs. For $l=1,\ldots,L$, define the tensors $C^{x^{l}},C^{h^{l}}\in\R^{k\times P\times k\times P}$ by $C_{isjt}^{x^{l}}\defeq C^{x_{s}^{l},\bar{x}_{t}^{l}}(\xi_{i},\xi_{j}),C_{isjt}^{h^{l}}\defeq C^{h_{s}^{l},\bar{h}_{t}^{l}}(\xi_{i},\xi_{j})$. Note that $s$ (and $t$) is a spatial index that may expand to double indices if $P$ is 2-dimensional (e.g. $P=[32]\times[32]$). For any tensor $C=\{C_{isjt}\}_{isjt}\in\R^{k\times P\times k\times P}$, define the linear operator
\[
\conv(C)_{isjt}\defeq\sum_{\kappa}C_{i,s+\kappa,j,t+\kappa}
\]
where the sum is over $\kappa\in K$ such that $s+\kappa,t+\kappa$ are both in $P$. This operator can be easily implemented via convolution in \texttt{pytorch} or \texttt{tensorflow}. Also define $C^{x^{0}}\in\R^{k\times P\times k\times P}$ by $C_{isjt}^{x^{0}}=\xi_{is}^{\trsp}\xi_{jt}/d$. Then assuming $K=-K$\footnote{Note, in general when $K\ne-K$, we have $D^{x^{l-1}}=\sigma_{w}^{2}\conv^{\dagger}(D^{h^{l}})$, where $\conv^{\dagger}$ is the adjoint of $\conv$. },
\begin{align*}
C^{x^{l}} & =\Vt{\phi}(C^{h^{l}})&
C^{h^{l}} & =\sigma_{w}^{2}\conv(C^{x^{l-1}})\\
D^{x^{l-1}} & =\sigma_{w}^{2}\conv(D^{h^{l}})&
D^{h^{l}} & =D^{x^{l}}\odot\Vt{\phi'}(C^{h^{l}}).
\end{align*}
The initial condition is
\[
D_{isjt}^{x^{L}}=\begin{cases}
\sigma_{v}^{2}/|P|^{2} & \text{if GAP}\\
\sigma_{v}^{2}\ind(s=t)/|P| & \text{else.}
\end{cases}
\]
Note in the vectorization case, we only need to compute the entries $C_{isjt}^{x^{l}},C_{isjt}^{h^{l}}$ where $s=t$, as everything else will be 0. Finally, the infinite-width NTK is given by
\[
\mathring{\NTK}_{ij}=\sum_{l=1}^{L}D^{h^{l}}\circ\conv(C^{x^{l-1}})+\begin{cases}
|P|^{-2} \sum_{s,t\in P}C_{isjt}^{x^{L}} & \text{if GAP}\\
|P|^{-1} \sum_{s\in P}C_{isjs}^{x^{L}} & \text{else}.
\end{cases}
\]
where $\circ$ contracts the $s,t$ indices: $(A\circ B)_{ij}=\sum_{s,t\in P}A_{isjt}B_{isjt}$.

\subsection{Batchnorm}

We first detail how to propagate the covariances of activations and of gradients before describing how we can combine them to compute the NTK.
We use the notation of \cref{eqn:BN} and let $\tilde \phi$ denote batchnorm followed by coordinatewise nonlinearity $\phi$.

\paragraph*{Forward Single Batch}

If $h^{1},\ldots,h^{B}$ are the pre-activations of a layer over a batch of size $B$, and
\[
x^{1},\ldots,x^{B}=\widetilde\relu(h^{1},\ldots,h^{B}),
\]
then as discussed in \cref{sec:BackpropInNetsorT}, this is a valid tensor program. If $Z^{h^{1}},\ldots,Z^{h^{B}}$ are jointly distributed as $\Gaus(0,\Sigma)$, then \cite{yang2019wide,yang_mean_2019} showed that $Z^{x^{1}},\ldots,Z^{x^{B}}$ has the 2nd moment matrix $\Sigma',\Sigma'_{ij}=\EV Z^{x^{1}}Z^{x^{B}}$, given by
\begin{equation}
\Sigma'=B\int_{0}^{\infty}\frac{\Vt{\relu}(\Sigma^{G}(I+2s\Sigma^{G})^{-1})}{\sqrt{\det(I+2s\Sigma^{G})}}\dd s
\label{eqn:VBNSingleBatchForward}
\end{equation}
where $\Vt{\relu}$ is as in \cref{fact:Vrelu}, and $\Sigma^{G}=G\tSigma G,G=I_{B}-\frac{1}{B}\mathbf{1}\mathbf{1}^{\trsp}.$

\paragraph*{Forward Cross Batch}

Suppose $\bar{h}^{1},\ldots,\bar{h}^{\bar{B}}$ are the pre-activations of a layer over another batch of size $\bar{B}$ (possibly $\bar{B}\ne B$), such that $(Z^{h^{1}},\ldots,Z^{h^{B}},Z^{\bar{h}^{1}},\ldots,Z^{\bar{h}^{\bar{B}}})$ is jointly distributed as $\Gaus\left(0,\begin{pmatrix}\Sigma & \Xi\\
\Xi^{\trsp} & \bar{\Sigma}
\end{pmatrix}\right)$. Let $\bar{x}^{1},\ldots,\bar{x}^{B}=\widetilde\relu(\bar{h}^{1},\ldots,\bar{h}^{B})$. Then the \emph{cross-batch moment matrix} $\Xi',\Xi'_{ij}=\EV Z^{x^{i}}Z^{\bar{x}^{j}}$, is given by
\begin{equation}
\Xi'=\sqrt{B\bar{B}}\pi^{-1}\int_{0}^{\infty}\dd s\int_{0}^{\infty}\dd t\ (st)^{-1/2}\det(I_{B+\bar{B}}+2\Omega)^{-1/2}\Vt{\relu}(\Pi)_{12}
\label{eqn:VBNCrossBatchForward}
\end{equation}
where 
\begin{align*}
\Omega & =D^{1/2}\begin{pmatrix}G\Sigma G & G\Xi\bar{G}\\
\bar{G}\Xi^{\trsp}G & \bar{G}\bar{\Sigma}\bar{G}
\end{pmatrix}D^{1/2}\\
\Pi & =D^{-1/2}\Omega(I+2\Omega)^{-1}D^{-1/2}\\
D & =sI_{B}\oplus tI_{\bar{B}}=\begin{pmatrix}sI_{B} & 0\\
0 & tI_{\bar{B}}
\end{pmatrix}\\
G & =I_{B}-B^{-1}\mathbf{1}\mathbf{1}^{\top}\\
\bar{G} & =I_{\bar{B}}-\bar{B}^{-1}\mathbf{1}\mathbf{1}^{\top}
\end{align*}
and $\Vt{\relu}(\Pi)_{12}$ is the block of $\Vt{\relu}(\Pi)$ on the first row, second column, of size $B\times\bar{B}$.

\paragraph*{Backward Single Batch}

Now, suppose $dx^{1},\ldots,dx^{B}$ are gradients such that $Z^{dx^{1}},\ldots,Z^{dx^{B}}$ are jointly distributed as $\Gaus(0,\Delta)$ independently from $Z^{h^{1}},\ldots,Z^{h^{B}}$. Let $dh^{1},\ldots,dh^{B}=d\ \widetilde\relu(dx^{1},\ldots,dx^{B}\mid h^{1},\ldots,h^{B})$. Then, by \citet{yang_mean_2019}, $\{Z^{dh^{i}}\}_{i}$ has 2nd-moment matrix $\Delta',\Delta'_{ij}=\EV Z^{dh^{i}}Z^{dh^{j}},$ given by
\begin{equation}
\Delta'=B\int_{0}^{\infty}\delta(\Lambda_{1}+\Lambda_{2}-\Lambda_{3})^{G}\dd s
\label{eqn:VBNSingleBatchBackward}
\end{equation}
where
\begin{align*}
\Lambda_{1} & =\Lambda_{1}(s)=\Delta\odot\Vt{\step}(K(s))\\
\Lambda_{2} & =\Lambda_{2}(s)=\frac{1}{2}s^{2}(\langle\Delta,\Vt{\relu}(K(s))\rangle K(s)+2K(s)J(s)K(s))\\
\Lambda_{3} & =\Lambda_{3}(s)=s(K(s)J(s)+J(s)K(s))
\end{align*}

and

\begin{align*}
\delta(s) & =1/\sqrt{\det(I+2s\Sigma^{G})}\\
K(s) & =\Sigma^{G}(I+2s\Sigma^{G})^{-1}\\
J(s) & =\frac{d\Vt{\relu}(K(s))}{dK(s)}^{\dagger}\{\Delta\}.
\end{align*}

Here, $\frac{d\Vt{\relu}(K(s))}{dK(s)}$ is a matrix-to-matrix linear operator, and $\frac{d\Vt{\relu}(K(s))}{dK(s)}^{\dagger}$ denotes its adjoint, which ``backprops'' a gradient of $\Vt{\relu}(K(s))$ to a gradient of $K(s)$.

\paragraph*{Backward Cross Batch}

Now, suppose $d\bar{x}^{1},\ldots,d\bar{x}^{B}$ are gradients such that $(Z^{dx^{1}},\ldots,Z^{dx^{B}},Z^{d\bar{x}^{1}},\ldots,Z^{d\bar{x}^{\bar{B}}})$ are jointly distributed as $\Gaus\left(0,\begin{pmatrix}\Delta & \chi\\
\chi^{\trsp} & \bar{\Delta}
\end{pmatrix}\right)$ independently from $(Z^{h^{1}},\ldots,Z^{h^{B}},Z^{\bar{h}^{1}},\ldots,Z^{\bar{h}^{\bar{B}}})$. Let
\begin{equation}
d\bar{h}^{1},\ldots,d\bar{h}^{\bar{B}}=d\ \widetilde{\relu}(d\bar{x}^{1},\ldots,d\bar{x}^{B}\mid\bar{h}^{1},\ldots,\bar{h}^{B}).
\label{eqn:VBNCrossBatchBackward}
\end{equation}
Then the \emph{cross-batch moment matrix} $\chi',\chi'_{ij}=\EV Z^{dh^{i}}Z^{d\bar{h}^{j}}$, is given by
\[
\chi'=\sqrt{B\bar{B}}\int_{0}^{\infty}\int_{0}^{\infty}\gamma\ G(\Gamma_{1}+\Gamma_{2}-\Gamma_{3})\bar{G\ }\dd s\ \dd t
\]
where 
\begin{align*}
\Gamma_{1}=\Gamma_{1}(s,t) & =\chi\odot\Vt{\step}(\Pi)_{12}\\
\Gamma_{2}=\Gamma_{2}(s,t) & =4st(\langle\Vt{\relu}(\Pi)_{12},\chi\rangle\Pi_{12}+(\Pi J\Pi)_{12})\\
\Gamma_{3}=\Gamma_{3}(s,t) & =2(t(J\Pi)_{12}+s(\Pi J)_{12})
\end{align*}
with $A_{12}$ denoting the off-diagonal block of $A$, and
\begin{align*}
\gamma=\gamma(s,t) & =\pi^{-1}s^{-1/2}t^{-1/2}\det(I_{B+\bar{B}}+2\Omega)^{-1/2}\\
D=D(s,t) & =\begin{pmatrix}sI_{B} & 0\\
0 & tI_{\bar{B}}
\end{pmatrix}\\
\Omega=\Omega(s,t) & =D^{1/2}\Sigma D^{1/2}\\
\Pi=\Pi(s,t) & =D^{-1/2}\Omega(I+2\Omega)^{-1}D^{-1/2}\\
J=J(s,t) & =\frac{d\Vt{\relu}(\Pi)}{d\Pi}^{\dagger}\{\begin{pmatrix}0 & \chi\\
\chi^{\trsp} & 0
\end{pmatrix}\}
\end{align*}

The backward equations \cref{eqn:VBNSingleBatchBackward,eqn:VBNCrossBatchBackward} are not explicit in \cite{yang_mean_2019} but can be derived from Lemma H.5, Eq~(52), Prop~G.8, and Lemma G.10 from \cite{yang_mean_2019}.

\subsubsection{NTK}

Because batchnorm turns the neural network into a batch-to-batch function, its infinite-width NTK is constructed slightly differently than other networks demonstrated in the main text. We summarize its computation below.

\paragraph*{Two inputs of the same batch}

Let $x_{1},\ldots,x_{B}\in\R^{d}$ be a batch of inputs.

Consider a batchnorm-ReLU MLP with $L$ hidden layers and width $n$. Its forward pass is given by
\[
h_{i}^{l}=\omega^{l}x_{i}^{l-1}\in\R^{n},\quad x^{l}=\widetilde{\relu}(h^{l})\in\R^{B\times n},\quad x_{i}^{1}=\omega^{1}x_{i}\in\R^{n}
\]
with weights $\omega_{\alpha\beta}^{l}\sim\Gaus(0,1)$, and it has output $y_{i}=\frac{1}{\sqrt{n}}v^{\trsp}x^{L}$ for parameters $v\sim\Gaus(0,1)$.

Starting with the $\Sigma_{ij}^{0}=\langle x_{i},x_{j}\rangle/d$, compute $\Sigma^{l}=(\Sigma^{l-1})',l=1,\ldots L,$ according to \cref{eqn:VBNSingleBatchForward}.

Suppose we want to compute the NTK's value on two inputs $x_{i},x_{j}$, possibly the same. Start with $\Delta^{L+1}=\frac{1}{2}\left(\delta_{ij}+\delta_{ji}\right)$ where $\delta_{ij}$ is the matrix with zero everywhere except 1 at the $(i,j)$th entry. Then compute $\Delta^{l}=(\Delta^{l+1})',l=L,\ldots,1$ according to \cref{eqn:VBNSingleBatchBackward}. Then

\[
\mathring{\NTK}(x_{i},x_{j})=\sum_{l=0}^{L}\langle\Sigma^{l},\Delta^{l+1}\rangle.
\]

\paragraph*{Two inputs of different batches}

Let $\bar{x}_{1},\ldots,\bar{x}_{B}\in\R^{d}$ be a second batch of inputs, and compute $\bar{\Sigma}^{l},l=0,\ldots,L$ for them just like how $\Sigma^{l}$ are computed for $x_{1},\ldots,x_{B}\in\R^{d}$ above. In addition, starting with $\Xi_{ij}^{0}=\langle x_{i},\bar{x}_{j}\rangle/d$, compute $\Xi^{l}=(\Xi^{l-1})',l=1,\ldots,L,$ according to \cref{eqn:VBNCrossBatchForward}.

Suppose we want to compute the NTK's value on two inputs $x_{i},\bar{x}_{j}$ from different batches. Start with $\chi^{L+1}=\delta_{ij}$, compute $\chi^{l}=(\chi^{l+1})',l=L,\ldots,1,$ according to \cref{eqn:VBNCrossBatchBackward}. Then
\[
\mathring{\NTK}(x_{i},\bar{x}_{j})=\sum_{l=0}^{L}\langle\Xi^{l},\chi^{l+1}\rangle.
\]

\subsection{Transformer}

\global\long\def\Layernorm{\mathrm{L}}%

\global\long\def\Attention{\mathrm{Attn}}%

\global\long\def\SoftMax{\mathrm{SoftMax}}%

\global\long\def\relu{\mathrm{relu}}%

We'll work with the following transformer variant. Let $x_{1}^{0},\ldots,x_{T}^{0}\in\R^{d}$ be a sequence of inputs (the superscript will be layer index, and the subscript will be token index). Then each layer $l$ of our transformer computes the following 
\begin{align*}
k_{i}^{l} & =U^{l}x_{i}^{l-1}\in\R^{n}\nonumber &
y_{i}^{l} & =\Attention(k_{i}^{l},k^{l},k^{l})+k_{i}^{l}&
z_{i}^{l} & =\Layernorm(y_{i}^{l})\\
g_{i}^{l} & =W^{l1}z_{i}^{l}&
h_{i}^{l} & =W^{l2}\mathrm{\phi}(g_{i}^{l})&
x_{i}^{l} & =\Layernorm(h_{i}^{l}+z_{i}^{l})%
\end{align*}
where $U^{l}$, $W^{l1},W^{l2}$ are weights, $\phi$ is nonlinearity (e.g. relu), $k^{l}=\{k_{j}^{l}\}_{j=1}^{T}$ and $\Attention$ and $\Layernorm$ are Attention and Layernorm as in \cref{sec:BackpropInNetsorT}. The network outputs a single scalar from the average of the final embeddings $x_{i}^{L}$:
\[
o=\frac{1}{T}\sum_{i=1}^{T}v^{\trsp}x_{i}^{L}/\sqrt{n}
\]
To compute the Transformer-NTK, we will need to use the (simplified) \netsortplus{} Master Theorem (\cref{thm:netsorTPlusMasterTheoremSimple}) due to the presence of Layernorm and Attention.

\paragraph*{Setup}
assume for all $\alpha,\beta\in[n]$, 
\begin{itemize}
\item $W_{\alpha\beta}^{l1},W_{\alpha\beta}^{l2}\sim\Gaus(0,\sigma_{w}^{2}/n)$ for all $l\ge1$ 
\item $U_{\alpha\beta}^{l}\sim\Gaus(0,\sigma_{u}^{2}/n)$ for all $l\ge2$ and $U_{\alpha\beta}^{1}\sim\Gaus(0,\sigma_{u}^{2}/d)$ 
\item $v_{\alpha}\sim\Gaus(0,\sigma_{v}^{2})$ 
\item Assume Layernorm $\epsilon=0$
\end{itemize}

\paragraph*{Forward pass}

Suppose we have two sequences $\{x_{1}^{0},\ldots,x_{T}^{0}\}$ and $\{\bar{x}_{1}^{0},\ldots,\bar{x}_{T}^{0}\}$, and we use $\bar{\bullet}$ to denote quantities $\bullet$ computed on the second sequence. Then we see that $\{Z^{h_{i}^{l}},Z^{\bar{h}_{j}^{l}}\}_{i,j},\{Z^{g_{i}^{l}},Z^{\bar{g}_{j}^{l}}\}_{i,j},\{Z^{k_{i}^{l}},Z^{\bar{k}_{j}^{l}}\}_{i,j}$ are mutually independent sets of random variables, each of which is jointly Gaussian with zero mean. Their covariances are given by

\begin{align*}
\Cov(Z^{h_{i}^{l}},Z^{\bar{h}_{j}^{l}}) & =\sigma_{w}^{2}\EV\phi(Z^{g_{i}^{l}})\phi(Z^{\bar{g}_{j}^{l}})\\
\Cov(Z^{g_{i}^{l}},Z^{\bar{g}_{j}^{l}}) & =\sigma_{w}^{2}\EV Z^{z_{i}^{l}}Z^{\bar{z}_{j}^{l}}\\
\Cov(Z^{k_{i}^{l}},Z^{\bar{k}_{j}^{l}}) & =\sigma_{u}^{2}\EV Z^{x_{i}^{l-1}}Z^{\bar{x}_{j}^{l-1}}
\end{align*}

In addition, 

\begin{align*}
Z^{x_{i}^{l}} & =\frac{Z^{h_{i}^{l}}+Z^{z_{i}^{l}}-\EV Z^{h_{i}^{l}}+Z^{z_{i}^{l}}}{std\left(Z^{h_{i}^{l}}+Z^{z_{i}^{l}}\right)}&
Z^{z_{i}^{l}} & =\frac{Z^{y_{i}^{l}}-\EV Z^{y_{i}^{l}}}{std(Z^{y_{i}^{l}})}&
Z^{y_{i}^{l}} & =\sum_{j=1}^{T}\mathring{a}_{ij}^{l}Z^{k_{j}^{l}}+Z^{k_{i}^{l}}
\end{align*}
where
\begin{align*}
    (\mathring{a}_{i1}^{l},\ldots,\mathring{a}_{iT}^{l}) & =\SoftMax(\mathring{c}_{i1}^{l},\ldots,\mathring{c}_{iT}^{l}),\qquad
    \mathring{c}_{ij}^{l}  =\EV Z^{k_{i}^{l}}Z^{k_{j}^{l}}.
\end{align*}

We can easily see that $\EV Z^{h_{i}^{l}}+Z^{z_{i}^{l}}=\EV Z^{y_{i}^{l}}=0$. So we can simplify

\begin{align*}
Z^{x_{i}^{l}} & =\frac{Z^{h_{i}^{l}}+Z^{z_{i}^{l}}}{\sqrt{\EV\left(Z^{h_{i}^{l}}+Z^{z_{i}^{l}}\right)^{2}}},\qquad
Z^{z_{i}^{l}} =\frac{Z^{y_{i}^{l}}}{\sqrt{\EV(Z^{y_{i}^{l}})^{2}}}.
\end{align*}

These equations yield a recursive way of computing all random variable $Z^{\bullet}$ associated to vector $\bullet$ in the forward pass.

\paragraph*{Backward pass}

Backprop is given by the following equations.
\begin{align*}
dx_{i}^{L} & =\frac{1}{T}v&
dh_{i}^{l} & =d\Layernorm(dx_{i}^{l}\mid h_{i}^{l}+z_{i}^{l})&
dg_{i}^{l} & =\phi'(g_{i}^{l})\odot W^{l2\trsp}dh_{i}^{l}\\
dz_{i}^{l} & =W^{l1\trsp}dg_{i}^{l}+dh_{i}^{l}&
dy_{i}^{l} & =d\Layernorm(dz_{i}^{l}\mid y_{i}^{l})&
dx_{i}^{l-1} & =U^{l\trsp}dk_{i}^{l}
\end{align*}
and
\begin{align*}
dk_{i}^{l} =dy_{i}^{l}&+\dqAttn{i}(\{dy_{i}^{l}\}_{i}\mid\{k_{i}^{l}\}_{i},\{k_{i}^{l}\}_{i},\{k_{i}^{l}\}_{i})\\
 &+\dkAttn{i}(\{dy_{i}^{l}\}_{i}\mid\{k_{i}^{l}\}_{i},\{k_{i}^{l}\}_{i},\{k_{i}^{l}\}_{i})\\
 &+\dvAttn{i}(\{dy_{i}^{l}\}_{i}\mid\{k_{i}^{l}\}_{i},\{k_{i}^{l}\}_{i},\{k_{i}^{l}\}_{i})
\end{align*}
This implies the following equations (via \cref{box:netsortPlusKeyIntuition}) for the associated random variables.

\begin{align*}
    Z^{dh_{i}^{l}} & =d\mathring{\Layernorm}\left(Z^{dx_{i}^{l}}\mid Z^{h_{i}^{l}}+Z^{z_{i}^{l}}\right),&
    Z^{dy_{i}^{l}} & =d\mathring{\Layernorm}\left(Z^{dz_{i}^{l}}\mid Z^{y_{i}^{l}}\right),&
    \EV Z^{dx_{i}^{l}}Z^{d\bar{x}_{j}^{l}} & =\sigma_{u}^{2}\EV Z^{dk_{i}^{l}}Z^{d\bar{k}_{j}^{l}}
\end{align*}
\begin{align*}
    Z^{dk_{i}^{l}} & =Z^{dy_{i}^{l}}+\dqAttnLim{i}(Z^{dy^{l}}\mid Z^{dk^{l}},Z^{dk^{l}},Z^{dk^{l}})\\
    & \phantom{{}=Z^{dy_{i}^{l}}}+\dkAttnLim{i}(Z^{dy^{l}}\mid Z^{dk^{l}},Z^{dk^{l}},Z^{dk^{l}})\\
    & \phantom{{}=Z^{dy_{i}^{l}}}+\dvAttnLim{i}(Z^{dy^{l}}\mid Z^{dk^{l}},Z^{dk^{l}},Z^{dk^{l}})\\
\end{align*}
\begin{align*}
    \EV Z^{dx_{i}^{L}}Z^{d\bar{x}_{j}^{L}} & =T^{-2}\sigma_{v}^{2}\\
    \EV Z^{dg_{i}^{l}}Z^{d\bar{g}_{j}^{l}} & =\sigma_{w}^{2}\EV Z^{dh_{i}^{l}}Z^{d\bar{h}_{j}^{l}}\EV\phi'(Z^{g_{i}^{l}})\phi'(Z^{\bar{g}_{j}^{l}})\\
    \EV Z^{dz_{i}^{l}}Z^{d\bar{z}_{j}^{l}} & =\sigma_{w}^{2}\EV Z^{dg_{i}^{l}}Z^{d\bar{g}_{j}^{l}}+\EV Z^{dh_{i}^{l}}Z^{d\bar{h}_{j}^{l}}\\
\end{align*}
where $Z^{dy^{l}}=\{Z^{dy_{i}^{l}}\}_{i},Z^{dk^{l}}=\{Z^{dk_{i}^{l}}\}_{i}$, and $d\mathring{\Attention}$ and $d\mathring{\Layernorm}$ are as in \cref{eqn:dLayernormLimit,eqn:dAttnLimit}.

\paragraph{Some Simplifications}
Now we can write
\[
Z^{dh_{i}^{l}}=d\mathring{\Layernorm}\left(Z^{dx_{i}^{l}}\mid Z^{h_{i}^{l}}+Z^{z_{i}^{l}}\right)=\frac{1}{std\left(Z^{h_{i}^{l}}+Z^{z_{i}^{l}}\right)}\mathrm{Center}\left(Z^{dx_{i}^{l}}-Z^{x_{i}^{l}}\EV Z^{dx_{i}^{l}}Z^{x_{i}^{l}}\right)
\]
as in \cref{eqn:LayernormLimit}. But because of the \textbf{MatMul} rule of \cref{box:netsortPlusKeyIntuition}, $Z^{dx_{i}^{l}}$ is a zero mean Gaussian independent from $Z^{x_{i}^{l}}$, and 
$
\EV Z^{dx_{i}^{l}}Z^{x_{i}^{l}}=0.
$
Therefore $\mathrm{Center}\left(Z^{dx_{i}^{l}}-Z^{x_{i}^{l}}\EV Z^{dx_{i}^{l}}Z^{x_{i}^{l}}\right)=Z^{dx_{i}^{l}}$, and
\[
Z^{dh_{i}^{l}}=\frac{Z^{dx_{i}^{l}}}{std\left(Z^{h_{i}^{l}}+Z^{z_{i}^{l}}\right)}.
\]
Likewise,
\[
Z^{dy_{i}^{l}}=d\mathring{\Layernorm}\left(Z^{dz_{i}^{l}}\mid Z^{y_{i}^{l}}\right)=\frac{Z^{dz_{i}^{l}}}{std(Z^{y_{i}^{l}})}.
\]
Finally, from \cref{eqn:dAttnLimit}, the $\dqAttnLim{i}$ and $\dkAttnLim{i}$ terms in $Z^{dk_{i}^{l}}$ depend linearly on $\{\EV Z^{dy_{i}^{l}}Z^{k_{j}^{l}}\}_{j}$ which vanish again by the \textbf{MatMul} rule of \cref{box:netsortPlusKeyIntuition}.
Therefore,

\[
Z^{dk_{i}^{l}}=Z^{dy_{i}^{l}}+\dvAttnLim{i}(Z^{dy^{l}}\mid Z^{dk^{l}},Z^{dk^{l}},Z^{dk^{l}})=\sum_{j}\mathring{a}_{ji}^{l}Z^{dy_{j}^{l}}+Z^{dy_{i}^{l}}
\]

with $\mathring{a}_{ji}^{l}$ as computed in the forward pass.

The complete simplification:

\begin{align*}
Z^{dh_{i}^{l}} & =\frac{1}{std\left(Z^{h_{i}^{l}}+Z^{z_{i}^{l}}\right)}Z^{dx_{i}^{l}}&
Z^{dy_{i}^{l}} & =\frac{1}{std\left(Z^{z_{i}^{l}}\right)}Z^{dz_{i}^{L}}&
Z^{dk_{i}^{l}} & =\sum_{j}\mathring{a}_{ji}^{l}Z^{dy_{j}^{l}}+Z^{dy_{i}^{l}}
\end{align*}

\begin{align*}
\EV Z^{dg_{i}^{l}}Z^{d\bar{g}_{j}^{l}} & =\sigma_{w}^{2}\EV Z^{dh_{i}^{l}}Z^{d\bar{h}_{j}^{l}}\EV\phi'(Z^{g_{i}^{l}})\phi'(Z^{\bar{g}_{j}^{l}})\\
\EV Z^{dz_{i}^{l}}Z^{d\bar{z}_{j}^{l}} & =\sigma_{w}^{2}\EV Z^{dg_{i}^{l}}Z^{d\bar{g}_{j}^{l}}+\EV Z^{dh_{i}^{l}}Z^{d\bar{h}_{j}^{l}}\\
\EV Z^{dx_{i}^{l}}Z^{d\bar{x}_{j}^{l}} & =\sigma_{u}^{2}\EV Z^{dk_{i}^{l}}Z^{d\bar{k}_{j}^{l}}
\end{align*}

As in \citet{yang2019wide}, all nonlinearities of the \netsortplus{} program (corresponding to nonlinearities and their derivatives in the network) are parameter controlled, and \cref{assm:asRankStab} is satisfied.
So Master Theorem holds, and the NTK has a well-defined almost sure limit which is given by \cref{eqn:NTKlimit}.
We can then summarize the above into the following vectorized formulas for computing this NTK limit. 

\subsubsection{NTK}
Suppose we have a collection of $M$ sequences $\{x_{a1}^{0},\ldots,x_{aT}^{0}\}_{a=1}^{M}$ each with $T$ tokens. We will use $a,b,\ldots$ as sequence indices and $i,j,\ldots$ as token indices.
We will work with 4-tensors in $\R^{M\times T\times M\times T}$, which can be also thought of as $M\times M$ blocks of $T\times T$ matrices.

\paragraph{Notations}
For 4-tensor $C \in \R^{M\times T\times M\times T}$:
\begin{enumerate}
    \item $BlockDiag(C)$ is the 4-tensor with $BlockDiag(C)_{aibi} = C_{aibj}\ind(a=b)$.
    \item $Diag(C)$ is the 4-tensor with $Diag(C)_{aibj} = C_{aibj}\ind(a=b)\ind(i=j)$.
    \item Juxtaposition represents multiplication of tensors reshaped as matrices $C\bar C=\texttt{einsum(`aibj,bjck->aick'}, C, \bar C\texttt{)}$.
    \item $Corr(C)=Diag(C)^{-1/2}C\ Diag(C)^{-1/2}$.
    \item $\SoftMax(C)$ applies SoftMax to $C$ in the last dimension.
\end{enumerate}

\paragraph{NTK Computation}
the NTK, as a $M\times M$ matrix, is
\[
\mathring \NTK = \frac{1}{T^{2}}\circ C^{x^{L}}+\sum_{l=1}^{L}D^{k^{l}}\circ C^{x^{l-1}}+D^{g^{l}}\circ C^{z^{l}}+D^{h^{l}}\circ V_{\phi}(C^{g^{l}})
\]
where $X\circ Y$ is a matrix for 4-tensors $X,Y$, with $(X\circ Y)_{ab}=\sum_{ij}X_{aibj}Y_{aibj}$, and the relevant tensors are computed by
\begin{multicols}{2}
Forward:
\begin{align*}
C_{aibj}^{x^{0}} & =x_{ai}^{\trsp}x_{bj}/d\\
C^{k^{l}} & =\sigma_{u}^{2}C^{x^{l-1}}\\
A^{l} & =BlockDiag(\SoftMax(C^{k^{l}}))\\
C^{y^{l}} & =(A^{l}+I)C^{k^{l}}(A^{l\trsp}+I)\\
C^{z^{l}} & =Corr(C^{y_{i}^{l}})\\
\Delta^{z^{l}} & =Diag(C^{y_{i}^{l}})^{-1/2}\\
C^{g^{l}} & =\sigma_{w}^{2}C^{z^{l}}\\
C^{h^{l}} & =\sigma_{w}^{2}V_{\phi}(C^{g^{l}})\\
C^{x^{l}} & =Corr(C^{h^{l}}+C^{z^{l}})\\
\Delta^{x^{l}} & =Diag(C^{h^{l}}+C^{z^{l}})^{-1/2}
\end{align*}

Backward:
\begin{align*}
D_{aibj}^{x^{L}} & =\sigma_{v}^{2}/T^{2}\\
D^{h^{l}} & =\Delta^{x^{l}}D^{x^{l}}\Delta^{x^{l}}\\
D^{g^{l}} & =\sigma_{w}^{2}D^{h^{l}}\odot V_{\phi'}(C^{g^{l}})\\
D^{z^{l}} & =\sigma_{w}^{2}D^{g^{l}}+D^{h^{l}}\\
D^{y^{l}} & =\Delta^{z^{l}}D^{z^{l}}\Delta^{z^{l}}\\
D^{k^{l}} & =(A^{l\trsp}+I)D^{y^{l}}(A^{l}+I)\\
D^{x^{l-1}} & =\sigma_{u}^{2}D^{k^{l}}
\end{align*}
\end{multicols}

\section{Theoretical Tools}

\label{sec:proofs}

We will use the following trivial but useful fact repeatedly.

\begin{lemma}\label{lem:powerbound}
For an integer $m$, and complex numbers $a_i \in \C$, $i \in [k]$,
\[
\left| \sum_{i=1}^k a_i \right|^m
\le 
k^{m-1} \sum_{i=1}^k \left|a_i\right|^m
.
\]
\end{lemma}
\begin{proof}
Expand the power in the LHS using the multinomial theorem, apply AM-GM to each summand, and finally aggregate using triangle inequality.
\end{proof}

\subsection{Probability Facts}

This section is largely the same as section G.1 of \citet{yang2019wide}.
All proofs can be found there.

\paragraph{Notations}
Given two random variables $X, Y$, and a $\sigma$-algebra $\Aa$, the notation $X \disteq_\Aa Y$ means that for any integrable function $\phi$ and for any random varible $Z$ measurable on $\Aa$, $\EV \phi(X) Z = \EV \phi(Y)Z$.
We say that $X$ is distributed as (or is equal in distribution to) $Y$ conditional on $\Aa$.
In case $\Aa$ is the trivial $\sigma$-algebra, we just write $X \disteq Y$.
The expression $X \distto Y$ (resp. $X \asto Y$) means $X$ converges to $Y$ in distribution (resp. almost surely).

\begin{lemma}\label{lemma:momentBoundASConvergence}
Let $\{X_n\}_{n \ge 1}$ be a sequence of random variables with zero mean.
If for some $p \in \N$ and for all $n$, $\EV X_n^{2p} \le c n^{-1-\rho}$, for some $\rho > 0$, then $X_n \to 0$ almost surely.
\end{lemma}

The following is a standard fact about multivariate Gaussian conditioning
\begin{prop}\label{prop:GaussianCondition}
Suppose $\R^{n_1 + n_2} \ni x \sim \Gaus(\mu, K)$, where we partition $x = (x_1, x_2) \in \R^{n_1} \times \R^{n_2}, \mu = (\mu_1, \mu_2) \in \R^{n_1} \times \R^{n_2}$, and $K = \begin{pmatrix} K_{11} & K_{12}\\ K_{21} & K_{22}\end{pmatrix}$.
Then
$x_1 \disteq_{x_2} \Gaus(\mu|_{x_2}, K|_{x_2})$
where
\begin{align*}
    \mu|_{x_2}
        &=
            \mu_1 - K_{12} K_{22}^+ (x_2 - \mu_2)\\
    K|_{x_2}
        &=
            K_{11} - K_{12} K_{22}^+ K_{21}.
\end{align*}

\end{prop}

\begin{lemma}[Stein's lemma]\label{lemma:stein}
For jointly Gaussian random variables $Z_1, Z_2$ with zero mean, and any function $\phi: \R \to \R$ where $\EV \phi'(Z_1)$ and $\EV Z_1 \phi(Z_2)$ exists, we have
\[\EV Z_1 \phi(Z_2) = \Cov(Z_1, Z_2) \EV \phi'(Z_2).\]
\end{lemma}

\begin{lemma}\label{lemma:gaussianDer}
Let $\Phi: \R^n \to \R$ be measurable.
Then for $z \sim \Gaus(\zeta, \Sigma)$, the following Hessian and gradient matrices are equal:
\begin{align*}
    \Jac{^2}{\zeta^2} \EV \Phi(z)
        &=
            2\Jac{}{\Sigma} \EV \Phi(z)
\end{align*}
whenever both sides exist.
\end{lemma}

\subsection{Gaussian Conditioning Trick}

\paragraph{Review of Moore-Penrose Pseudoinverse}

Let $A^+$ denote the Moore-Penrose pseudo-inverse of a matrix $A$.
\begin{lemma}\label{lemma:condTrick}
Let $A \in \R^{n \times m}$ be a matrix with random Gaussian entries, $A_{ij} \sim \Gaus(0, \sigma^2)$.
Consider fixed matrices $Q \in \R^{m \times q}, Y \in \R^{n \times q}, P \in \R^{n \times p}, X \in \R^{m \times p}$.
Suppose there exists a solution in $A$ to the equations $Y = AQ$ and $X = A^\trsp P$.
Then the distribution of $A$ conditioned on $Y = AQ$ and $X = A^\trsp P$ is
\begin{align*}
    A &\disteq_{Y=AQ, X=A^\trsp P} E + \Pi_P^\perp \tilde A \Pi_Q^\perp
\end{align*}
where
\begin{align*}
    E
        &=
            Y Q^+
            + P^{+\trsp} X^\trsp
            - P^{+\trsp} P^\trsp
                YQ^+,
\end{align*}
$\tilde A$ is an iid copy of $A$,
and $\Pi_P^\perp = I - \Pi_P$ and $\Pi_Q^\perp = I - \Pi_Q$ in which $\Pi_P = PP^+$ and $\Pi_Q = QQ^+$ are the orthogonal projection to the space spanned by the column spaces of $P$ and $Q$ respectively.
\end{lemma}
\begin{proof}
    See \citet[Lemma G.7]{yang2019wide}.
\end{proof}

\subsection{Law of Large Numbers for Images of Weakly Correlated Gaussians}

\begin{lemma}\label{lemma:projectionDiagonal}
    Let $\Pi \in \R^{n \times n}$ be an orthogonal projection matrix.
    Then each diagonal entry $\Pi_{ii} \in [0, 1].$
    \end{lemma}
    \begin{proof}
    Because $\Pi = \Pi^2$, we have for each $i$, $\Pi_{ii} = \sum_{j} \Pi_{ij}^2 \implies \Pi_{ii} (1 - \Pi_{ii}) = \sum_{j \ne i} \Pi_{ij}^2 \ge 0 \implies \Pi_{ii} \in [0, 1].$
    \end{proof}

\begin{thm}\label{thm:controlHighMoments}
    \newcommand{\LL}{2p}
    \newcommand{\CC}{\mathcal{C}}
    Let $z \sim \Gaus(0, \Pi)$ where $\Pi \in \R^{n \times n}$ is a matrix such that its correlation matrix $C = D^{-1/2} \Pi D^{-1/2}, D = \Diag(\Pi),$ has $\sum_{i<j} C_{ij}^2 \le R$ for some constant $R$ as $n \to \infty$.
    (So an orthogonal projection matrix of rank $n - O(1)$ satisfies this condition).
    Consider functions $\phi_i: \R \to \R$ for each $i \in [n]$ with mean $\mu_i = \EV_x \phi_i(x)$ under $x \sim \Gaus(0, \Pi_{ii})$.
    Suppose each $\phi_i$ has finite ($\LL$)th centered moment $\EV_x (\phi_i(x) - \mu_i)^{\LL}$, for $x \sim \Gaus(0, \Pi_{ii})$, where $p \ge 6$.
    Then for $Q \defeq \f 1 n \sum_{i=1}^n \phi_i(z_i),$ as $n \to \infty$,
    \begin{align*}
        \EV[(Q - \EV Q)^{2p}]
            &\le
               \CC
                n^{-1.5} \max_{i \in [n]}
                    \EV_{x \sim \Gaus(0, \Pi_{ii})}
                        \left(\phi_{i}(x) - \mu_i \right)^{2p}
    \end{align*}
    for some constant $\CC$ depending on $p$ and $R$, but not on $n$ or the functions $\phi_i$.
    If in addition, each $\phi_i$ has finite centered moments of order $2p L$ for some $L > 1$, then
    \begin{align*}
        \EV[(Q - \EV Q)^{2p}]
            &\le
               \CC
                n^{-1.5+1/L}
                    \sqrt[L]{\f 1 n \sum_{i=1}^n 
                        \EV_{x \sim \Gaus(0, \Pi_{ii})}
                        \left(\phi_{i}(x) - \mu_i \right)^{2p L}
                        }
                .
    \end{align*}
\end{thm}

\begin{proof}
    See \citet{yangTP3}.
\end{proof}

\section{Proof of Main Theorem}
\label{sec:ProofMainTheorem}

\renewcommand{\nu}{\eta}
\newcommand{\rankext}[1]{{\color{green}#1}}

\newcommand{\coreset}{{\mathcal{M}}}
\newcommand{\basespace}{\mathcal{U}}
\newcommand{\MM}{m}

In this section, we will give the proof for \cref{thm:netsorTMasterTheoremFormal} which is equivalent to \cref{thm:netsorTMasterTheoremSimple}.
We reproduce the statement below

\NetsorTMasterTheoremFormal*

\paragraph{Comparison against \netsor{} Master Theorem \citep{yang2019wide}}
We will follow the general outline of the inductive proof of \netsor{} Master Theorem in \citet{yang2019wide}.
Here, the correlation of a matrix with its transpose (\cref{remk:addedComplexity})
causes additional difficulty in proving rank stability and zero stability properties (\cref{sec:rankStabilityZeroStability}) as well as the induction hypothesis (\ref{IH:MomConv}$(m)$).
The main sections dealing with these difficulties are \cref{sec:inductiveSetup} which describes the setup for the induction, \cref{sec:sigmalimit>0} which proves part of rank and zero stability properties, and \cref{sec:probA,sec:probB} which prove parts of the inductive step using the law of large numbers for weakly correlated random variables \cref{thm:controlHighMoments}.

\paragraph{A Bit of Notation and Terminology}
Note that, for each $n$, the randomness of our program specified by \cref{thm:netsorTMasterTheoremFormal} comes from the sampling of the input variables.
Let $\basespace$ be the product space obtained from multiplying together the corresponding probability space for each $n$.
Each sample from this product probability space thus correspond to a sequence $\{S(n)\}_n$ of instantiatiations of input variables.
Below, when we say ``almost surely'' (often abbreviated ``a.s.''), we mean ``almost surely over the probability of $\basespace{}$.''
We will also often make statements of the form 
\begin{equation*}
\text{\emph{almost surely (or, a.s.), for all large $n$, \quad $\mathcal A(n)$ is true}}
\end{equation*}
where $\mathcal A(n)$ is a claim parametrized by $n$.
This means that for all but a $\basespace{}$-probability-zero set of sequences $\{S(n)\}_n$ of input variable instantiations, $\mathcal A(n)$ is true for large enough $n$.
Note that the order of the qualifiers is very important here.

\paragraph{We induct, but on what?}
A natural way of going about proving \cref{thm:netsorTMasterTheoremFormal} is by inducting on the number of variables in a program.
It turns out this is not enough to prove our claim in its full generality, and it would be more fruitful to perform a simultaneous induction on our claim (\ref{IH:MomConv}) along with another statement, parametrized by $\MM$,
\begin{description}
\item[Moments\label{IH:MomConv}]\!\!\!$(\MM)$\ \ \ 
    For any polynomially-bounded $\psi: \R^\MM \to \R$, as $n \to \infty$,
\begin{align*}
    \f 1 n \sum_{\alpha=1}^n \psi(g^1_\alpha, \ldots, g^\MM_\alpha) \asto \EV_{Z \sim \Gaus(\tmu, \tSigma)}\psi(Z).
\end{align*}
\item[CoreSet\label{IH:coreSet}]\!\!\!$(\MM)$\ \ \ 
    There exists a ``core set'' $\coreset \sbe [\MM]$ such that, 
 \begin{description}
    \item[Basis\label{prop:basis}]\!\!\!$(\MM)$\ \ \ 
    almost surely, for large enough $n$, for every $i \in [\MM]$, there exist \emph{unique} constants (not depending on $n$) $\{a_j\}_{j \in \coreset}$ such that $g^i = \sum_{j \in \coreset} a_j g^j$.
    Note the uniqueness implies that $\{g^i\}_{i \in \coreset}$ is linearly independent.
    \item[NullAvoid\label{prop:nullAvoid}]\!\!\!$(\MM)$\ \ \ 
    for every triangular array of Lesbegue measure zero sets $\{A_{n\alpha} \in \R^{\coreset} \}_{n \in \N, \alpha \in [n]}$, almost surely for all large enough $n$, for all $\alpha \in [n]$, we have
    \[\{g^i_\alpha\}_{i \in \coreset} \not \in A_{n \alpha}.\]
    In other words, the values $\{g^i_\alpha\}_{\alpha \in \coreset}$ of the core set ``avoid'' Lebesgue measure zero sets asymptotically.
    Intuitively, this says that the distribution of these values are not singular.
    (Note the LHS depends on $n$ although we are suppressing it notationally)
\end{description}
\end{description}

Let us explain in brief why we need to consider \ref{IH:coreSet} satisfying \ref{prop:basis} and \ref{prop:nullAvoid}.
\begin{itemize}
\item
    \ref{prop:basis} reduces the consideration of \ref{IH:MomConv} to only the core set G-vars, since every other G-var is asymptotically a linear combination of them.
\item
    When we apply the Gaussian conditioning technique \cref{prop:GaussianCondition}, we need to reason about the pseudo-inverse $\Lambda^+$ of some submatrix $\Lambda$ of a covariance matrix.
    Each entry of $\Lambda$ is of the form $\f 1 n \sum_{\alpha=1}^n \phi_i(g^1_\alpha, \ldots, g^{\MM-1}_\alpha) \phi_j(g^1_\alpha, \ldots, g^{\MM-1}_\alpha)$ for a collection of polynomially bounded scalar functions $\{\phi_i\}_i$.
    This $\Lambda$ will be a random variable which converges a.s.\ to a determinstic limit $\mathring \Lambda$  as $n \to \infty$.
    It should be generically true that $\Lambda^+ \asto \mathring \Lambda^+$ as well, which is essential to make the Gaussian conditioning argument go through.
    But in general, this is guaranteed only if $\Lambda$'s rank doesn't drop suddenly in the $n \to \infty$ limit.
    We thus need to guard against the possibility that $g^1, \ldots, g^\MM$, in the limit, suddenly concentrate on a small set on which $\{\phi_i(g^1, \ldots, g^\MM)\}_i$ are linearly dependent.
    This is where \ref{prop:nullAvoid} comes in.
    It tells us that $g^1, \ldots, g^\MM$ will avoid any such small set asymptotically, so that indeed the rank of $\Lambda$ will not drop in the limit.
\end{itemize}

\paragraph{Proof organization}

We will show that \ref{IH:MomConv} and \ref{IH:coreSet} are true for input variables, as the base case, and
\begin{equation*}
\text{\ref{IH:MomConv}}(\MM-1) \text{ and } \text{\ref{IH:coreSet}}(\MM-1) \implies \text{\ref{IH:MomConv}}(\MM) \text{ and } \text{\ref{IH:coreSet}}(\MM)
\end{equation*}
as the inductive step.
By induction, we obtain \ref{IH:MomConv}$(M)$, which is \cref{thm:netsorTMasterTheoremFormal}.

The base cases are easy and we will dispatch with them immediately after this in \cref{sec:basecases}, but the inductive step is much more complicated, and we will need to set up notation in \cref{sec:inductiveSetup}.
During this setup, we prove some basic limit theorems using the induction hypothesis.
However, the full generality of these claims requires some consequences of \ref{IH:coreSet}, which we call ``rank stability'' and ``zero stability.''
These notions are introduced and proved in \cref{sec:rankStabilityZeroStability}.

We would then finally be able to handle the inductive steps at this point.
We first prove
\begin{equation*}
\text{\ref{IH:MomConv}$(\MM-1)$ and \ref{IH:coreSet}$(\MM-1)$} \implies \text{\ref{IH:coreSet}}(\MM)
\end{equation*}
in \cref{sec:inductiveCoreSet} because it is easier.
Then we prove
\begin{equation*}
\text{\ref{IH:MomConv}$(\MM-1)$ and \ref{IH:coreSet}$(\MM-1)$} \implies \text{\ref{IH:MomConv}}(\MM)
\end{equation*}
in \cref{sec:inductiveMoments}.

\subsection{Base Cases: \ref{IH:MomConv} and \ref{IH:coreSet} for Input Variables}
\label{sec:basecases}

\paragraph{Base case: \ref{IH:MomConv}(input vars)}
Suppose the input variables are $x^1, \ldots, x^k: \Gtype(n)$ (so that $\mu \in \R^k, \Sigma \in \R^{k \times k}$).
We need to show that for any polynomially-bounded function $\psi: \R^k \to \R$,
\begin{align*}
    \f 1 n \sum_{\alpha=1}^n \psi(x^1_\alpha, \ldots, x^k_\alpha) \asto \EV_{Z \sim \Gaus(\tmu, \tSigma)}\psi(Z),
\end{align*}
where $\psi$ on the RHS ignores all coordinates corresponding to non-input G-vars.
Since $\tmu$ and $\tSigma$ restricted to input variables are just $\mu$ and $\Sigma$ (see \cref{eqn:extendedMuSigma}), the RHS expectation is just
\begin{align*}
    \EV_{Z \sim \Gaus(\tmu, \tSigma)}\psi(Z) = \EV_{Z^{in} \sim \Gaus(\mu, \Sigma)} \psi(Z^{in})
\end{align*}
and the almost sure convergence we desire is just a result of the law of large numbers.

\paragraph{Base Case: \ref{IH:coreSet}(input vars)}
Let $x^1, \ldots, x^k$ be the input G-vars as above.
Pick the core set $\coreset$ to be any subset of $[k]$ such that $\rank \Sigma|_\coreset = \rank \Sigma$.
Then it's straightforward to verify \ref{prop:basis} and \ref{prop:nullAvoid}.

\subsection{Inductive Case: Setup}
\label{sec:inductiveSetup}
We now assume \ref{IH:MomConv}$(\MM-1)$ and \ref{IH:coreSet}$(\MM-1)$ and want to reason about $g^\MM$ to show \ref{IH:MomConv}$(\MM)$ and \ref{IH:coreSet}$(\MM)$.
Suppose
\begin{align*}
    g^{\MM} := A h \quad \text{where} \quad A: \Atype(n,n) \text{ and $h: \Htype(n)$ was introduced by $h := \phi(g^{1}, \ldots, g^{\MM - 1})$}
\end{align*}
(WLOG padding input slots if necessary; if $h = g^i$ is a G-var, then just let $\phi$ be the projection to the $i$th coordinate).
For brevity, we will just write $g = g^{\MM}$.
Consider all previous instances where $A$ or $A^\trsp$ is used: 
\[\hat g^{i} := A \hat h^{i}, i = 1, \ldots, r, \quad\text{and}\quad
\check g^{j} := A^\trsp \check h^{j}, j = 1, \ldots, s.\]
Define
\begin{equation}
\hat G \defeq [\hat g^1| \ldots| \hat g^{r}] \in \R^{n \times r}, \check G \defeq [\check g^1| \ldots| \check g^{s}] \in \R^{n \times s}, \hat H \defeq [\hat h^1| \ldots| \hat h^r], \check H \defeq [\check h^1| \ldots| \check h^s]
.
\end{equation}
We will also use $\hat G$ to denote the \emph{set} of G-vars $\{\hat g^1, \ldots, \hat g^r\}$ when we later write expressions like $\tSigma(\hat G, \hat G)$.
Let $\Bb$ be the $\sigma$-algebra spanned by all previous G-vars $g^1, \ldots, g^{\MM-1}$ (and hence also all previous H-vars).
Conditioning on $\Bb$, $A$ is constrained by $\hat G = A\hat H, \check G = A^\trsp \check H$, and we have by \cref{lemma:condTrick},
\begin{align*}
    g \disteq_{\Bb} (E + \Pi_{\check H}^\perp \tilde A \Pi_{\hat H}^\perp) h
\end{align*}
where
\begin{align*}
    E
        &=
            \hat G \hat H^+
            + \check H^{+\trsp} \check G^{\trsp}
            - \check H^{+\trsp} \check G^\trsp \hat H \hat H^+
            \\
        &=
            \hat G (\hat H^\trsp \hat H)^+ \hat H^{\trsp}
            + \check H (\check H^{\trsp} \check H)^+ \check G^{\trsp}
            - \check H (\check H^{\trsp} \check H)^+ \check G^\trsp \hat H (\hat H^\trsp \hat H)^+ \hat H^\trsp,
            \numberthis
            \label{eqn:Eexpansion}
\end{align*}
$\tilde A$ is an independent copy of $A$ and $\Pi_{\hat H} = \hat H \hat H^+ = \hat H(\hat H^\trsp \hat H)^+ \hat H^\trsp$ is the projection to the column space of $\hat H$ (likewise for $\Pi_{\check H}$).

\begin{remk}\label{remk:addedComplexity}
    Note if \ref{linetype:Trsp} is not allowed (as in \netsor{} Master Theorem \citep{yang2019wide}), then this would simplify a lot to $g\disteq_\Bb (\hat G \hat H^+ + \tilde A \Pi_{\hat H}^\perp)h.$
    In particular, compared to the \netsor{} Master Theorem, we cannot straightforwardly think of $g$ as having iid Gaussian coordinates because of the projection $\Pi_{\check H}^\perp$ in front of $\tilde A$.
    Most of the added complexity of this proof of \netsort{} Master Theorem comes from this fact.
\end{remk}

\begin{remk}
    On the other hand, with the BP-like assumption, $E$ is roughly equal to $\hat G \hat H^+$; see \cref{lemma:omegaExpansion}.
    However, this is very far from true when we don't assume the program is BP-like.
\end{remk}

If we define
\begin{align}
    \omega
        &\defeq
            E h,
            \quad
    \sigma \defeq
        \sigma_A
        \sqrt{\|\Pi_{\hat H}^\perp h\|^2/n}
    \label{eqn:meanvardef}
\end{align}
then
\begin{align}
    g \disteq_\Bb \omega + \sigma \Pi_{\check H}^\perp y,\ \text{with $y \sim \Gaus(0, I_n)$}
    \label{eqn:gConditionedOnB}
\end{align}
For brevity, we will define the following matrices and vectors of fixed dimension
\begin{equation}
\begin{aligned}
    \hat \Lambda
        &\defeq
            \hat H^\trsp \hat H/n \in \R^{r \times r}
            &
    \check \Lambda
        &\defeq
            \check H^\trsp \check H/n \in \R^{s \times s}
            &
    \Gamma
        &\defeq
            \check G^\trsp \hat H/n \in \R^{s \times r}
            \\
    \hat \nu
        &\defeq
            \hat H^\trsp h/n \in \R^{r}
            &
    \check \nu
        &\defeq
            \check G^\trsp h /n \in \R^s
            .
\end{aligned}
    \label{eqn:momentMatrices}
\end{equation}

Suppose $\hat h^i$ was introduced by $\hat h^i := \hat \phi^i(g^1, \ldots, g^M)$, and $\check h^j$ was introduced by $\check h^j := \check \phi^j(g^1, \ldots, g^M)$, where $\hat \phi^i$ and $\check \phi^j$ depend at most on $g^1, \ldots, g^{\MM - 1}$.
By induction hypothesis \ref{IH:MomConv}$(\MM-1)$, $\hat \Lambda, \check \Lambda, \Gamma, \hat \nu, \check \nu$ all converge a.s. to corresponding limit values $\mathring{\hat \Lambda}, \mathring{\check \Lambda}, \mathring{\hat \nu}$, since their entries are moments of $Z^1, \ldots, Z^{\MM - 1}$:
\begin{align*}
    \hat \Lambda_{ij}
        &\asto
            \mathring{\hat \Lambda}_{ij}
        \defeq
            \EV \hat\phi^i(Z) \hat \phi^j(Z)
        =
            (\sigma_A)^{-2} \tSigma(\hat g^i, \hat g^j)
            \\
    \check \Lambda_{ij}
        &\asto
            \mathring{\check \Lambda}_{ij}
        \defeq
            \EV \check \phi^i(Z) \check \phi^j(Z)
        =
            (\sigma_A)^{-2} \tSigma(\check g^i, \check g^j)
            \\
    \hat \nu_i
        &\asto
            \mathring{\hat \nu}_i
        \defeq
            \EV \hat \phi^i(Z) \phi(Z)
        =
            (\sigma_A)^{-2} \tSigma(\hat g^i, g)
\end{align*}
and $\Gamma \asto 0, \check \nu \asto 0$ because, by the BP-like assumption, $\check G$ is odd in some input G-vars $v^1, \ldots, v^k$ independent from all other input G-vars, so that the limiting expectation is 0.

It turns out that, as a consequence of \cref{lemma:rankStability} below, a.s.\ for all large enough $n$, $\rank \hat \Lambda = \rank \mathring {\hat \Lambda}$ and $\rank \check \Lambda = \rank \mathring{\check \Lambda}$.
Therefore, as pseudoinverse is continuous on matrices of fixed rank,
we get the following proposition
\begin{prop}\label{prop:pseudoinverseLambda}
$\hat \Lambda^+ \asto \mathring{\hat \Lambda}^+$ and $\check \Lambda^+ \asto \mathring{\check \Lambda}^+$.
\end{prop}

Using this proposition, we compute the limits of the conditional mean $\omega$ and variance $\sigma^2$.
\begin{lemma}\label{lemma:sigmaConverges}
$\sigma^2 \asto \mathring \sigma^2 \defeq \tSigma(g, g) - \tSigma(g, \hat G) \tSigma(\hat G, \hat G)^+ \tSigma(\hat G, g)$
\end{lemma}
\begin{proof}
Note that
\begin{align*}
    \sigma^2 = \f {\sigma_A^2} n (h^\trsp h - h^\trsp \Pi_{\hat H} h)
        = \f {\sigma_A^2} n (h^\trsp h - h^\trsp {\hat H} (\hat H^\trsp \hat H)^+ \hat H^\trsp h)
        = \f {\sigma_A^2} n (h^\trsp h - \hat \nu^\trsp \hat \Lambda^+ \hat \nu).
\end{align*}
Because $\phi$ is polynomially-bounded, so is $\phi(z)^2$ as well.
By induction hypothesis,
\begin{align*}
    \f 1 n h^\trsp h = \f 1 n \sum_{\alpha = 1}^n \phi(g^1_\alpha, \ldots, g^{\MM - 1}_\alpha)^2
    \asto\EV_{Z \sim \Gaus(\tmu, \tSigma)}
        \phi(Z)^2 = 
    \sigma_A^{-2} \tSigma(g, g).
\end{align*}
Likewise, $\hat \nu \asto \mathring{\hat \nu}$ and $\hat \Lambda \asto \mathring{\hat \Lambda}$.
By \cref{prop:pseudoinverseLambda}, $\hat \Lambda^+ \asto \mathring{\hat \Lambda}^+$.
Combining all of these limits together yields the desired claim.
\end{proof}

\begin{lemma}\label{lemma:omegaExpansion}
Let $v \defeq \hat \Lambda^+ \hat \nu$, so that $v \asto \mathring v \defeq \mathring{\hat \Lambda}^+ \mathring{\hat \nu}.$
Then for some vector $\hat \varepsilon \in \R^r, \check \varepsilon \in \R^s$ that go to 0 a.s.\ with $n$, $\omega = Eh = \hat G(\mathring v + \hat \varepsilon) + \check H \check \varepsilon$
\end{lemma}

\begin{proof}
Using \cref{eqn:momentMatrices,eqn:Eexpansion}, we can re-express $\omega$ as
\begin{align*}
\omega
    &=
        \hat G \hat \Lambda^+ \hat \nu
        + \check H \check \Lambda^+ \check \nu
        - \check H \check \Lambda^+ \Gamma \hat \Lambda^+ \hat \nu
        .
\end{align*}
Because $\Gamma \asto 0, \check \nu \asto 0$ as discussed above, we can set $\check \varepsilon \defeq \check \Lambda^+ \check \nu - \check \Lambda^+ \Gamma \hat \Lambda^+ \hat \nu$, so that $\check \varepsilon \asto 0$.

In addition, by \cref{prop:pseudoinverseLambda}, $\hat \Lambda^+ \asto \mathring{\hat \Lambda}^+$, so that setting $\hat \varepsilon \defeq v - \mathring v $, we get $\hat \varepsilon \asto 0$.

Altogether, we have
\begin{equation*}
\omega = \hat G(\mathring v + \hat \varepsilon) + \check H \check \varepsilon
\end{equation*}
as desired.
\end{proof}

\subsection{Rank Stability and Zero Stability}
\label{sec:rankStabilityZeroStability}
In this section, we prove the following consequence of \ref{IH:coreSet}$(\MM-1)$ and \ref{IH:MomConv}$(\MM-1)$.

\begin{lemma}[Rank Stability]\label{lemma:rankStability}
For any collection of polynomially-bounded functions $\{\psi_j: \R^{\MM-1} \to \R\}_{j=1}^l$, let $K \in \R^{l \times l}$ be the random matrix (depending on $n$) defined by
\begin{equation*}
K_{ij} = \f 1 n \sum_{\alpha=1}^n \psi_i(g^1_\alpha, \ldots, g^{\MM-1}_\alpha) \psi_j(g^1_\alpha, \ldots, g^{\MM-1}_\alpha).
\end{equation*}
By \ref{IH:MomConv}$(\MM-1)$,
\[K \asto \mathring K\]
for some matrix $\mathring K \in \R^{l \times l}$.
\begin{enumerate}
\item
    Then, almost surely, for large enough $n$,
    \begin{equation*}
    \ker K = \ker \mathring K, \quad \im K = \im \mathring K, \quad\text{and}\quad \rank K = \rank \mathring K.
    \end{equation*}
    Here $\ker$ denotes null space and $\im$ denotes image space.
\item
    Suppose $I \sbe [l]$ is any subset such that $\mathring K|_I$, the restriction of $\mathring K$ to rows and columns corresponding to $I$, satisfies
    \[|I| = \rank \mathring K|_I = \rank \mathring K.\]
    There are unique coefficients $\{F_{ij}\}_{i \in [l], j \in I}$ that expresses each row of $\mathring K$ as linear combinations of rows corresponding to $I$:
    \[
    \forall i \in [l],\quad 
    \mathring K_i = \sum_{j \in I} F_{ij} \mathring K_j.
    \]
    Then, a.s.\ for all large $n$, for all $\alpha \in [n]$,
    \begin{equation*}
    \psi_i(g^1_\alpha, \ldots, g^{\MM-1}_\alpha)
    = \sum_{j \in I} F_{ij} \psi_j(g^1_\alpha, \ldots, g^{\MM-1}_\alpha).
    \end{equation*}
\end{enumerate}

\end{lemma}

This will be primarily a corollary of the following \cref{lemma:zerofunStability}.

\begin{lemma}[Zero Stability]\label{lemma:zerofunStability}
If $\psi: \R^{\MM-1} \to \R^{\ge 0}$ is a nonnegative function such that
\begin{align*}
\f 1 n \sum_{\alpha = 1}^n \psi(g^1_\alpha, \ldots, g^{\MM-1}_\alpha) \asto 0
\end{align*}
then, almost surely, for large enough $n$,
\[\psi(g^1_\alpha, \ldots, g^{\MM-1}_\alpha) = 0\]
for all $\alpha \in [n]$.
\end{lemma}

We give the proof of \cref{lemma:rankStability} now, assuming \cref{lemma:zerofunStability}.
\begin{proof}
Let $v \in \R^l$ be in the null space of $\mathring K$, i.e. $v^\trsp \mathring K v = 0$.
Then we also have $v^\trsp K v \asto v^\trsp \mathring K v = 0$.
But
\begin{align*}
v^\trsp K v
    &=
        \f 1 n \sum_{\alpha=1}^n \Psi(g^1_\alpha, \ldots, g^{\MM-1}_\alpha)
        ,
        \quad
        \text{where}
        \quad
        \Psi(g^1_\alpha, \ldots, g^{\MM-1}_\alpha) \defeq
        \lp
            \sum_{i=1} v_i \psi_i(g^1_\alpha, \ldots, g^{\MM-1}_\alpha)
        \rp^2
\end{align*}
and $\Psi$ is a nonnegative function.
By \cref{lemma:zerofunStability}, we have that: almost surely, for large enough $n$, 
\[\Psi(g^1_\alpha, \ldots, g^{\MM-1}_\alpha) = 0 \quad \text{for all $\alpha \in [n]$}
\quad
\implies v^\trsp K v = 0\]
\textit{Claim 1.}\ \ 
If we apply this argument to a basis $\{v^1, \ldots, v^t\}$ of $\ker \mathring K$, then we get, 
\[\text{a.s.\ for all large $n$,}\quad 
\ker \mathring K \sbe \ker K,\]
so that
\[\text{a.s.\ for all large $n$,}\quad 
\rank \mathring K \ge \rank K.\]
Because the rank function is lower semicontinuous (i.e.\ the rank can drop suddenly, but cannot increase suddenly), and $K \asto \mathring K$, we also have
\[\text{a.s.\ for all large $n$,}\quad 
\rank \mathring K \le \rank K.\]
Combined with the above, this gives the desired result on rank.
The equality of null space then follows from the equality of rank, and the equality of image space follows immediately, as the image space is the orthogonal complement of the null space.

\textit{Claim 2.}\ \ 
If we apply the above argument to each $v^i$ defined by inner product as
\[\forall x \in \R^l,\quad x^\trsp v^i \defeq x_i - \sum_{j \in I} F_{ij} x_j,
\]
(note that only for $i \not \in I$ is $v^i$ nonzero),
then we have, a.s.\ for large $n$, $v^i{}^\trsp K v^i = 0$, or
\begin{equation*}
    \psi_i(g^1_\alpha, \ldots, g^{\MM-1}_\alpha)
    = \sum_{j \in I} F_{ij} \psi_j(g^1_\alpha, \ldots, g^{\MM-1}_\alpha).
\end{equation*}
\end{proof}

In the rest of this section, we prove \cref{lemma:zerofunStability}.
It helps to first show that the linear relations given in \ref{prop:basis} carries over to the $n\to\infty$ limit.
\begin{prop}\label{prop:limitbasis}
Let $\tSigma|_\coreset$ be the submatrix of $\tSigma$ with rows and columns corresponding to $\{g^i: i \in \coreset\}$.
Then $\rank \tSigma = \rank \tSigma|_\coreset = |\coreset|$.
Furthermore, if $Z = (Z^1, \ldots, Z^{\MM-1}) \sim \Gaus(\tmu|_{\MM-1}, \tSigma|_{\MM-1})$, where $\tmu|_{\MM-1}, \tSigma|_{\MM-1}$ are the restrictions of $\tmu, \tSigma$ to $g^1, \ldots, g^{\MM-1}$, then for each $i$,
\[Z^i \disteq \sum_{j \in \coreset} a_j Z^j\]
where $\{a_j\}_{j \in \coreset}$ are the coefficients corresponding to $g^i$ given in \ref{prop:basis}.
\end{prop}
\begin{proof}
By \ref{prop:basis} property, each $g^i, i \in \coreset$, has a set of unique constants $\{a_j\}_{j \in \coreset}$ (independent of $n$) such that, almost surely, for large enough $n$,
\[g^i = \sum_{j \in \coreset} a_j g^j.\]
Let $\psi(x^1, \ldots, x^{\MM-1}) \defeq (x^i - \sum_{j \in \coreset} a_j x^j)^2$.
Then by \ref{prop:basis}$(\MM-1)$ and \ref{IH:MomConv}$(\MM-1)$,
\begin{align*}
\f 1 n \sum_{\alpha=1}^n \psi(g^1_\alpha, \ldots, g^{\MM-1}_\alpha) \asto \EV_{Z \sim \Gaus(\tmu|_{\MM-1}, \tSigma|_{\MM-1})} \psi(Z) = 0.
\end{align*}
where $\tmu|_{\MM-1}, \tSigma|_{\MM-1}$ are the restrictions of $\tmu, \tSigma$ to $g^1, \ldots, g^{\MM-1}$.
This implies that for $Z = (Z^1, \ldots, Z^{\MM-1}) \sim \Gaus(\tmu|_{\MM-1}, \tSigma|_{\MM-1})$,
\[Z^i \disteq \sum_{j \in \coreset} a_j Z^j.\]
Repeating this argument for all $i \in [{\MM-1}]$ implies that $\{Z^j\}_{j \in \coreset}$ is a ``spanning set'' of $Z^1, \ldots, Z^{\MM-1}$.
Furthermore, by the uniqueness of the coefficients, we also have that $\{Z^j\}_{j \in \coreset}$ is linearly independent as well.
This then implies the rank consequence we want.
\end{proof}

Now we show \cref{lemma:zerofunStability}.

\begin{proof}[Proof of \cref{lemma:zerofunStability}]
By \ref{IH:MomConv}$(\MM-1)$,
\begin{align*}
\f 1 n \sum_{\alpha = 1}^n \psi(g^1_\alpha, \ldots, g^{\MM-1}_\alpha) \to \EV_{Z \sim \Gaus(\tmu|_{\MM-1}, \tSigma|_{\MM-1})} \psi(Z).
\end{align*}
By \cref{prop:limitbasis}, if $Z \sim \Gaus(\tmu|_{\MM-1}, \tSigma|_{\MM-1})$ and $Z|_\coreset$ is the part of $Z$ corresponding to $\coreset$, then 
\begin{description}
\item[$Z|_\coreset$ has density.]
    The law of $Z|_\coreset$ (namely $\Gaus(\tmu|_\coreset, \tSigma|_\coreset)$, where $\tmu|_\coreset, \tSigma|_\coreset$ are the restriction of $\tmu$ and $\tSigma$ to $\coreset$) is absolutely continuous against the Lebesgue measure of $\R^\coreset$ and vice versa, so that a set of Lebesgue measure zero is measure zero under $\Gaus(\tmu|_\coreset, \tSigma|_\coreset)$, and vice versa; and
\item[$Z|_\coreset$ is basis of $Z$.]
    \ref{prop:basis} yields a linear function $\lambda$ such that $\lambda(\{g^j_\alpha\}_{j \in \coreset}) = \{g^i_\alpha\}_{i=1}^{m-1}$ for all $\alpha$, almost surely asymptotically, and
    $\lambda(Z|_\coreset) \disteq Z$, so that
    \begin{equation*}
    \EV_{Z \sim \Gaus(\tmu|_{\MM-1}, \tSigma|_{\MM-1})} \psi(Z) = \EV_{Z' \sim \Gaus(\tmu|_\coreset, \tSigma|_\coreset)} \psi \circ \lambda(Z').
    \end{equation*}
    This expectation is 0 by our premise.
\end{description}
Because $\psi$, and thus $\psi \circ \lambda$, is a nonnegative function, the nullity of the expectation implies that, other than a set $U$ of $\Gaus(\tmu|_\coreset, \tSigma|_\coreset)$-measure zero, $\psi\circ \lambda$ is 0.
This set $U$ also has Lebesgue measure zero as $Z|_\coreset$ has density, by our reasoning above.

If in \ref{prop:nullAvoid}, we set $A_{n\alpha} = U$ for all $n$ and all $\alpha \in [n]$, then we get that: almost surely, for all large enough $n$, for all $\alpha \in [n]$,

\begin{equation*}
\{g^i_\alpha\}_{i \in \coreset} \not\in U
\iff
\psi\circ \lambda(\{g^i_\alpha\}_{i \in \coreset}) = 0
\iff
\psi(g^1_\alpha, \ldots, g^{\MM-1}_\alpha) = 0,
\end{equation*}
as desired.
\end{proof}

\subsection{Inductive Step: \ref{IH:coreSet}\texorpdfstring{$(\MM)$}{(m)}}
\label{sec:inductiveCoreSet}
In this section, we show
\begin{equation*}
\text{\ref{IH:MomConv}$(\MM-1)$ and \ref{IH:coreSet}$(\MM-1)$} \implies \text{\ref{IH:coreSet}}(\MM).
\end{equation*}
More explicitly, we need to think about whether to add $\MM$ to the core set $\coreset$ of $[\MM-1]$ in order to maintain the \ref{prop:basis} and \ref{prop:nullAvoid} properties.

We proceed by casework on whether $\mathring \sigma = 0$.

\newcommand{\Lsq}{\mathcal{L}}
\subsubsection{If \texorpdfstring{$\mathring \sigma = 0$}{sigma converges to 0 a.s.}.}

We will show that the core set properties are maintained if we don't add $m$ to the core set.

Consider the space $\Lsq \defeq L^2(\Gaus(\tmu|_\coreset, \tSigma|_\coreset))$ of square-integrable real functions against the measure $\Gaus(\tmu|_\coreset, \tSigma|_\coreset)$ defined on $\R^{\coreset}$.
Let $\la \phi, \psi \ra = \EV_{Y \sim \Gaus(\tmu|_\coreset, \tSigma|_\coreset)} \phi(Y) \psi(Y)$ be the inner product of this space.
Just like in a finite-dimensional inner product space, given a finite collection of functions $S = \{\psi^i\}_{i=1}^k$, the orthogonal projection operator $\Pi_S$ to the span of $S$ (inside $\Lsq$) is given by
\[\Pi_S \phi = \sum_{i=1}^k a_i \psi^i,
\]
for any $\phi \in \Lsq$, where
\begin{align*}
a &= \Lambda^+ b \in \R^k,\\
b_j &= \la \psi^j, \phi \ra, b \in \R^k,\\
\Lambda_{ij} &= \la \psi^i, \psi^j\ra, \Lambda \in \R^{k \times k}.
\end{align*}

Recall that $g = Ah$ where $h$ was introduced by $h := \phi(g^{1}, \ldots, g^{\MM - 1})$, for some polynomially-bounded $\phi$, and likewise $\hat g^i = A \hat h^i$ where $\hat h^i = \hat \phi^i(g^1, \ldots, g^{\MM-1})$, for each $i \in [r]$.
By \ref{prop:basis}, we know that, a.s.\ for large enough $n$, each of $g^1, \ldots, g^{\MM-1}$ is a (unique, constant-in-$n$) linear combination of $\{g^j\}_{j \in \coreset}$.
Therefore, we can express
\[h = \underline\phi(\{g^j\}_{j \in \coreset}),\quad\text{and}\quad
\forall i \in [r], \hat h^i = \underline {\hat \phi^i}(\{g^j\}_{j \in \coreset})
\]
for some functions $\underline \phi, \underline {\hat \phi^i} \in \Lsq$.
For convenience, set $S \defeq \{\underline{\hat \phi^i}\}_i.$

One can see then,
as in the proof of \cref{lemma:sigmaConverges},
\[
\mathring\sigma^2
    =
        \sigma_A^2 (\EV \phi(Z)^2 - \mathring{\hat \nu}^\trsp \mathring{\hat \Lambda}^+ \mathring{\hat \nu})
        \\
    =
        \sigma_A^2 (\la \underline\phi, \underline\phi \ra - \la \underline\phi, \Pi_{S} \underline \phi \ra)
\]
by expanding the definition of $\mathring{\hat \nu}$ and $\mathring{\hat \Lambda}$.
Therefore, $\mathring \sigma = 0$ implies that 
\[\la \underline\phi, \underline\phi \ra = \la \underline\phi, \Pi_{S} \underline \phi \ra
\]
so that: after changing its values on a set $U$ of measure zero under $\Gaus(\tmu|_\coreset, \tSigma|_\coreset)$ (and thus also under Lebesgue measure by \cref{lemma:rankStability}), $\underline \phi$ is a linear combination of $\{\underline{\hat \phi^i}\}_{i=1}^r$, i.e.
\begin{equation*}
\forall \vec x \not \in U, \underline\phi(\vec x) = \sum_{i \in [r]} c_i \underline{\hat \phi^i}(\vec x)
\end{equation*}
for some coefficients $\{c_i\}_{i \in [r]}$.
By \ref{prop:nullAvoid} applied to $A_{n\alpha} = U$ for all $n$ and $\alpha \in [n]$, we also have that: a.s.\ for large enough $n$,
\[
\phi(g^1, \ldots, g^\alpha) = \underline \phi(\{g^j\}_{j \in \coreset})
= \sum_{i \in [r]} c_i \underline{\hat \phi^i}(\{g^j\}_{j \in \coreset}) 
= \sum_{i \in [r]} c_i \hat \phi^i(g^1, \ldots, g^\alpha)
,
\]
and therefore, under the same condition, (recall $A$ is the matrix giving rise to $g$ in $g:= A h$)
\[
g = A \phi(g^1, \ldots, g^\alpha)
= \sum_{i \in [r]} c_i A \hat \phi^i(g^1, \ldots, g^\alpha)
= \sum_{i \in [r]} c_i \hat g^i.
\]
This shows that, if we keep the core set as $\coreset$, then \ref{prop:basis} is still satisfied.
Since the core set is not changing, \ref{prop:nullAvoid} just follows from the induction hypothesis.

For usage later in the proof of \ref{IH:MomConv}$(\MM)$, we record our observation here as follows
\begin{lemma}\label{lemma:limitSigmaIsZero}
If $\mathring \sigma = 0$, then there are coefficients $\{c_i\}_{i=1}^r$ independent of $n$ such that a.s.\ for large enough $n$,
\[
g = \sum_{i \in [r]} c_i \hat g^i.
\]
\end{lemma}

\subsubsection{If \texorpdfstring{$\mathring \sigma > 0$}{sigma converges to nonzero value a.s.}.}
\label{sec:sigmalimit>0}

It's clear that $g$ cannot be in the linear span of $\{\hat g^i\}_{i\in[r]}$ asymptotically, so we will add $g$ to the core set, and the \ref{prop:basis} property follows immediately.
In the below, we shall write $\coreset$ for the old core set, and $\coreset' \defeq \coreset \cup \{g\}$ for the new one.

It remains to show \ref{prop:nullAvoid} for $\coreset'$.
First, let's assume that, a.s.\ for large enough $n$, $\Pi_{\check H}^\perp$ has no zero diagonal entry;
we shall show this fact below in \cref{lemma:PiCheckHDiagonal}.
Because the conditional variance of $g^\MM_\alpha$ given $g^1, \ldots, g^{\MM-1}$ is $\sigma^2 {(\Pi_{\check H}^\perp)_{\alpha\alpha}}$, and because $\mathring \sigma > 0$, this assumption implies that, a.s.\ for all large enough $n$,
\begin{equation}
\text{$g^\MM_\alpha|g^1, \ldots, g^{\MM-1}$ has density for all $\alpha \in [n]$.}
\label{eqn:conditionalDistributionHasDensity}
\end{equation}
By ``has density'' here, we in particular mean that any Lesbegue measure zero set in $\R$ has zero probability under the conditional distribution of $g^\MM_\alpha$ given $g^1, \ldots, g^{\MM-1}$.

Now, assuming \cref{lemma:PiCheckHDiagonal}, we prove \ref{prop:nullAvoid} holds for $\coreset'$.

Let $\{A_{n\alpha} \sbe \R^{\coreset'}\}_{n\in\N, \alpha \in [n]}$ be a triangular array of Lesbegue measure zero sets.
For each $A_{n \alpha}$, define $B_{n\alpha} \defeq \{\vec x \in \R^{\coreset}: \lambda(A_{n\alpha}|_{\vec x}) \ne 0\}$, where $A_{n\alpha}|_{\vec x} = \{y \in \R: (\vec x, y) \in A_{n\alpha} \sbe \R^{\coreset} \times \R\}$ is the ``slice'' of $A_{n\alpha}$ at $\vec x$, and $\lambda$ is the 1-dimensional Lebesgue measure.
Because each $A_{n\alpha}$ has measure zero in $\R^{\coreset'}$, necessarily each $B_{n\alpha}$ also has measure zero in $\R^{\coreset}$.
Applying \ref{prop:nullAvoid} to the triangular array $\{B_{n\alpha} \sbe \R^{\coreset}\}_{n \in \N, \alpha \in [n]}$, we get that: a.s.\ for large enough $n$,
\[
\forall \alpha \in [n], \{g^i_\alpha\}_{i \in \coreset} \not \in B_{n\alpha}.\]
Therefore, by \cref{eqn:conditionalDistributionHasDensity}, a.s.\ for large enough $n$,
\[
\forall \alpha \in [n], \{g^i_\alpha\}_{i \in \coreset'} \not \in A_{n\alpha}.
\]
This finishes the proof of \ref{prop:nullAvoid} for $\coreset'$, and also \ref{IH:coreSet}$(\MM)$, save for \cref{lemma:PiCheckHDiagonal} below.

\begin{lemma}\label{lemma:PiCheckHDiagonal}
Almost surely, for large enough $n$, $\Pi_{\check H}^\perp$ has no zero diagonal entry.
\end{lemma}
\begin{proof}

WLOG, assume $\mathring{\check \Lambda}$ is full rank.
Otherwise, by \cref{lemma:rankStability}(2), we can replace $\check h^1,\ldots, \check h^s$ by a linearly independent spanning set $\check h^{i_1}, \ldots, \check h^{i_k}$ such that 1) each $\check h^j$ is almost surely, for all large $n$, a linear combination of them and such that 2) their 2nd moment matrix is full rank in the limit.
Then the projection matrix associated to $\check h^{i_1}, \ldots, \check h^{i_k}$ is, almost surely, for all large $n$, the same as $\Pi_{\check H}$.

By the Sherman-Morrison formula (\cref{fact:ShermanMorrison}),
\begin{equation*}
(\Pi_{\check H})_{\alpha\alpha}
    =
        f\lp \f 1 n \check h_\alpha{}^\trsp \check \Lambda_{-\alpha}^{-1} \check h_\alpha\rp
\end{equation*}
where $f(x) = x/(1+x)$, $\check h_\alpha$ is the column vector $(\check h^1_\alpha, \ldots, \check h^s_\alpha)^\trsp$,
and $\check \Lambda_{-\alpha} = \f 1 n \sum_{\beta \ne \alpha} \check h_\beta \check h_\beta{}^\trsp$.
Thus, unless $\check \Lambda_{-\alpha}$ is singular for some $\alpha$, all diagonal entries of $\Pi_{\check H}^\perp = I - \Pi_{\check H}$ are nonzero.
So it suffices to show that, 
\begin{equation*}
\text{a.s.\ for large enough $n$, \quad $\check \Lambda_{-\alpha}$ is nonsingular for all $\alpha$.}
\end{equation*}

To do this, it pays to note that $\check \Lambda_{-\alpha} = \check \Lambda - \f 1 n \check h_\alpha \check h_\alpha^\trsp$, so that
\[|\lambda_{\mathrm{min}}(\check \Lambda_{-\alpha}) - \lambda_{\mathrm{min}}(\check \Lambda)| \le \|\f 1 n \check h_\alpha \check h_\alpha^\trsp\|_{\mathrm{op}} = \f 1 n \check h_\alpha^\trsp \check h_\alpha.\]

By \cref{lemma:maxbound} below (which bounds the max by a high moment),
\[\max_{\alpha \in n} \f 1 n \check h_\alpha^\trsp \check h_\alpha \asto 0,\]
and consequently
\[\max_{\alpha \in [n]} |\lambda_{\mathrm{min}}(\check \Lambda_{-\alpha}) - \lambda_{\mathrm{min}}(\check \Lambda)| \asto 0.\]
Because $\check \Lambda \asto \mathring{\check \Lambda}$, we know that, a.s.\ for large enough $n$, $\lambda_{\mathrm{min}}(\check \Lambda)$ is bounded away from 0 by a constant (independent of $n$).
Altogether, this implies that all $\check \Lambda_{-\alpha}$ are nonsingular, as desired.
\end{proof}

\begin{fact}[Sherman-Morrison formula]\label{fact:ShermanMorrison}
For any nonsingular matrix $A \in \R^{l \times l}$ and vector $a \in \R^l$,
we have
\[a^\trsp (A + aa^\trsp)^{-1} a = \f{a^\trsp A^{-1} a}{1 + a^\trsp \inv A a}.
\]
Consequently, for any full rank matrix $H$, the $\alpha$th diagonal entry of its associated projection matrix $\Pi_H = H (H^\trsp H)^{-1} H^\trsp$ can be written as
\[(\Pi_H)_{\alpha\alpha} = \f{H_\alpha (H_{-\alpha}^\trsp H_{-\alpha})^{-1} H_\alpha^\trsp}{1 + H_\alpha (H_{-\alpha}^\trsp H_{-\alpha})^{-1} H_\alpha^\trsp}
\]
where $H_\alpha$ is the $\alpha$th row of $H$, and $H_{-\alpha}$ is $H$ with the $\alpha$th row removed.
\end{fact}

\begin{lemma}\label{lemma:maxbound}
Assume \ref{IH:MomConv}$(\MM-1)$.
Suppose $\psi: \R^{\MM-1} \to \R$ is polynomially bounded.
Then as $n \to \infty,$
\[\f 1 {n^p} \max_{\alpha \in [n]} |\psi(g^1_\alpha, \ldots, g^{\MM-1}_\alpha)| \asto 0\]
for any $p > 0$.
\end{lemma}
\begin{proof}
For any $q > 0$, we have the elementary bound
\begin{equation*}
\max_{\alpha \in [n]} |\psi(g^1_\alpha, \ldots, g^{\MM-1}_\alpha)|
\le
\sqrt[q]{\sum_{\alpha \in [n]}
|\psi(g^1_\alpha, \ldots, g^{\MM-1}_\alpha)|^q}.
\end{equation*}
Thus, for any $q > 0$,
\begin{align*}
\f 1 {n^p} \max_{\alpha \in [n]} |\psi(g^1_\alpha, \ldots, g^{\MM-1}_\alpha)|
&\le
    \f 1 {n^{p-1/q}} 
    \sqrt[q]{\f 1 n
        \sum_{\alpha \in [n]}
        |\psi(g^1_\alpha, \ldots, g^{\MM-1}_\alpha)|^q}.
\end{align*}
Because, by \ref{IH:MomConv}$(\MM-1)$, $\f 1 n
        \sum_{\alpha \in [n]}
        |\psi(g^1_\alpha, \ldots, g^{\MM-1}_\alpha)|^q \asto C$ for some constant $C$ as $n \to \infty$,
the RHS above converges a.s.\ to 0 as soon as we take $q > 1/p$, and therefore so does the LHS.

\end{proof}

\subsection{Inductive Step: \ref{IH:MomConv}\texorpdfstring{$(\MM)$}{(m)}}
\label{sec:inductiveMoments}
In this section, we show
\begin{equation*}
\text{\ref{IH:MomConv}$(\MM-1)$ and \ref{IH:coreSet}$(\MM-1)$} \implies \text{\ref{IH:MomConv}}(\MM).
\end{equation*}

More specifically,
we will show that for any polynomially-bounded $\psi: \R^{\MM} \to \R$,
\begin{align*}
    \f 1 n \sum_{\alpha=1}^n \psi(g^1_\alpha, \ldots, g^{\MM}_\alpha) 
    \asto
    \EV_{Z \sim \Gaus(\tmu, \tSigma)} \psi(Z)
\end{align*}
where again on the RHS $\psi$ ignores all coordinates $Z^{\MM +1},\ldots, Z^M$ (corresponding to $g^{\MM +1}, \ldots, g^{M}$).

By \cref{lemma:limitSigmaIsZero}, if $\mathring \sigma = 0$, then almost surely, for large enough $n$, $g = g^\MM$ is just a (fixed) linear combination of $g^1, \ldots, g^{\MM-1}$, so \ref{IH:MomConv} is trivially true.
Therefore, in the below, we assume 
\begin{equation}
\mathring \sigma > 0.
\label{assm:mathringSigmaPositive}
\tag{$\star$}
\end{equation}
This assumption will be crucial for our arguments involving smoothness induced by Gaussian averaging.

\newcommand{\probA}{\mathsf{A}}
\newcommand{\probB}{\mathsf{B}}
\newcommand{\probC}{\mathsf{C}}
\newcommand{\probD}{\mathsf{D}}

\newcommand{\EVbr}[2][]{\EV_{#1}\left[#2\right]}
\newcommand{\EVcond}[2]{\EV\left[\left.#1\right| #2 \right]}
\newcommand{\extcom}[1]{{\color{blue}{#1}}}
\newcommand{\exttcom}[1]{{\color{olive}{#1}}}

To clarify notation in the following, we will write $\EVbr[X]{expression}$ to denote the expectation over only the randomization in $X$, and $\EVcond{expression}{\Bb}$ to denote the expectation taken over all randomness except those in $\Bb$.

\paragraph{Proof Plan}
Note that
\begin{align*}
    &\phantomeq
        \left|\f 1 n \sum_{\alpha = 1}^n \psi(g^1_\alpha, \ldots, g^{\MM}_\alpha)
        - \EV_{Z \sim \Gaus(\tmu, \tSigma)} \psi(Z)
        \right|
    \le
        \probA + \probB + \probC
\end{align*}
where
\begin{align*}
    \probA
        &\defeq
            \left|\f 1 n \sum_{\alpha = 1}^n \psi(g^1_\alpha, \ldots, g^{\MM}_\alpha)
            - \EV_z\psi\left(g^1_\alpha, \ldots, g^{\MM-1}_\alpha, \omega_\alpha + \sigma z \sqrt{(\Pi^\perp_{\check H})_{\alpha\alpha}}\right)\right|
            \\
    \probB
        &\defeq
            \left|
            \f 1 n \sum_{\alpha = 1}^n \EV_z\psi\left(g^1_\alpha, \ldots, g^{\MM-1}_\alpha, \omega_\alpha + \sigma z \sqrt{(\Pi^\perp_{\check H})_{\alpha\alpha}}\right)
            -
            \EV_z
                {
                    \psi\lp
                        g^1_\alpha, \ldots, g^{\MM-1}_\alpha,
                        \sum_{i=1}^r \mathring v_i \hat g^i_\alpha + \mathring \sigma z
                        \rp
                }
            \right|
            \\
    \probC
        &\defeq
            \left|
            \f 1 n \sum_{\alpha = 1}^n
            \EV_z
                {
                    \psi\lp
                        g^1_\alpha, \ldots, g^{\MM-1}_\alpha,
                        \sum_{i=1}^r \mathring v_i \hat g^i_\alpha + \mathring \sigma z
                        \rp
                }
            -
            \EV_{Z \sim \Gaus(\tmu, \tSigma)} \psi(Z)
            \right|
\end{align*}
with $z  \sim \Gaus(0, 1)$.
Note that $\probB$ and $\probC$ are random variables in $\Bb$.
We will show that each of $\probA, \probB, \probC$ goes to 0 almost surely, which would finish the proof of \cref{thm:netsorTMasterTheoremFormal}.

Roughly speaking, $\probA \asto 0$ because of a law of large number, $\probB \asto 0$ because of the smoothness in $\EV_z \psi$ induced by Gaussian averaging, and $\probC \asto 0$ by induction hypothesis.
We start with the last item, since it's the easiest.

\newcommand{\step}{\mathrm{step}}

\subsubsection{\texorpdfstring{$\probC$}{C} Converges Almost Surely to 0}

In this section we show that $\probC \asto 0$ by a straightforward reduction to the inductive hypothesis.

Let $\hat Z^1, \ldots, \hat Z^r$ be the components of $Z \sim \Gaus(\tmu, \tSigma)$ corresponding to $\hat g^1, \ldots, \hat g^r$, and let $\hat Z$ be the column vector with these entries.
Note that, by \cref{prop:GaussianCondition}, $Z^{\MM}$ (corresponding to $g^{\MM}$), conditioned on $Z^1, \ldots, Z^{\MM-1}$, is distributed as a Gaussian with mean $
\tSigma(g, \hat G)\tSigma(\hat G, \hat G)^+  \hat Z
    = \mathring{\hat \nu}^\trsp \mathring{\hat \Lambda}^+  \hat Z
    = \mathring v^\trsp \hat Z$
and variance
$\tSigma(g, g) - \tSigma(g, \hat G) \tSigma(\hat G, \hat G)^+ \tSigma(\hat G, g)
    = \mathring \sigma$.
Thus
\begin{align*}
\EV_{Z} \psi(Z)
    &=
        \EV_{Z^1, \ldots, Z^{\MM - 1}} \EV[\psi(Z) | Z^1, \ldots, Z^{\MM - 1}]
        \\
    &=
        \EV_{Z^1, \ldots, Z^{\MM - 1}}
        \EV_{z \sim \Gaus(0, 1)}\psi(Z^1,\ldots, Z^{\MM - 1}, \mathring v^\trsp \hat Z + \mathring \sigma z)
        \\
    &=
        \EV_{Z^1, \ldots, Z^{\MM - 1}}
        \Psi(Z^1, \ldots, Z^{\MM - 1})
\end{align*}
where we have set $\Psi(Z^1, \ldots, Z^{\MM - 1}) \defeq \EV_{z \sim \Gaus(0, 1)}\psi(Z^1,\ldots, Z^{\MM - 1}, \mathring v^\trsp \hat Z + \mathring \sigma z)$.
$\Psi$ is a polynomially bounded function since $\psi$ is.
Applying the induction hypothesis to $\Psi$, we obtain
\begin{align*}
    &\phantomeq
        \f 1 n \sum_{\alpha = 1}^n
            \EV_z
                {
                    \psi\lp
                        g^1_\alpha, \ldots, g^{\MM-1}_\alpha,
                        \sum_{i=1}^r \mathring v_i \hat g^i_\alpha + \mathring \sigma z
                        \rp
                }
        \\
    &=
        \f 1 n \sum_{\alpha = 1}^n
                {
                    \Psi\lp
                        g^1_\alpha, \ldots, g^{\MM-1}_\alpha
                        \rp
                }
        \\
    &\asto
        \EV_{Z^1, \ldots, Z^{\MM-1}} \Psi(Z^1, \ldots, Z^{\MM-1})
        \\
    &\pushright{\text{by induction hypothesis}}
        \\
    &=
        \EV_{Z^1, \ldots, Z^{\MM - 1}}
        \EV_{z \sim \Gaus(0, 1)}\psi(Z^1,\ldots, Z^{\MM - 1}, \mathring{v}^\trsp \hat Z + \mathring \sigma z)
        \\
    &=
        \EV_Z \psi(Z)
\end{align*}
as desired.

\subsubsection{\texorpdfstring{$\probA$}{A} Converges Almost Surely to 0}
\label{sec:probA}

In this section we show $\probA \asto 0.$

\renewcommand{\rho}{\lambda}
For each $\alpha \in [n]$, let $\psi_\alpha (x) \defeq \psi(g^1_\alpha, \ldots, g^{\MM-1}_\alpha, \omega_\alpha + \sigma x)$, with $\omega$ and $\sigma$ defined in \cref{eqn:meanvardef}.
This is a random function depending on the randomness of $g^1_\alpha, \ldots, g^{\MM-1}_\alpha$, and it changes with $n$ as well.
Also consider the ``centered version'' of $\psi_\alpha,$ $\tilde \psi_\alpha(x) \defeq \psi_\alpha(x) - \EV\psi_\alpha(x')$ with expectation taken over $x' \sim \Gaus(0, (\Pi^\perp_{\check H})_{\alpha\alpha})$ (but not $g^1_\alpha, \ldots, g^{\MM-1}_\alpha$).
Note by \cref{eqn:gConditionedOnB},
\[ \probA \disteq_\Bb \f 1 n \sum_{\alpha =1}^n \tilde \psi_\alpha(\xi_\alpha)\]
where $\xi \sim \Gaus(0, \Pi_{\check H}^\perp)$.

\paragraph{Proof idea.}
To prove our claim, we will show that, for almost all (i.e. probability 1 in $\basespace$) sequences of $(g^1, \ldots, g^{\MM-1}) = (g^1(n), \ldots, g^{\MM-1}(n))$ in $n$ --- which we shall call \emph{amenable sequences of $g^1, \ldots, g^{\MM-1}$} --- 
we have a moment bound
\begin{equation}
\EV[\probA^{2\rho} | \Bb] = 
\EV_{\xi \sim \Gaus(0, \Pi^\perp_{\check H})} 
    \lp \f 1 n \sum_{\alpha =1}^n \tilde \psi_\alpha(\xi_\alpha)\rp^{2\rho}
    < C n^{-1.25}
    \label{eqn:highMomentBound}
\end{equation}
for some large $\rho$ and some constant $C > 0$ depending only on $\rho$ and the particular sequence of $\{(g^1(n), \ldots, g^{\MM-1}(n))\}_n$.
Then we apply \cref{lemma:momentBoundASConvergence} to show that, conditioned on any amenable sequence, $\probA$ converges to 0 almost surely over all randomness remaining after conditioning.
Since almost all sequences are amenable, this shows that the convergence is also almost sure without the conditioning.

\paragraph{The moment bound.}
For $\rho \ge 6$ and any $q > 1$, we first apply \cref{thm:controlHighMoments} to get the bound
\begin{equation*}
\EV_{\xi} 
    \lp \f 1 n \sum_{\alpha =1}^n \tilde \psi_\alpha(\xi_\alpha)\rp^{2\rho}
    \le c n^{-1.5 + 1/q}
        \sqrt[q]{\f 1 n \sum_{\alpha=1}^n \EV \tilde \psi_\alpha(\xi_\alpha)^{2\rho q}}
\end{equation*}
where on both sides $\xi \sim \Gaus(0, \Pi^\perp_{\check H})$, and $c$ is a constant depending only on $\rho$ and $\MM$, but not on $n$, the functions $\psi_\alpha$, or $g^1, \ldots, g^{\MM-1}$.
To obtain \cref{eqn:highMomentBound}, we will show that
\begin{equation}
\f 1 n \sum_{\alpha=1}^n \EV \tilde \psi_\alpha(\xi_\alpha)^{2\rho q}
\label{eqn:moment}
\end{equation}
is uniformly bounded (in $n$), almost surely over the randomness of the sequences $\{g^1(n), \ldots, g^{\MM -1}(n)\}_n$.
We take all such sequences to be the \emph{amenable sequences}.
For $q > 4$, we then get the desired moment bound \cref{eqn:highMomentBound}.

It remains to show the almost sure uniform boundedness.

\paragraph{Almost sure uniform boundedness.}
Intuitively, \cref{eqn:moment} should converge almost surely to a deterministic value by applying some version of the induction hypothesis, so it should be almost surely uniformly bounded in $n$.
The obstacle is that $\xi_\alpha$ is not purely a function of $g^1_\alpha, \ldots, g^{\MM-1}_\alpha$, and \textit{a priori} it is not clear how to apply the induction hypothesis in a straightforward way.
We thus first process \cref{eqn:moment} a bit.
Let $\mu_\alpha \defeq \EV_{x \sim \Gaus(0, (\Pi^\perp_{\check H})_{\alpha\alpha})} \psi_\alpha(x)$.
Then, abbreviating $\EV$ for expectation taken over $\xi \sim \Gaus(0, \Pi^\perp_{\check H})$, we have the following inequalities of random variables in $\Bb$:
\begin{align*}
\f 1 n \sum_{\alpha=1}^n \EV \tilde \psi_\alpha(\xi_\alpha)^{2\rho q}
&=
    \f 1 n \sum_{\alpha=1}^n \EV (\psi_\alpha(\xi_\alpha) - \mu_\alpha)^{2\rho q}
    \\
&\le
    \f 1 n 2^{2\rho q - 1}
        \sum_{\alpha=1}^n
            \EV 
                \left[
                    \psi_\alpha(\xi_\alpha)^{2\rho q} + \mu_\alpha^{2\rho q}
                \right]
    \\
&\pushright{\text{by \cref{lem:powerbound}}}
    \\
&\le
    \f 1 n 2^{2\rho q}
        \sum_{\alpha=1}^n
            \EV 
                \psi_\alpha(\xi_\alpha)^{2\rho q}
    \\
&\pushright{
    \text{
        by power mean inequality $\mu_\alpha \le \sqrt[2\rho q]{\EV\psi_\alpha(\xi_\alpha)^{2\rho q}}$
        }
    }
    \\
&=
    \f 1 n 2^{2\rho q}
    \sum_{\alpha=1}^n
        \EV 
            \psi(g^1_\alpha, \ldots, g^{\MM-1}_\alpha, \omega_\alpha + \sigma \xi_\alpha )^{2\rho q}
    .
\end{align*}
Suppose, WLOG, that $\psi$ is polynomially bounded by an inequality $|\psi(x)| \le C \|x\|^p_p + c$ for some $p, C, c > 0$.
In the below, we will silently introduce constants $C_1, C_2, \ldots$ via \cref{lem:powerbound} and merge with old constants, such that they will only depend on $\rho, p, q$.
Continuing the chain of inequalities above
\begin{align*}
\f 1 n \sum_{\alpha=1}^n \EV \tilde \psi_\alpha(\xi_\alpha)^{2\rho q}
&\le
    c + 
    \f 1 n C 2^{2\rho q}
    \sum_{\alpha=1}^n
        \EV 
            \lp 
                |g^1_\alpha|^p + \ldots + |g^{\MM-1}_\alpha|^p + |\omega_\alpha + \sigma \xi_\alpha|^p
            \rp^{2\rho q}
    \\
&\le
    c + 
    \f 1 n C_1
    \sum_{\alpha=1}^n
        \EV 
            \lp 
                |g^1_\alpha|^p + \ldots + |g^{\MM-1}_\alpha|^p + |\omega_\alpha|^p + |\sigma \xi_\alpha|^p
            \rp^{2\rho q}
    \\
&\le
    c +
    \f 1 n C_2
    \sum_{\alpha=1}^n
        |g^1_\alpha|^{2 \rho q p} + \ldots + |g^{\MM-1}_\alpha|^{2 \rho q p} + |\omega_\alpha|^{2 \rho q p} +
        \EV 
             |\sigma \xi_\alpha|^{2 \rho q p}
    .
    \numberthis \label{eqn:momentUpperBoundDecomposition}
\end{align*}

We now proceed to show that the summands of \cref{eqn:momentUpperBoundDecomposition} are almost surely uniformly bounded, which finishes our proof of $\probA \asto 0$.

\begin{itemize}
    \item
        By induction hypothesis,
        \[
        \f 1 n
            \sum_{\alpha=1}^n
                \EV 
                    |g^1_\alpha|^{2 \rho q p} + \ldots + |g^{\MM-1}_\alpha|^{2 \rho q p}
        \]
        almost surely converges to a deterministic value, so that it is almost surely uniformly bounded in $n$.

    \item
        In addition, $\sigma \asto \mathring \sigma$, so that, almost surely, for large enough $n$, $\sigma \le \mathring \sigma + 1$.
        (The order of the qualifiers is important here; in general this statement cannot be made uniformly in $n$).
        Therefore, \emph{almost surely, for large enough $n$},
        \begin{align*}
        \f 1 n
            \sum_{\alpha=1}^n
            \EV |\sigma \xi_\alpha|^{2 \rho q p}
        &\le
            \f 1 n
            \sum_{\alpha=1}^n
                |\mathring \sigma + 1|^{2\rho q p}
                \EV |\xi_\alpha|^{2 \rho q p}
                .
        \end{align*}
        This is almost surely uniformly bounded in $n$ because $\Var(\xi_\alpha) = (\Pi_{\check H}^\perp)_{\alpha\alpha} \in [0, 1]$ for all $\alpha$ by \cref{lemma:projectionDiagonal}.
    \item
        It remains to bound $\f 1 n \sum_{\alpha=1}^n |\omega_\alpha|^{2\rho q p}$.
        We extract our reasoning here into the \cref{lemma:omegaAlphaUnifomBounded} below, as we will need to reuse this for later.
        This finishes the proof of $\probA\asto 0.$

\end{itemize}

\begin{lemma}\label{lemma:omegaAlphaUnifomBounded}
For any polynomially bounded function $\varphi: \R \to \R,$
\[
\f 1 n \sum_{\alpha=1}^n |\varphi(\omega_\alpha)|
\]
is almost surely uniformly bounded in $n$.
\end{lemma}
\begin{proof}
It suffices to prove this for $\varphi(x) = |x|^d$ for any $d > 0$.

Expanding $\omega$ according to \cref{lemma:omegaExpansion}, we get
\begin{align*}
    \f 1 n \sum_{\alpha=1}^n |\omega_\alpha|^{d}
    &=
        \f 1 n \sum_{\alpha=1}^n \left|
            \sum_{i=1}^r
                \hat g_\alpha^i (\mathring v_i + \hat \epsilon_i)
            + \sum_{j=1}^s
                \check h_\alpha^j \check \epsilon_j
        \right|^d
\end{align*}
for (fixed dimensional) $\hat \epsilon \in \R^r, \check \epsilon \in \R^s$ that go to 0 almost surely with $n$.
Applying \cref{lem:powerbound}, we get
\begin{align*}
    \f 1 n \sum_{\alpha=1}^n |\omega_\alpha|^d
    &\le
        \f 1 n C_3\sum_{\alpha=1}^n
            \left|
                \sum_{i=1}^r
                \hat g_\alpha^i \mathring v_i
            \right|^d
            +
            \left|
                \sum_{i=1}^r
                \hat g_\alpha^i \hat \epsilon_i
            \right|^d
            +
            \left|
                \sum_{j=1}^s
                \check h_\alpha^j \check \epsilon_j
            \right|^d
            .
\end{align*}
We bound each summand separately.
\begin{itemize}
    \item
        By induction hypothesis, 
        \begin{align*}
            \f 1 n \sum_{\alpha=1}^n
            \left|
                \sum_{i=1}^r
                \hat g_\alpha^i \mathring v_i
            \right|^d
        \end{align*}
        converges a.s.\ to a deterministic value, so it is a.s uniformly bounded in $n.$
    \item
        By the a.s.\ decaying property of $\hat \epsilon$, we have
        almost surely, for large enough $n$, $|\sum_{i=1}^r \hat g_\alpha^i \hat \epsilon_i| \le \sum_{i=1}^r |\hat g_\alpha^i|$ (again, the order of qualifier is very important here).
        By induction hypothesis,
        \begin{align*}
        \f 1 n \sum_{\alpha=1}^n
            \lp \sum_{i=1}^r |\hat g_\alpha^i|\rp^d
        \end{align*}
        converges a.s.\ to a deterministic value, yielding the a.s.\ uniformly-boundedness of it and of
        \begin{align*}
        \f 1 n \sum_{\alpha=1}^n
            \left| \sum_{i=1}^r \hat g_\alpha^i \hat \epsilon_i \right|^d.
        \end{align*}
    \item
        Likewise, because for each $j$, $\check h^j$ is a polynomially-bounded function of $g^1, \ldots, g^{\MM-1}$\ \footnote{This is the only place where we need the assumption that all nonlinearities in the program are polynomially bounded. Otherwise, the compositions of such nonlinearities might not be integrable against the Gaussian measure}, the summands of
        \begin{align*}
        \f 1 n \sum_{\alpha=1}^n
            \lp \sum_{j=1}^s |\check h_\alpha^j|\rp^d
        \end{align*}
        are polynomially-bounded functions of $g^1, \ldots, g^{\MM-1}$ too.
        So by induction hypothesis, this sum converges a.s., implying the a.s. uniform boundedness of it and
        \begin{align*}
        \f 1 n \sum_{\alpha=1}^n
                    \left|
                        \sum_{j=1}^s
                        \check h_\alpha^j \check \epsilon_j
                    \right|^d
            .
       \end{align*}
\end{itemize}
\end{proof}

\subsubsection{\texorpdfstring{$\probB$}{B} Converges Almost Surely to 0}
\label{sec:probB}

In this section we show  $\probB \asto 0.$

\paragraph{Some Notations}
For brevity, we will set $d_\alpha \defeq (\Pi^\perp_{\check H})_{\alpha\alpha}$.
In addition, for each $\alpha \in [n]$, $w \in \R$, $\tau \ge 0$, let
\[\Psi_\alpha(w; \tau^2) \defeq
            \EV_{z\sim\Gaus(0, 1)}
            \psi\lp
                g^1_\alpha, \ldots, g^{\MM-1}_\alpha, w + \tau z
                \rp.
\]
(Here and in all that follows, $\tau^2$ is the square of $\tau$, and the $2$ is not an index).
This is a random function, with randomness induced by $g^1, \ldots, g^{\MM-1}$.

\paragraph{Our proof idea} is to write
\begin{align*}
\probB
    &=
        \left| \f 1 n \sum_{\alpha=1}^n
            \Psi_\alpha\lp \omega_\alpha; \sigma^2 d_\alpha \rp
            -
            \Psi_\alpha\lp \sum_{i=1}^r \mathring v_i \hat g^i_\alpha; \mathring \sigma^2 \rp
        \right|
        \\
    &\le
        \f 1 n
        \sum_{\alpha\in U}
        \left|\Psi_\alpha\lp \omega_\alpha; \sigma^2 d_\alpha\rp\right|
        +
        \left|\Psi_\alpha\lp \sum_{i=1}^r \mathring v_i \hat g^i_\alpha; \mathring \sigma^2 \rp\right|
        \numberthis \label{eqn:smalldiagonal}
        \\
    &\qquad
        +
        \f 1 n \sum_{\alpha\in V}
        \left|
        \Psi_\alpha\lp \omega_\alpha; \sigma^2 d_\alpha\rp
        -
        \Psi_\alpha\lp \sum_{i=1}^r \mathring v_i \hat g^i_\alpha; \mathring \sigma^2 \rp
        \numberthis \label{eqn:largediagonal}
        \right|
\end{align*}
where $U \sqcup V = [n]$ is a partition of $[n]$ with $U \defeq \{\alpha: d_\alpha < 1/2\}$ and $V$ is its complement.
Note that $|U| \le 2 \rank \check H \le 2 s$ is uniformly bounded in $n$.
We then show each summand of \cref{eqn:smalldiagonal} goes to 0 a.s. independently.
Finally we use the smoothness of $\Psi_\alpha$ (\cref{eqn:PsiSmoothness}) induced by the Gaussian averaging in $\Psi_\alpha$ to show each summand of \cref{eqn:largediagonal} is almost surely $o(1/n)$, finishing the proof.

\paragraph{\cref{eqn:smalldiagonal} converges to 0 a.s.}
We first look at the term
\begin{align*}
\f 1 n
\sum_{\alpha\in U}
\left|\Psi_\alpha\lp \sum_{i=1}^r \mathring v_i \hat g^i_\alpha; \mathring \sigma^2 \rp\right|
&\le
    \f{|U|}n \max_{\alpha \in [n]}
        \left|\Psi_\alpha\lp \sum_{i=1}^r \mathring v_i \hat g^i_\alpha; \mathring \sigma^2 \rp\right|
    \\
&\le
    \f{2s}{n^{1-1/q}} \sqrt[q]{\f 1 n \sum_{\alpha \in [n]} \left|\Psi_\alpha\lp \sum_{i=1}^r \mathring v_i \hat g^i_\alpha; \mathring \sigma^2 \rp\right|^q}
    \numberthis\label{eqn:smalldiagonalEasy}
\end{align*}
for any $q > 0$.
Now $\left|\Psi_\alpha\lp \sum_{i=1}^r \mathring v_i \hat g^i_\alpha; \mathring \sigma^2 \rp\right|^q$ is a fixed (independent of $\alpha$) polynomially-bounded function of $g^1_\alpha, \ldots, g^{\MM-1}_\alpha$, so by induction hypothesis, 
\[\f 1 n \sum_{\alpha \in [n]} \left|\Psi_\alpha\lp \sum_{i=1}^r \mathring v_i \hat g^i_\alpha; \mathring \sigma^2 \rp\right|^q\]
is a.s. uniformly bounded in $n$, so that using a large $q \ge 2$, we see \cref{eqn:smalldiagonalEasy} converges a.s. to 0.

Next, we apply a similar reasoning to the other term and obtain
\begin{align*}
\f 1 n
\sum_{\alpha\in U}
\left|\Psi_\alpha\lp \omega_\alpha; \sigma^2 d_\alpha\rp\right|
&\le
    \f{2s}{n^{1-1/q}} \sqrt[q]{\f 1 n \sum_{\alpha \in [n]}
    \left|\Psi_\alpha\lp \omega_\alpha; \sigma^2 d_\alpha\rp\right|^q
    }
\end{align*}
We in fact already know that
\[\f 1 n \sum_{\alpha \in [n]}
    \left|\Psi_\alpha\lp \omega_\alpha; \sigma^2 d_\alpha\rp\right|^q\]
is a.s. uniformly bounded in $n$
from \cref{eqn:momentUpperBoundDecomposition} in \cref{sec:probA}, so that 
\[
\f 1 n
\sum_{\alpha\in U}
\left|\Psi_\alpha\lp \omega_\alpha; \sigma^2 d_\alpha\rp\right|
\asto 0\]
from which follows the same for \cref{eqn:smalldiagonal}.

\paragraph{\cref{eqn:largediagonal} converges to 0 a.s.}

As mentioned above, to prove this we will use the following smoothness bound of $\Psi_\alpha$, whose proof will be delayed to the end of the section.
Suppose, WLOG, that the polynomially boundedness of $\psi$ presents itself in an inequality $|\psi(x)| \le C \|x\|^p_{p} + C$, for some $p, C > 0$, where $p$ is an integer.
This $p$ will appear explicitly in this smoothness bound below.

\begin{lemma}[Smoothness of $\Psi_\alpha$]\label{lemma:PsiAlphaSmoothnessBound}
Let $w, \Delta w \in \R, \tau^2, \Delta \tau^2 \in \R^{\ge 0}$.
Then
\begin{align*}
&\left| \Psi_\alpha(w+\Delta w ; \tau^2 + \Delta \tau^2) - \Psi_\alpha(w; \tau^2)\right|
\\
&\quad\quad\quad\le
    R
    (|\Delta w| + \Delta \tau^2)
    (1 + \tau^{-2})
        \lp S_\alpha + |w|^p + |\Delta w|^p + \tau^p + (\Delta\tau^2)^{p/2} \rp
    \numberthis
    \label{eqn:PsiSmoothness}
\end{align*}
for some constant $R > 0$, and where
\[S_\alpha \defeq 1 + |g^1_\alpha|^{p} + \cdots + |g^{\MM-1}_\alpha|^p.\]
\end{lemma}

To bound \cref{eqn:largediagonal}, first we expand
\[
\omega_\alpha =
\sum_{i=1}^r \hat g_\alpha^i (\mathring v_i + \hat \epsilon_i)
+ \sum_{j=1}^s \check h_\alpha^j \check \epsilon_j
\]
where, by \cref{lemma:omegaExpansion}, $\hat \epsilon \in \R^r, \check \epsilon \in \R^s$ are vectors that go to 0 almost surely with $n$.
Then we apply the smoothness bound \cref{eqn:PsiSmoothness} to get, for each $\alpha \in V$
\begin{align*}
    \left|
    \Psi_\alpha\lp \omega_\alpha; \sigma^2 d_\alpha\rp
    -
    \Psi_\alpha\lp \sum_{i=1}^r \mathring v_i \hat g^i_\alpha; \mathring \sigma^2 \rp
    \right|
    &\le
        R
        \lp 1 + \min(\sigma^2 d_\alpha, \mathring \sigma^2)^{-1} \rp
        X_\alpha
        Y_\alpha
        \\
    &\le
        R
        \lp 1 + \min(\sigma^2/2, \mathring \sigma^2)^{-1} \rp
        X_\alpha
        Y_\alpha
\end{align*}
using the fact that $d_\alpha \ge 1/2, \forall \alpha \in V$.
Here
\begin{align*}
X_\alpha
    &\defeq
        |\omega_\alpha - \sum_{i=1}^r \mathring v_i \hat g^i_\alpha|
        + |\sigma^2 d_\alpha - \mathring \sigma^2|
        \\
    &=
        \left|\sum_{i=1}^r \hat g_\alpha^i \hat \epsilon_i
                + \sum_{j=1}^s \check h_\alpha^j \check \epsilon_j
        \right|
        + |\sigma^2 d_\alpha - \mathring \sigma^2|
        \\
Y_\alpha
    &\defeq
        S_\alpha + |\omega_\alpha|^p
        + \left|\sum_{i=1}^r \hat g_\alpha^i \hat \epsilon_i
                + \sum_{j=1}^s \check h_\alpha^j \check \epsilon_j
        \right|^p
        + \max(\sigma^2 d_\alpha, \mathring \sigma^2)^{p/2}
        + |\sigma^2 d_\alpha - \mathring \sigma^2|^{p/2}
        .
        \\
\end{align*}
Thus,
\begin{align*}
\cref{eqn:largediagonal}
    &=
        \f 1 n \sum_{\alpha\in V}
        \left|
        \Psi_\alpha\lp \omega_\alpha; \sigma^2 d_\alpha\rp
        -
        \Psi_\alpha\lp \sum_{i=1}^r \mathring v_i \hat g^i_\alpha; \mathring \sigma^2 \rp
        \right|
        \\
    &\le
        R
        \f 1 n
        \lp 1 + \min(\sigma^2/2, \mathring \sigma^2)^{-1} \rp
        \sum_{\alpha \in V}
            X_\alpha
            Y_\alpha
        \\
    &\le
        R
        \lp 1 + \min(\sigma^2/2, \mathring \sigma^2)^{-1} \rp
        \sqrt{\f 1 n \sum_{\alpha \in V}
            X_\alpha^2}
        \sqrt{\f 1 n \sum_{\alpha \in V}
            Y_\alpha^2
            }
        .
\end{align*}
Since $\sigma \asto \mathring \sigma$ and we have assumed $\mathring \sigma > 0$ by \cref{assm:mathringSigmaPositive}, we have $\lp 1 + \min(\sigma^2/2, \mathring \sigma^2)^{-1} \rp$ is almost surely uniformly bounded in $n$.

Thus, \cref{eqn:largediagonal} can be shown to converge a.s.\ to 0 if we show
\begin{align*}
&\sqrt{\f 1 n \sum_{\alpha \in V} Y_\alpha^2} \quad\text{is a.s.\ uniformly bounded in $n$, and}\\
&\sqrt{\f 1 n \sum_{\alpha \in V} X_\alpha^2} \asto 0
\end{align*}

We prove these two claims in \cref{lemma:YAlphaUnifBounded,lemma:XalphaAsto0} below, which would finish our proof of $\probB \asto 0$, and of our main theorem \cref{thm:netsorTMasterTheoremFormal} as well.

\begin{lemma}\label{lemma:XalphaAsto0}
$\sqrt{\f 1 n \sum_{\alpha \in V} X_\alpha^2} \asto 0$.
\end{lemma}
\begin{proof}
    Note that
    \begin{align*}
    X_\alpha
        &\le
            \left|\sum_{i=1}^r \hat g_\alpha^i \hat \epsilon_i
                    + \sum_{j=1}^s \check h_\alpha^j \check \epsilon_j
            \right|
            + |\mathring \sigma^2 - \sigma^2|
            + |\sigma^2 - \sigma^2 d_\alpha|
            \\
        &\defeq
            P_\alpha + Q_\alpha + R_\alpha
            .
    \end{align*}
    Then by triangle inequality (in $\ell_2$-norm),
    \begin{align*}
    \sqrt{\f 1 n \sum_{\alpha \in V} X_\alpha^2}
        &\le
            \sqrt{\f 1 n \sum_{\alpha \in V} P_\alpha^2}
            + \sqrt{\f 1 n \sum_{\alpha \in V} Q_\alpha^2}
            + \sqrt{\f 1 n \sum_{\alpha \in V} R_\alpha^2}
            .
    \end{align*}
    We now show that each term above converges a.s.\ to 0, which would finish the proof of \cref{lemma:XalphaAsto0}.
    \begin{itemize}
    \item
        Because $\hat \epsilon \asto 0$ and $\check \epsilon \asto 0$, we have
        \begin{align*}
        \f 1 n \sum_{\alpha \in V} P_\alpha^2
            &\le
                C_8 \f 1 n \sum_{\alpha \in V} \lp
                    \sum_{i=1}^r (\hat g_\alpha^i \hat \epsilon_i)^2
                    + \sum_{j=1}^s (\check h_\alpha^j \check \epsilon_j)^2
                    \rp
                \\
            &\le
                C_8 \max_{i,j}\{|\hat \epsilon_i|, |\check \epsilon_j|\} \times 
                \f 1 n \sum_{\alpha \in V} \lp
                    \sum_{i=1}^r (\hat g_\alpha^i)^2
                    + \sum_{j=1}^s (\check h_\alpha^j)^2
                    \rp
                \\
            &\le
                C_8 \max_{i,j}\{|\hat \epsilon_i|, |\check \epsilon_j|\} \times 
                \f 1 n \sum_{\alpha \in [n]} \lp
                    \sum_{i=1}^r (\hat g_\alpha^i)^2
                    + \sum_{j=1}^s (\check h_\alpha^j)^2
                    \rp
                \\
            &\asto
                C_8 \times 0 \times \mathcal{E} = 0
        \end{align*}
        where $\mathcal{E}$ is the Gaussian expectation that $\f 1 n \sum_{\alpha \in [n]} \lp
                    \sum_{i=1}^r (\hat g_\alpha^i)^2
                    + \sum_{j=1}^s (\check h_\alpha^j)^2
                    \rp$ converges a.s. to, by inductive hypothesis.
    \item
        The quantity $Q_\alpha$ actually doesn't depend on $\alpha$, so that
        \[\sqrt{\f 1 n \sum_{\alpha \in V} Q_\alpha^2} \le |\mathring \sigma^2 - \sigma^2| \asto 0\]
        by \cref{lemma:sigmaConverges}.
    \item
        Notice $R_\alpha^2 = \sigma^4 (1 - d_\alpha)^2 \le \sigma^4 (1 - d_\alpha)$ because $1 - d_\alpha \in [0, 1/2]$.
        Thus,
        \begin{align*}
        \f 1 n \sum_{\alpha \in V} R_\alpha^2
            &\le
                \sigma^4 \f 1 n \sum_{\alpha \in V} 1 - d_\alpha
                \\
            &\le 
                \sigma^4 \f 1 n \sum_{\alpha \in [n]} 1 - d_\alpha
                \\
            &=
                \sigma^4 \f 1 n \rank \check H
        \end{align*}
        by the definition that $d_\alpha = (\Pi^\perp_{\check H})_{\alpha\alpha}$.
        But of course $\rank \check H \le s$ is bounded relative to $n$.
        So this quantity goes to 0 (surely) as desired.
    \end{itemize}
\end{proof}

\begin{lemma}\label{lemma:YAlphaUnifBounded}
$\sqrt{\f 1 n \sum_{\alpha \in V} Y_\alpha^2}$ is a.s.\ uniformly bounded in $n$.
\end{lemma}
\begin{proof}
We have
\begin{align*}
\sqrt{\f 1 n \sum_{\alpha \in V} Y_\alpha^2}
    &\le
        \sqrt{\f 1 n \sum_{\alpha \in V} S_\alpha^2}
        + \sqrt{\f 1 n \sum_{\alpha \in V} |\omega_\alpha|^{2p}}
        + \sqrt{\f 1 n \sum_{\alpha \in V} X'_\alpha{}^2}
        + \sqrt{\f 1 n \sum_{\alpha \in V} \max(\sigma^2 d_\alpha, \mathring \sigma^2)^p}
        \\
    &\le
        \sqrt{\f 1 n \sum_{\alpha \in [n]} S_\alpha^2}
        + \sqrt{\f 1 n \sum_{\alpha \in [n]} |\omega_\alpha|^{2p}}
        + \sqrt{\f 1 n \sum_{\alpha \in [n]} X'_\alpha{}^2}
        + \sqrt{\f 1 n \sum_{\alpha \in [n]} \max(\sigma^2 d_\alpha, \mathring \sigma^2)^p}
\end{align*}
where
\[X'_\alpha \defeq \left|\sum_{i=1}^r \hat g_\alpha^i \hat \epsilon_i
                + \sum_{j=1}^s \check h_\alpha^j \check \epsilon_j
        \right|^p
        + |\sigma^2 d_\alpha - \mathring \sigma^2|^p
\]
We proceed to show that each of 4 summands above are individually a.s.\ uniformly bounded in $n$.
\begin{itemize}
\item
    $S_\alpha^2$ is a polynomially bounded function of $g^1_\alpha, \ldots, g^{\MM-1}_\alpha$, so that by \ref{IH:MomConv}$(\MM-1)$,
    \[\f 1 n \sum_{\alpha \in [n]} S_\alpha^2 \asto C\]
    for some constant $C$, so it is also a.s.\ uniformly bounded in $n$.
\item
    By \cref{lemma:omegaAlphaUnifomBounded}, we get
    \begin{equation*}
    \f 1 n \sum_{\alpha \in [n]} |\omega_\alpha|^{2p}
    \end{equation*}
    is a.s.\ uniformly bounded in $n$.
\item
    Using the same reasoning as in the proof of \cref{lemma:XalphaAsto0}, one can easily show
    \[
    \f 1 n \sum_{\alpha \in [n]} X'_\alpha{}^2 \asto 0
    \]
    so it is also a.s.\ uniformly bounded. 
\item
    Since $d_\alpha \le 1$, we have $\max(\sigma^2 d_\alpha, \mathring\sigma^2) \le \max(\sigma^2, \mathring \sigma^2)$, which is independent of $\alpha$.
    Therefore,
    \begin{align*}
    \f 1 n \sum_{\alpha \in [n]} \max(\sigma^2 d_\alpha, \mathring \sigma^2)^p
    &\le
        \f 1 n \sum_{\alpha \in [n]} \max(\sigma^2, \mathring \sigma^2)^p
        \\
    &=
        \max(\sigma^2, \mathring \sigma^2)^{p/2}
    \asto \mathring \sigma^p.
    \end{align*}
    Therefore, it is also a.s.\ uniformly bounded in $n$.
\end{itemize}

\end{proof}

Finally, we deliver the promised proof of \cref{lemma:PsiAlphaSmoothnessBound}.

\begin{proof}[Proof of \cref{lemma:PsiAlphaSmoothnessBound}]

By \cref{lemma:stein}, $\Psi_\alpha$ is differentiable in $w$, and
\begin{align}
    \pd_w \Psi_\alpha(w; \tau^2)
    &=
        \inv \tau \EV_{z \sim \Gaus(0, 1)} z \psi(g^1_\alpha, \ldots, g^{\MM - 1}_\alpha, w+ \tau z)
        \label{eqn:PsiAlphaDW}
        \\
    \pd_{\tau^2} \Psi_\alpha(w; \tau^2)
    &=
        \f 1 2 \tau^{-2}
        \EV_{z \sim \Gaus(0, 1)} (z^2-1) \psi(g^1_\alpha, \ldots, g^{\MM - 1}_\alpha, w+ \tau z)
        .
        \label{eqn:PsiAlphaDTau2}
\end{align}

Recall that $|\psi(x)| \le C \|x\|^p_{p} + C$.
We will silently introduce constants $C_1, C_2, \ldots$ depending only on $p$, merging with old constants, typically via \cref{lem:powerbound} or by integrating out some integrands depending only on $p$.
With $z \sim \Gaus(0, 1)$,
\begin{align*}
|\pd_w \Psi_\alpha(w; \tau^2)|
    &\le
        \inv \tau \EV_{z} |z| |\psi(g^1_\alpha, \ldots, g^{\MM - 1}_\alpha, w+ \tau z)|
        \\
    &\le
        \inv \tau C\EV_z |z| \lp 1 + |g^1_\alpha|^{p} + \cdots + |g^{\MM-1}_\alpha|^p + |w + \tau z|^p \rp
        \\
    &\le
        \inv \tau C_1\EV_z |z| \lp 1 + |g^1_\alpha|^{p} + \cdots + |g^{\MM-1}_\alpha|^p + |w|^p + \tau^p |z|^p \rp
        \\
    &\le
        \inv \tau C_2 \lp 1 + |g^1_\alpha|^{p} + \cdots + |g^{\MM-1}_\alpha|^p + |w|^p + \tau^p \rp
        .
\end{align*}
Similarly,
\begin{align*}
|\pd_{\tau^2} \Psi_\alpha(w; \tau^2)|
    &\le
        \f 1 2 \tau^{-2} \EV_z |z^2-1| |\psi(g^1_\alpha, \ldots, g^{\MM - 1}_\alpha, w+ \tau z)|
        \\
    &\le
        \tau^{-2} C_3 \lp 1 + |g^1_\alpha|^{p} + \cdots + |g^{\MM-1}_\alpha|^p + |w|^p + \tau^p \rp
        .
\end{align*}

Therefore, for any $\Delta w \in \R, \Delta \tau^2 \in \R^{\ge 0}$, we have
\begin{align*}
&\phantomeq
    \left| \Psi_\alpha(w+\Delta w; \tau^2 + \Delta \tau^2) - \Psi_\alpha(w; \tau^2)\right|
    \\
&=
    \left|
    \int_0^1 \dd t \lp 
        \Delta w \cdot \pd_w \Psi_\alpha(w + \Delta w t; \tau^2 + \Delta \tau^2 t) +
        \Delta \tau^2 \cdot \pd_{\tau^2} \Psi_\alpha(w + \Delta w t; \tau^2 + \Delta \tau^2 t)
        \rp
    \right|
    \\
&\le
    \int_0^1 \dd t \lp 
        |\Delta w| \cdot |\pd_w \Psi_\alpha(w + \Delta w t; \tau^2 + \Delta \tau^2 t)| +
        |\Delta \tau^2| \cdot |\pd_{\tau^2} \Psi_\alpha(w + \Delta w t; \tau^2 + \Delta \tau^2 t)|
        \rp
    \\
&\le
    (C_2 + C_3)
    (|\Delta w| + |\Delta \tau^2|)
    \\
&\quad
    \int_0^1 \dd t
        ((\tau^2 + \Delta \tau^2 t)^{-1/2} + (\tau^2 + \Delta \tau^2 t)^{-1})
        \times
        \lp S_\alpha + |w+\Delta w t|^p + (\tau^2 + \Delta\tau^2 t)^{p/2} \rp
\end{align*}
where for brevity we have set 
\[S_\alpha \defeq 1 + |g^1_\alpha|^{p} + \cdots + |g^{\MM-1}_\alpha|^p,\]
which is independent of $t$.

Since $\Delta \tau^2 \ge 0$, $(\tau^2 + \Delta \tau^2 t)^{-1} \le \tau^{-2}$, and we get
\begin{align*}
&\phantomeq
    \left| \Psi_\alpha(w+\Delta w; \tau^2 + \Delta \tau^2) - \Psi_\alpha(w; \tau^2)\right|
    \\
&\le
    C_4
    (|\Delta w| + \Delta \tau^2)
    (\inv \tau + \tau^{-2})
    \int_0^1 \dd t
        \lp S_\alpha + |w+\Delta w t|^p + (\tau^2 + \Delta\tau^2 t)^{p/2} \rp
    \\
&\le
    C_5
    (|\Delta w| + \Delta \tau^2)
    (\inv \tau + \tau^{-2})
    \int_0^1 \dd t
        \lp S_\alpha + |w|^p + |\Delta w|^p t^p + \tau^p + (\Delta\tau^2)^{p/2} t^{p/2} \rp
    \\
&\le
    C_6
    (|\Delta w| + \Delta \tau^2)
    (\inv \tau + \tau^{-2})
        \lp S_\alpha + |w|^p + |\Delta w|^p + \tau^p + (\Delta\tau^2)^{p/2} \rp
\end{align*}
where in the end we have integrated out $t^p$ and $t^{p/2}$.
We finally apply the simplification $\tau^{-1} \le \f 1 2 + \f 1 2 \tau^{-2}$ by AM-GM to get the desired \cref{eqn:PsiSmoothness}.

\end{proof}

\end{document}